\documentclass[5p,times]{elsarticle}

\usepackage{tikz}
\usetikzlibrary{shapes.geometric, arrows}
\tikzstyle{startstop} = [rectangle, rounded corners, minimum width=3cm, minimum height=1cm,text centered, draw=black, fill=red!30]
\tikzstyle{io} = [trapezium, trapezium left angle=70, trapezium right angle=110, minimum width=3cm, minimum height=1cm, text centered, draw=black, fill=blue!30]
\tikzstyle{process} = [rectangle, text width= 6cm, minimum height=1cm, text centered, draw=black,fill=white!30]
\tikzstyle{decision} = [diamond, minimum width=3cm, minimum height=1cm, text centered, draw=black, fill=green!30]
\tikzstyle{arrow} = [thick,->,>=stealth]
\usepackage{multirow}
\usetikzlibrary{backgrounds,calc,positioning}

\usepackage[super]{nth}
\usepackage{setspace}

\usepackage{enumitem}
\usepackage{pgfplots}
\usepackage{comment}
\usepackage{booktabs}

\usepackage{mathptmx}
\usepackage{subcaption}
\usepackage{rotating}
\usepackage{graphicx}
\usepackage{array,etoolbox}
\preto\tabular{\setcounter{magicrownumbers}{0}}
\newcounter{magicrownumbers}
\def\rownumber{}

\usepackage[
  separate-uncertainty = true,
  multi-part-units = repeat
]{siunitx}

\usepackage{soul}

\begin{document}
\begin{frontmatter}

\title{Sci-Net: Scale Invariant Model for Buildings Segmentation from Aerial Imagery}

\author[First]{Hasan Nasrallah}
\author[Second]{Mustafa Shukor}
\author[Fourth]{Ali J. Ghandour\corref{cor1}}
\address[First]{CRSI, Faculty of Engineering, Lebanese University | hnasrallah@geogroup.ai}
\address[Second]{ISIR (MLIA), Sorbonne University, Paris, France | mshukor@geogroup.ai}
\cortext[cor1]{Corresponding Author.\\Email: aghandour@cnrs.edu.lb}
\address[Fourth]{National Center for Remote Sensing - CNRS, Beirut, Lebanon | aghandour@cnrs.edu.lb}

\begin{abstract}
Buildings' segmentation is a fundamental task in the field of earth observation and aerial imagery analysis. Most existing deep learning-based methods in the literature can be applied to a fixed or narrow-range spatial resolution imagery. In practical scenarios, users deal with a broad spectrum of image resolutions. Thus, a given aerial image often needs to be re-sampled to match the spatial resolution of the dataset used to train the deep learning model, which results in a degradation in segmentation performance. To overcome this challenge, we propose, in this manuscript, Scale-invariant Neural Network (Sci-Net) architecture that segments buildings from wide-range spatial resolution aerial images. Specifically, our approach leverages UNet hierarchical representation and Dense Atrous Spatial Pyramid Pooling to extract fine-grained multi-scale representations. Sci-Net significantly outperforms state of the art models on the Open Cities AI and the Multi-Scale Building datasets with a steady improvement margin across different spatial resolutions.
\end{abstract}


\end{frontmatter}

\section{Introduction}
Semantic segmentation is one of the most investigated computer vision topics, where the aim is to provide a pixel-wise classification over several classes in a particular image. With the current deep learning breakthrough, several fully connected neural network models are proposed for semantic segmentation~\cite{deeplabv3}~\cite{panet}~\cite{FPN_paper}~\cite{unet}~\cite{hrnet}~\cite{psp}, and employed for various applications such as autonomous driving \cite{kaymak2019brief_autonomous}, buildings footprint extraction \cite{delassus2018cnns}\cite{jiwani2021semantic} and medical applications \cite{lei2020medical}.

Buildings' footprint segmentation from aerial imagery \cite{LFE-small}\\ \cite{ternausv2}\cite{undoc-builds}\cite{ra-fcn}\cite{ggcnn} is important for wide range of applications such as urban planning, disaster assessment and change analysis. In addition, several international competitions have addressed deep-leaning-based buildings' segmentation challenging topics such as nadir-angle selection, noisy data impact and fusion of optical and non-optical collections. DIUx's xView2 \cite{gupta2019creating_xbd}, SpaceNet challenges (1, 2, 4, 5 and 7) \cite{van2021multi}, and Open Cities AI \cite{open_cities} are examples of recent well-known competitions focusing on this research topic.

\begin{figure*}[t]
    \begin{center}
        \subfloat[2 cm/pixel]{\includegraphics[width = 2in, height = 0.9in]{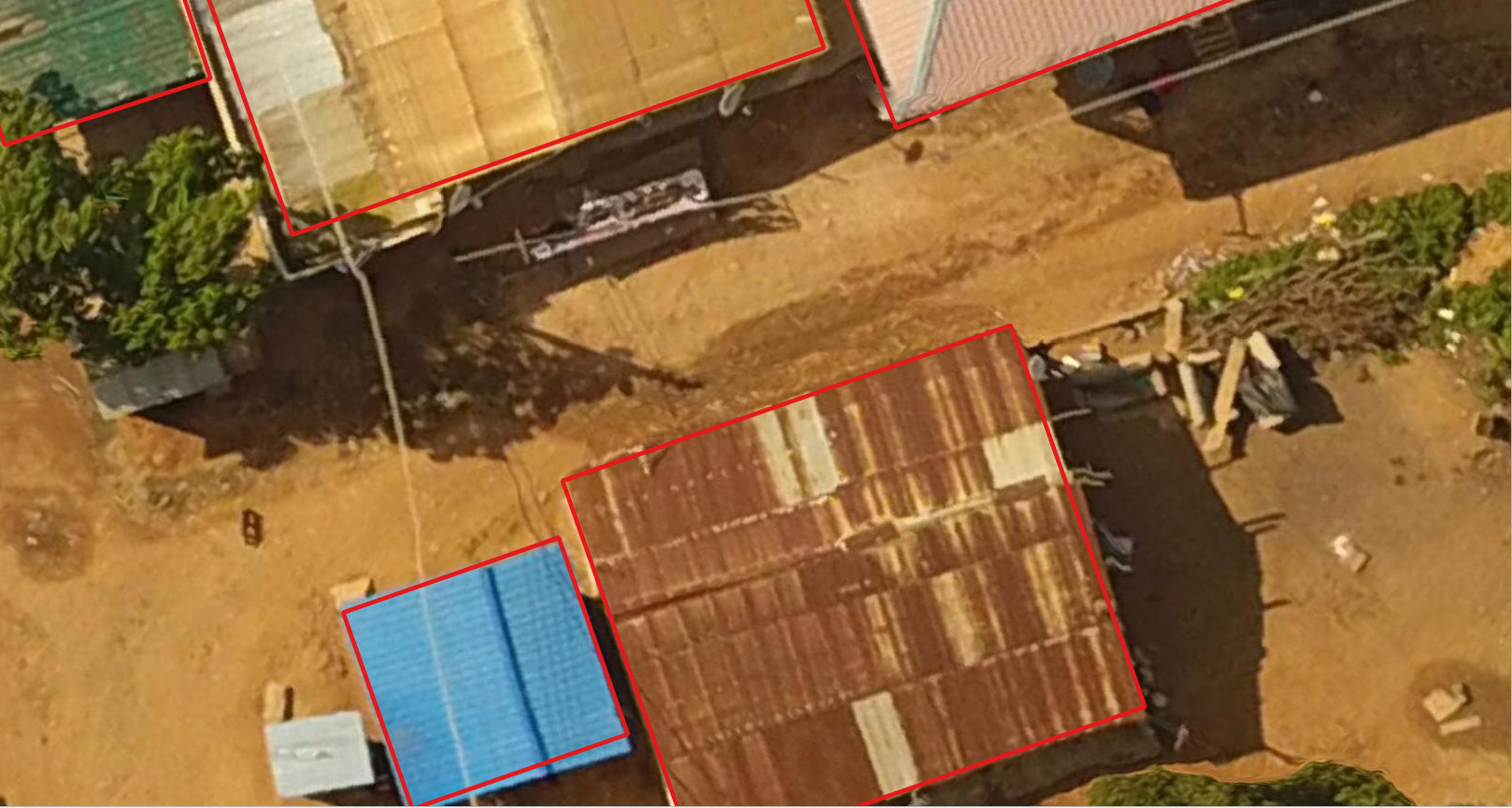}}
        \subfloat[3 cm/pixel]{\includegraphics[width = 2in, height = 0.9in]{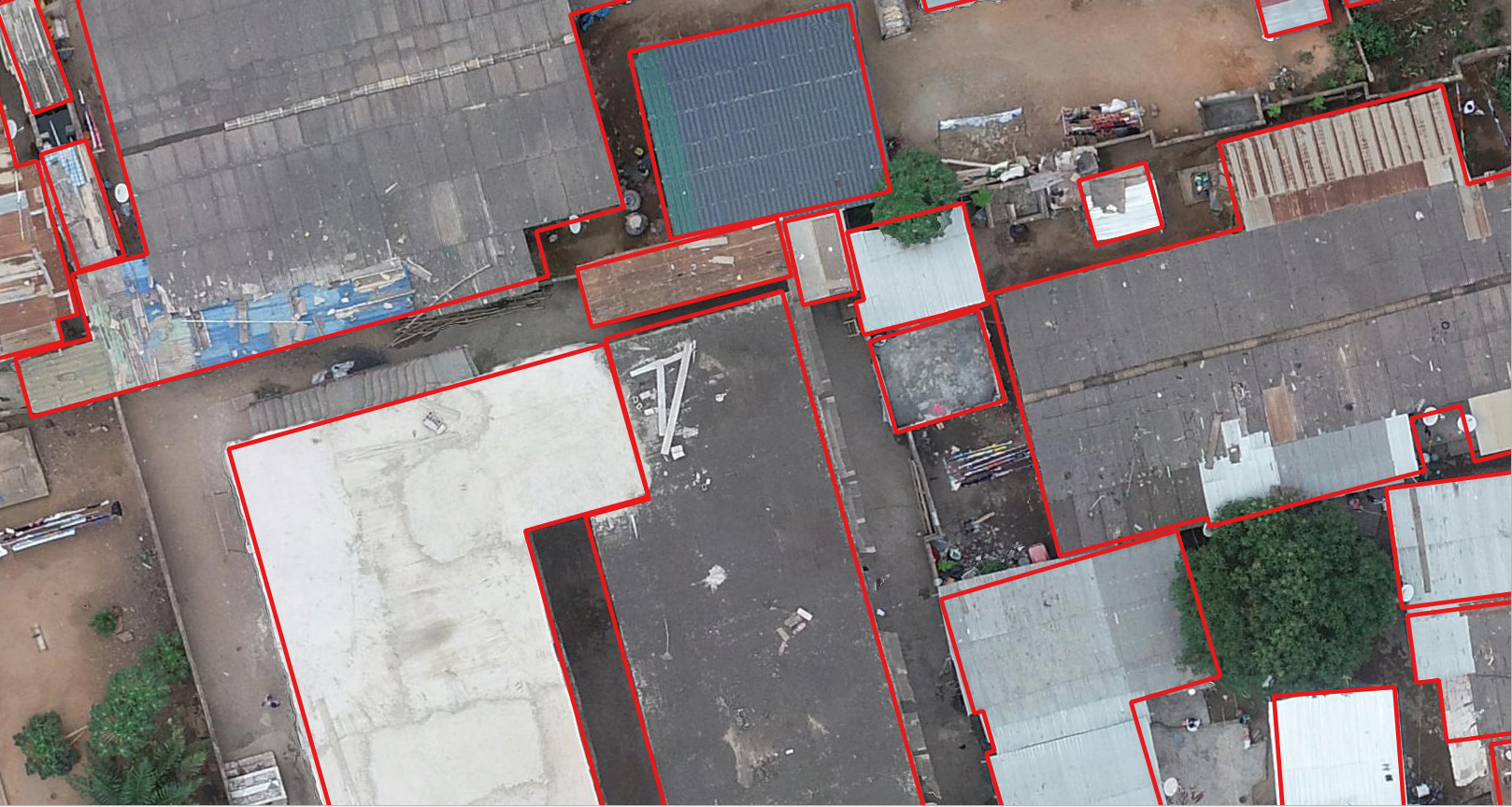}}
        \subfloat[4 cm/pixel]{\includegraphics[width = 2in, height = 0.9in]{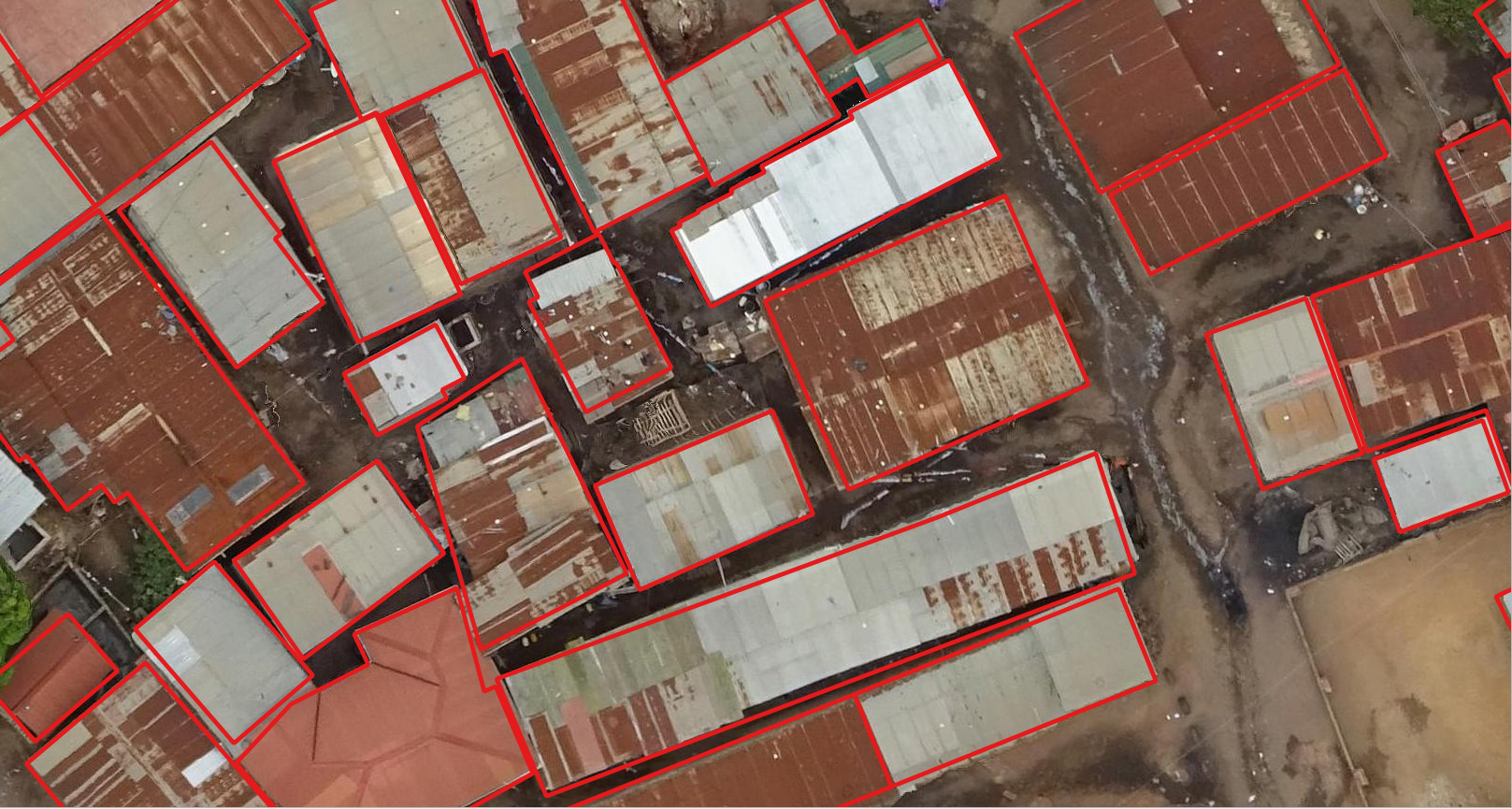}} \\
        \subfloat[5 cm/pixel]{\includegraphics[width = 2in, height = 0.9in]{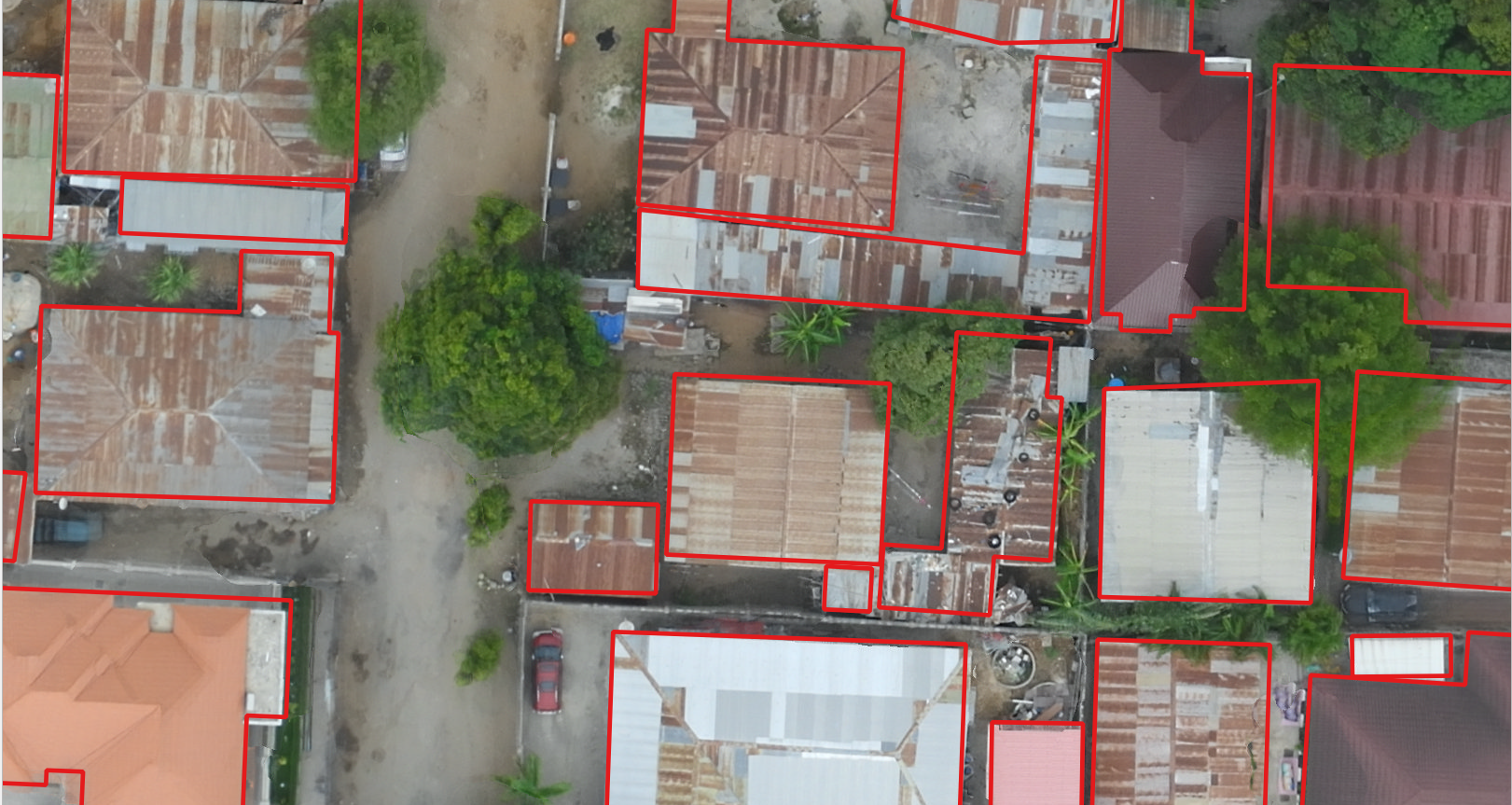}}
        \subfloat[6 cm/pixel]{\includegraphics[width = 2in, height = 0.9in]{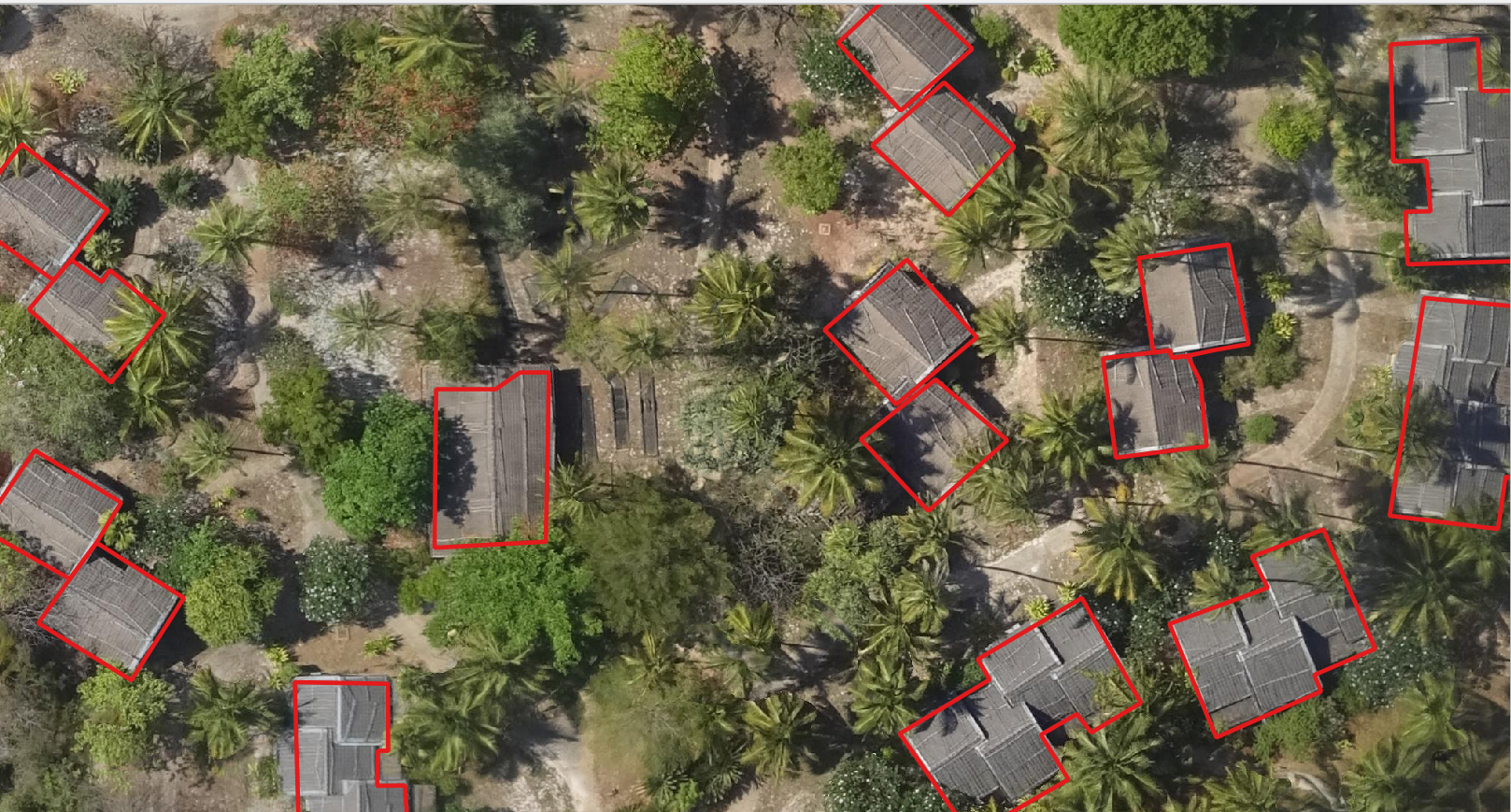}}
        \subfloat[7 cm/pixel]{\includegraphics[width = 2in, height = 0.9in]{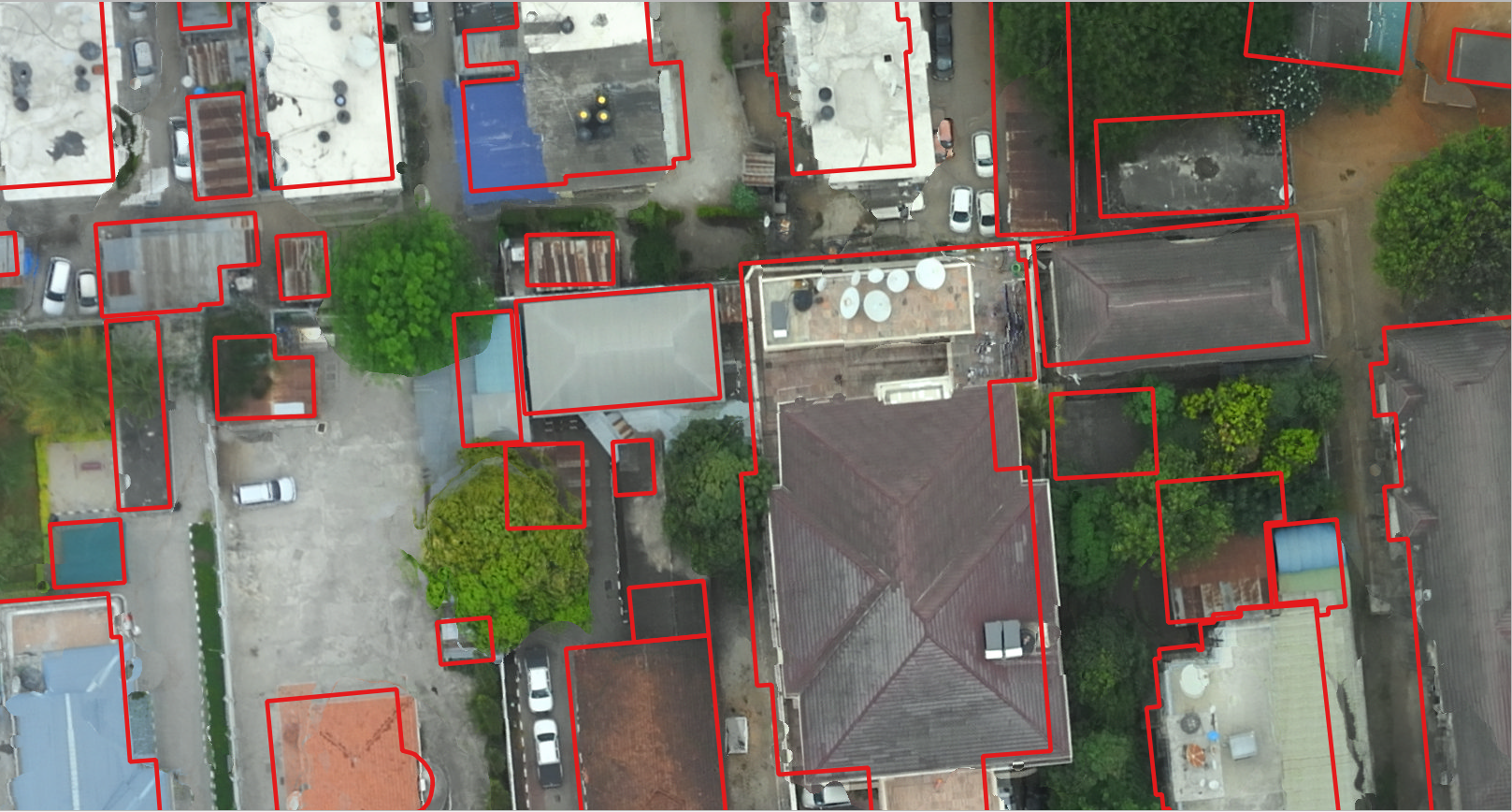}} \\
        \subfloat[8 cm/pixel]{\includegraphics[width = 2in, height = 0.9in]{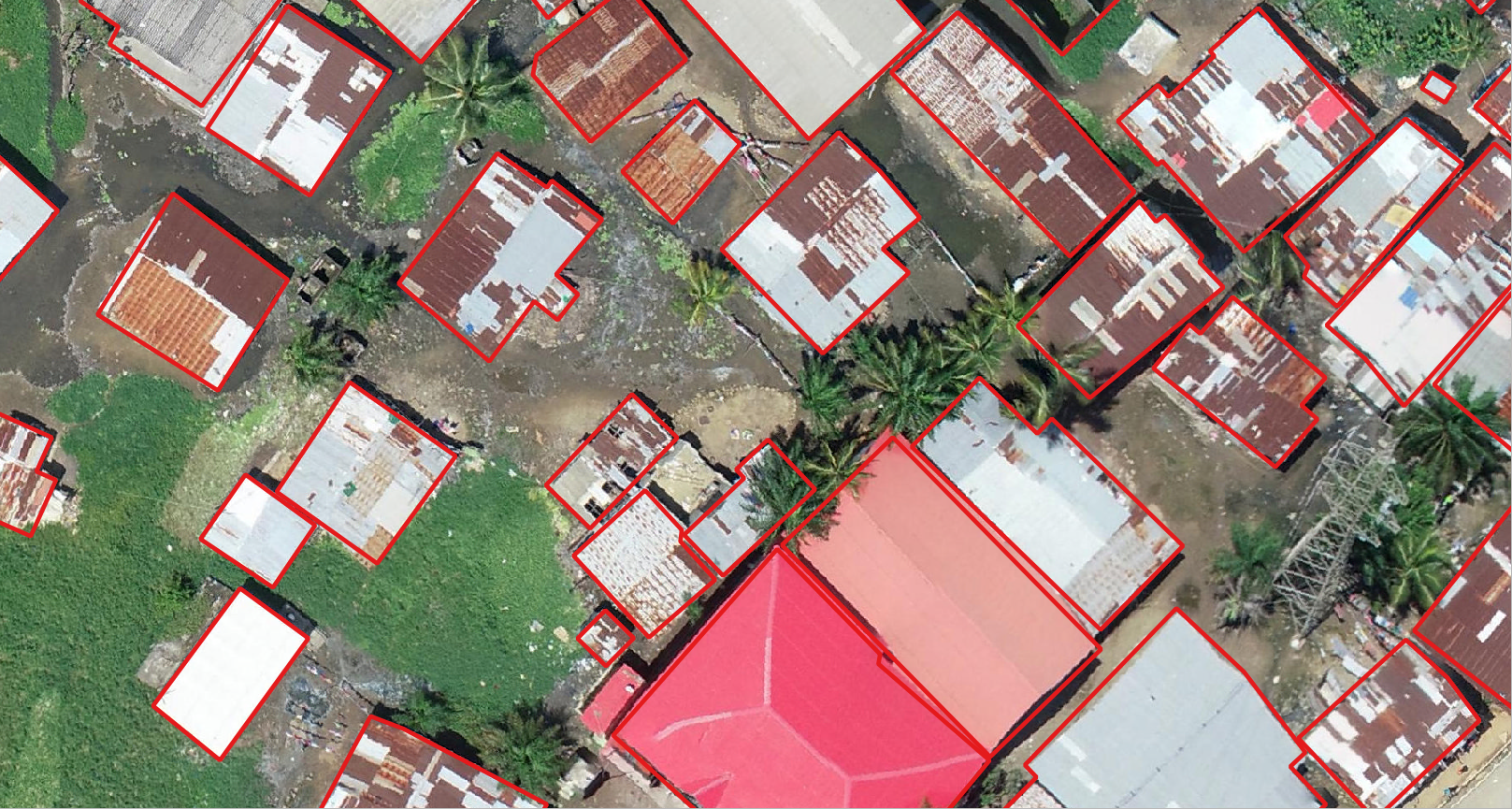}}
        \subfloat[10 cm/pixel]{\includegraphics[width = 2in, height = 0.9in]{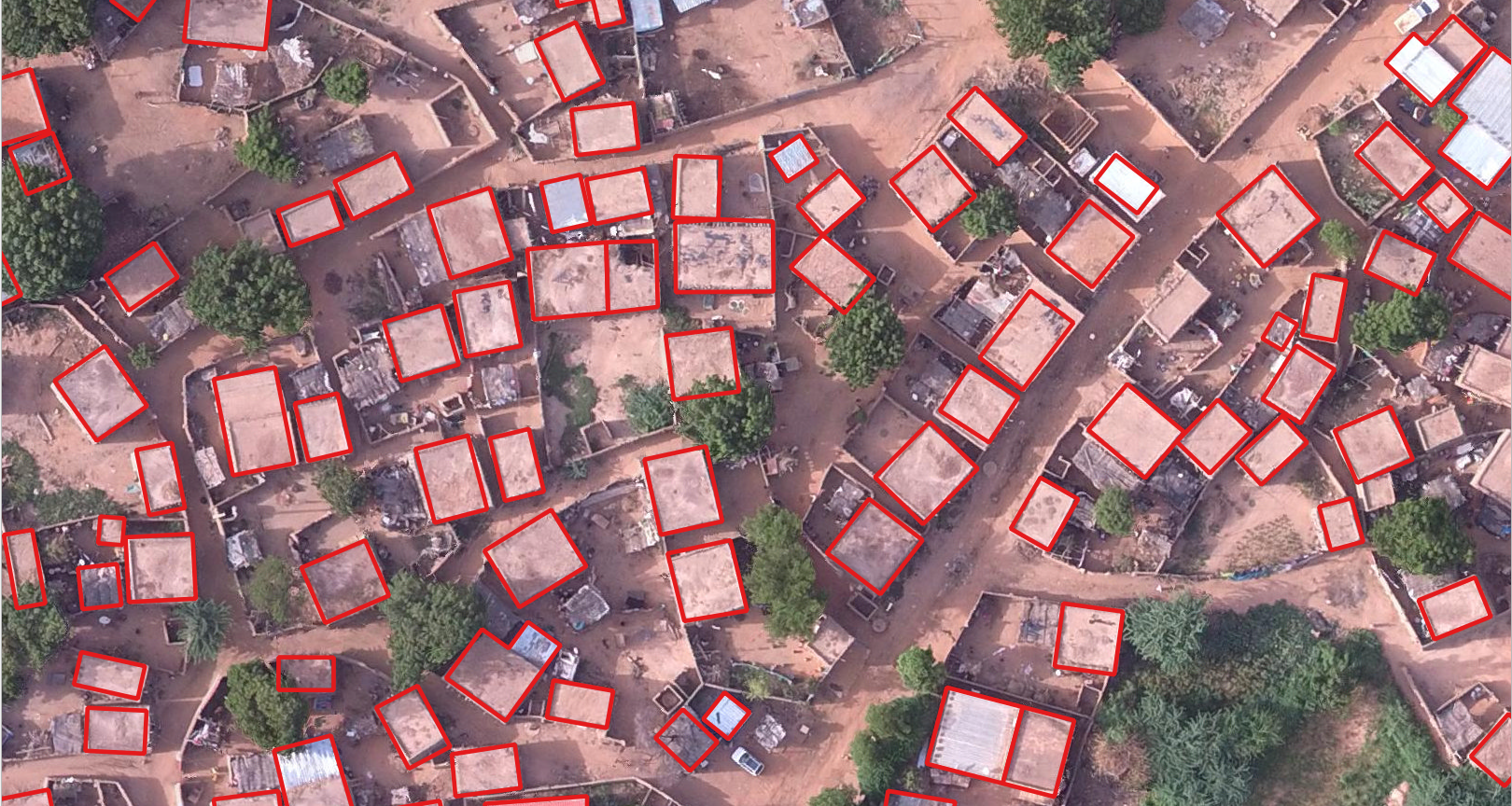}}
        \subfloat[20 cm/pixel]{\includegraphics[width = 2in, height = 0.9in]{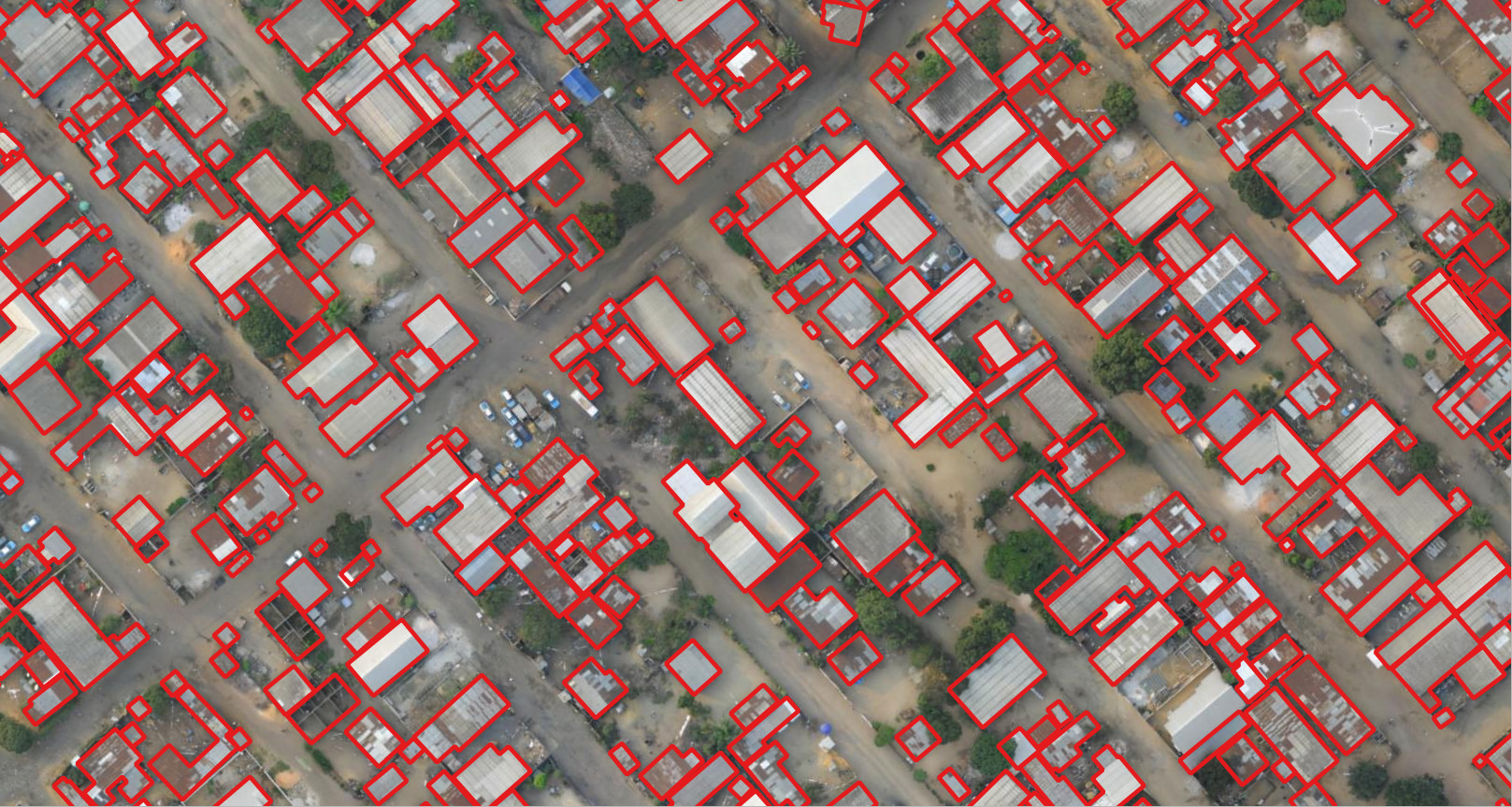}}
    \end{center}
    \caption{Sample images of different native resolution (in cm/pixel) from the Open Cities AI dataset.}
\label{samples_res}
\end{figure*}

Existing buildings' segmentation models are trained on a fixed spatial resolution. Training a robust and accurate deep learning model capable of segmenting buildings from a wide range of input spatial resolution images remains little investigated in the literature. Sate-of-the-art buildings' segmentation models perform well on test images of the exact spatial resolution as the training dataset used to generate the model. However, in practical scenarios, test images might be of various resolutions, resulting in non-optimal performance. This can be attributed to several issues: First, the fragmentation of building segments in high-resolution images, as the model fails to acquire a large enough receptive field to classify pixels closer to the center of large buildings accurately. Second, in low-resolution test images the model seamlessly merges the pixels of small building instances with the background, this is known as under-segmentation of buildings' instances, where the model suffers from an over-segmentation of the background, leaving false negative holes in the mask. These problems are commonly addressed by re-sampling the test images to match the resolution of the training dataset. However, as the gap between the spatial resolution of inference and training images increases, in both directions, the quality of the resultant segmentation masks deteriorates.

In this context, we propose Scale-invariant neural network (Sci-Net) architecture that is able to extract a multi-scale representation with wider receptive field of an aerial image to cope with varying spatial resolutions during inference time. The contribution of this paper is three-fold: \textit{(i)} show that existing SoA buildings' segmentation models often suffer from fragmentation, under-segmentation, or over-segmentation, \textit{(ii)} propose Sci-Net, a new model that significantly outperforms existing approaches across different image resolutions and finally \textit{(iii)} present model evaluation results using Open Cities AI scale-invariant testset with spatial resolution varying from 2cm/pixel to 20cm/pixel and Multi-Scale Building dataset with spatial resolution varying from 5cm/pixel to 2.5m/pixel.

The rest of the paper is organized as follows: Section \ref{review} reviews current research in the literature related to buildings' segmentation from aerial images. Practical problems in the process of buildings' segmentation from aerial images are discussed in Section \ref{problemdes}. Section \ref{method} introduces the proposed Sci-Net model and related background details. Section \ref{datasets} describes the Open Cities AI and the Multi-Scale Building datasets along with the training procedure adopted. Experimental results and ablation studies are reported in Section \ref{resultss}. Finally, the manuscript is concluded in Section \ref{conclusion}.

\section{Related Work}
\label{review}
In this section, we details some related work relevant to the problem in hand.
\subsection{Multi-Scale Context for Semantic Segmentation:}

Capturing multi-scale context has gained a lot of attention due to its importance for semantic segmentation.

To this end, several methods have been proposed. Image Pyramid methods \cite{chen2016attention}\cite{Farabet_multi}\cite{lin2016efficient}\cite{pinheiro2014recurrent} are one of the first approaches, where the feature extractor is applied on the same input with varying resolutions, then different aggregation mechanisms are applied to gather the features from all resolutions.

Encoder-Decoder approaches \cite{segnet}\cite{refine_gated}\cite{laplacian}\cite{lin2017refinenet}\cite{unet} have proven to be successful, where the input is processed by a feature extractor that reduces the spatial size and increases the number of channels progressively, then the decoder tries to decode the features and produce the output map. These approaches exploit the multi-scale features in the encoder (\emph{e.g.} using skip connections \cite{unet} or transferred pool indices \cite{segnet}).

Spatial Pyramid Pooling are another way to capture global context, methods like DeepLab \cite{deeplabv3} rely on dilated convolutions \cite{dilated-convs} to process the feature map using different rates in parallel. PSPNet \cite{psp} proposes to process different pooled feature maps with different resolution.
In~\cite{SA-Net}, authors introduce a modified version of squeeze and excitation blocks denoted as Squeeze and Attention (SA) module. SA module re-weights spatial locations in the features according to the local and global context, thus, improving semantic segmentation. Dilated or Atrous convolutions are widely used in this context \cite{wang2018understanding, dai2017deformable}

\subsection{Multi-Scale Context for Semantic Segmentation in Remote Sensing:}
Being able to segment objects with multiple resolutions is of high interest for the remote sensing community. While most of the work proposed in the computer vision community can be adapted to satellite images, several work have been also proposed in this context. \cite{Hamaguchi_2018_CVPR_Workshops} leverage multi-task learning and distillation to produce several output maps depending on the buildings size. \cite{Li2018AMR} propose to reuse previous feature maps by the help of the connections from each layer to the same sized subsequent layers. \cite{ESFNet} propose an efficient model based on separable factorized residual block in addition to dilated convolution.

Authors in~\cite{ra-fcn} introduce channel relation module that applies global average pooling over the features and spatial relation module to obtain global spatial relation features capable of capturing global contextual dependencies for identifying various objects. However, the validation of their results is based on the Postdam dataset with a fixed high resolution.

Furthermore, authors in~\cite{LFE-small} introduce the Local Feature Extractor (LFE) module, which is composed of a series of dilated convolutions of decreasing rates, after aggressively increasing the rates of dilated convolutions used in the front-end module to attain a high receptive field throughout the feature extraction process. They show that LFE module helps with tiny objects by recovering the spatial inconsistency and extracting local structure at higher layers.

A semantic segmentation network for building footprint extraction from satellite imagery using a modified DeepLabV3+ model was suggested in~\cite{jiwani2021semantic}. The performance of the proposed model is measured on test datasets with fixed resolution (and not scale-invariant). Authors in ~\cite{wang2021scale} present a scale-aware model for multi-resolution aerial imagery semantic segmentation using the LandCover.ai dataset which contains images of two different resolutions (0.25 and 0.5 m/pixel). They also conduct experiments on resized versions of the MSR Vaihingen dataset at different scales. Authors in \cite{msb_dataset} introduce MSB, a deep dilated CNN for automatic building footprint extraction. The model was trained and evaluated on the Multi-Scale Building dataset; a relatively small dataset collected from different public datasets, after undergoing horizontally/vertically flip augmentation and saturation transform, resulting in a total of 2,868 images. Finally, building extraction from remote sensing images using a Unet +inally, building extraction from remote (ASPP) model on the WHU dataset is discussed in ~\cite{wang2022building}.

\section{Problem Description}
\label{problemdes}

\subsection{Challenges}
Designing a model that is capable of segmenting buildings footprint from aerial images at different spatial resolutions faces the following challenges:

\begin{enumerate}[label=(\roman*)]
    \item \textbf{Features Resolution}: Most Feature extractors\cite{dpn}~\cite{resnet}\cite{inception}\\ \cite{effnet}\cite{resnext} are a series of five down-sampling stages, where each stage outputs denser and more meaningful representations than the previous one. However, features spatial resolution is reduced to half at each stage using a 2x2 pooling operation or a convolution with a stride $= 2$. This reduction leads to a loss in spatial information the deeper we go in the network, as the features extracted by the last stage have a resolution $32\times$ smaller than the input size (output stride $= 32$).

    \item \textbf{Field of View}: In feature extraction networks, convolutions are applied with a $3\times3$ receptive field (kernel size). Although this works very well in segmenting small to medium-sized objects, it often fails when dealing with larger objects, (\emph{i.e.}, building footprints at very high resolutions such as $2$cm/pixel). In the latter case, predicted segments often suffer from fragmentation, under-segmentation, and noise because the field of view is too small for the network to decide if the pixel belongs to a larger object or the background. On the contrary, increasing receptive field by applying convolutions with an increasing kernel size leads to losing local spatial information and exponential growth in both time and computational complexity.
\end{enumerate}

\subsection{Motivation}
\label{motivation}
Motivated by the observations above, we propose the following solution:
\begin{enumerate}[label=(\roman*)]
\item To avoid losing spatial features details and keep a reasonable computational complexity, we adopt the skip connections from the encoder to the decoder.
\item To increase the field of view or the receptive field: \textit{(a)} we use an encoder/decoder framework where the feature map spatial size decreases/increases with depth for the encoder/decoder, \textit{(b)} we include Dense ASPP with dilated convolutions, at the bottleneck of the encoder.
\end{enumerate}
To this end, we propose Sci-Net; a UNet like architecture augmented with Dense ASPP modules. We provide more details about Sci-Net architecture in the following section.

\begin{figure}[h]
\begin{center}
\includegraphics[width = 1.0\linewidth]{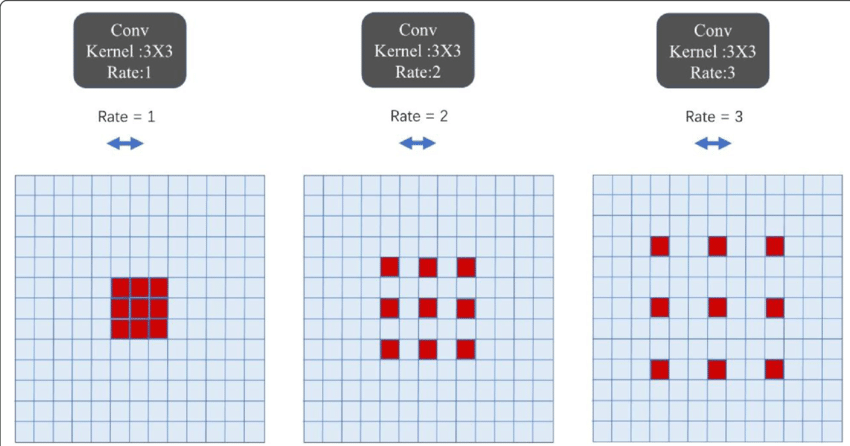}
\caption{Dilated $3\times3$ convolutions with rates = 1, 2 and 3 acquiring $3\times3$, $5\times5$ and $7\times7$ receptive fields, respectively.}
\label{atrousfig}
\end{center}
\end{figure}

\section{Sci-Net}
\label{method}

\subsection{Dilated Convolutions}
Dilated convolutions work just like regular convolutions; however, they manage to increase the size of the receptive field by the insertion of holes between the weights of the kernel, according to a selected rate denoted by $r$. For a $2$-dimensional input feature map $x$, the output feature map $y$ obtained as a result of a dilated convolution at every spatial location $i$ with a rate $r$ is defined according to Equation \ref{dilated_conv_eq}:

\begin{equation}
y[i] = \sum(x[i+r*k]*w[k]).
\label{dilated_conv_eq}
\end{equation}

The rate $r$ corresponds to the distance between the kernel weights. In this manner, a $3\times3$ convolution with rates = 1, 2 and 3 acquires the same receptive field size as $3\times3$, $5\times5$, and $7\times7$ regular convolutions respectively with the same number of parameters as a $3\times3$ regular convolution as shown in Figure~{\ref{atrousfig}}. Dilated convolution increases the receptive field size without incurring additional time or computational complexity.

\begin{figure}[h]
\begin{center}
    \subfloat[ASPP]{\includegraphics[width =           1.0\linewidth]{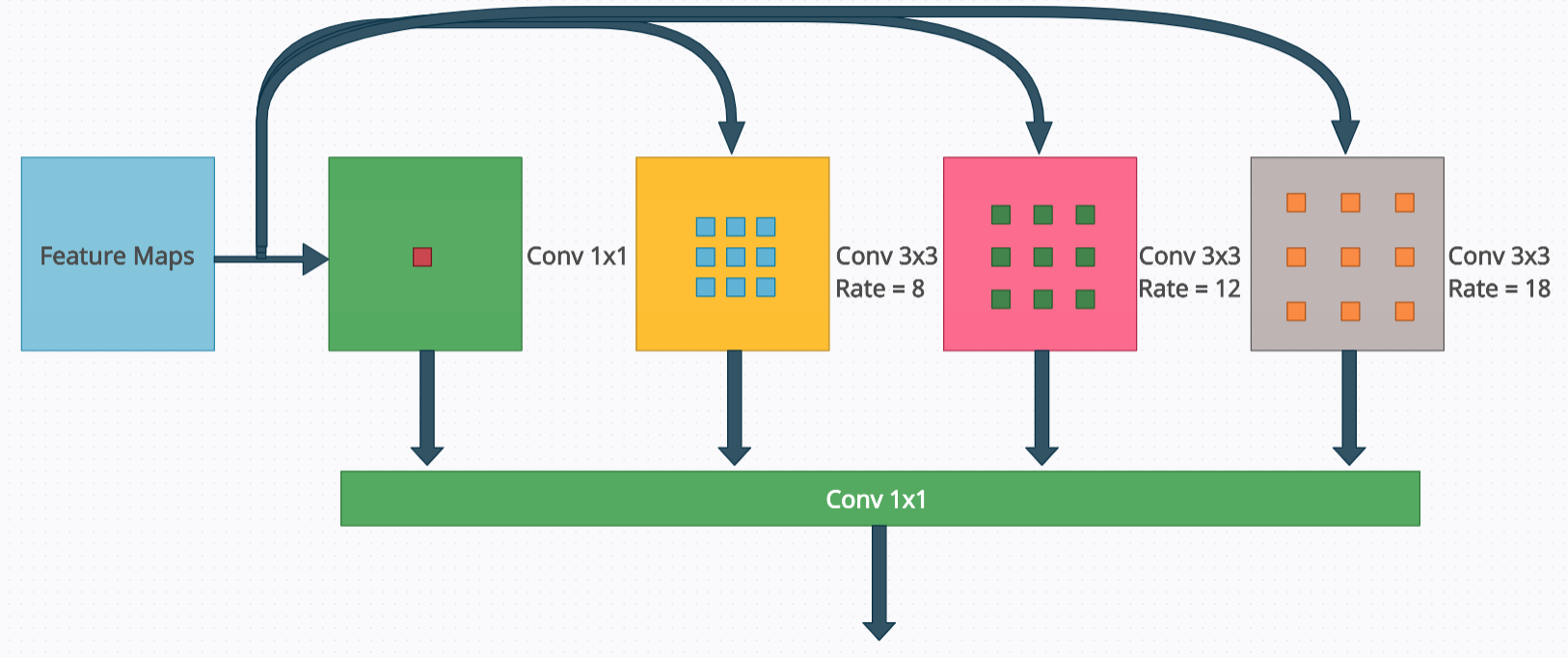}} \\
    \subfloat[Dense ASPP]{\includegraphics[width =           1.0\linewidth]{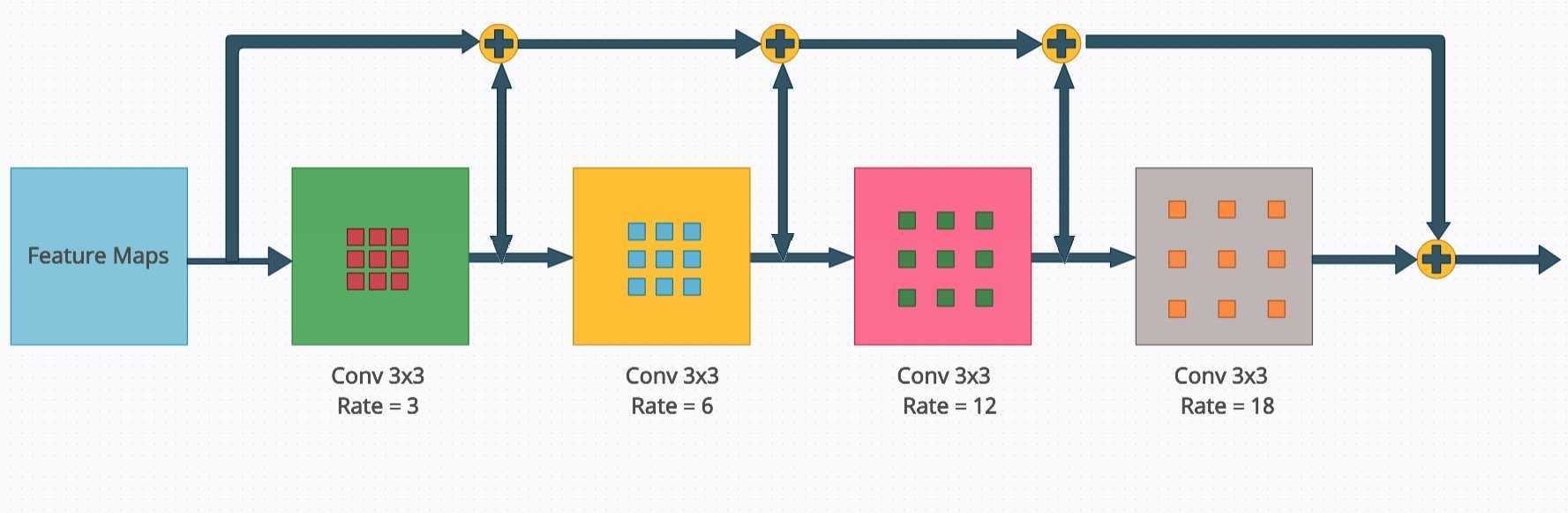}}

\caption{ASPP vs. Dense ASPP architectures.}
\label{asppfig}
\end{center}
\end{figure}

\begin{figure*}[h]
\begin{center}
\includegraphics[width = 1.0\linewidth]{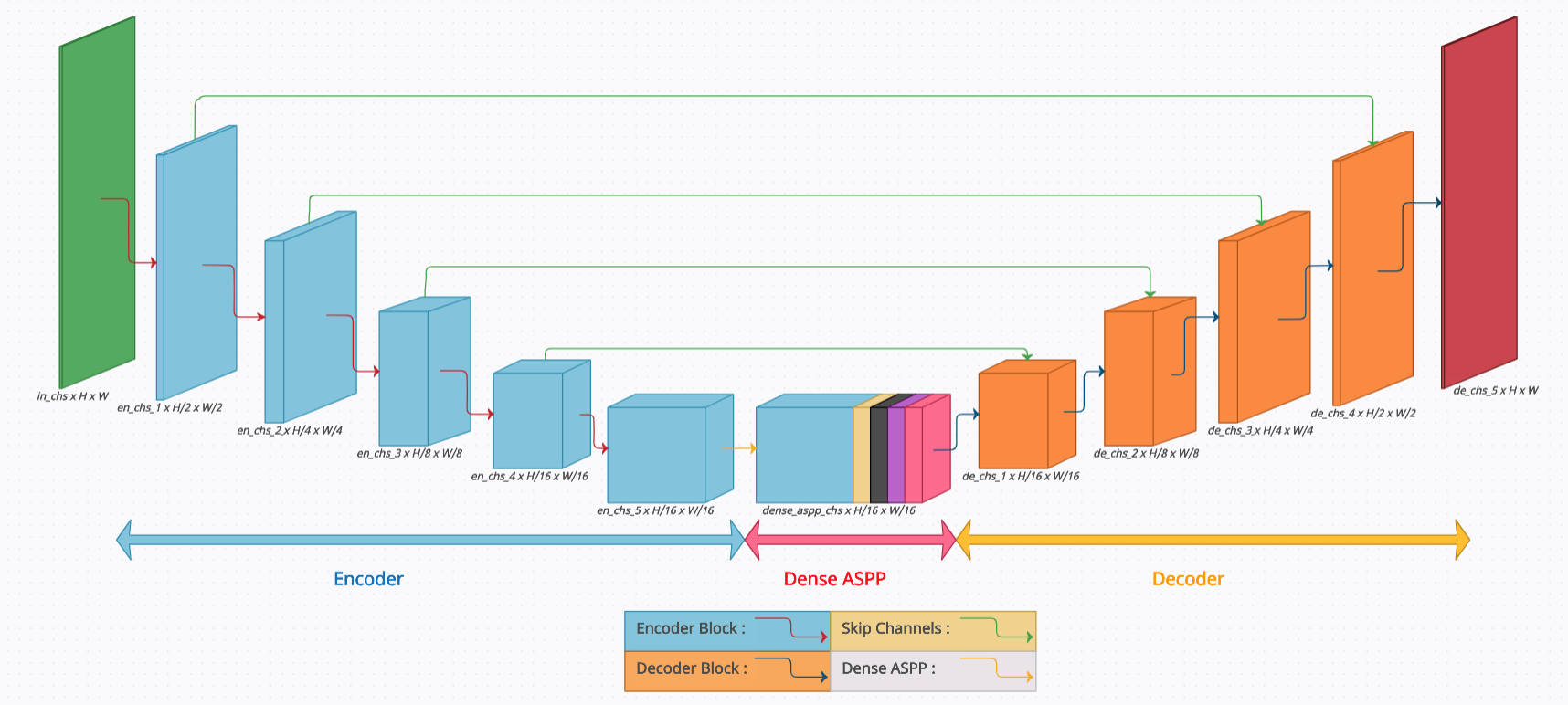}
\caption{Proposed Sci-Net Architecture.}
\label{modelfig}
\end{center}
\end{figure*}

\subsection{Atrous Spatial Pyramid Pooling (ASPP)}
ASPP is a module that applies multiple dilated convolutions with different rates on the feature maps to capture multi-scale representations. The concept has been first introduced in~\cite{deeplab} and then further developed in~\cite{deeplabv3,deeplabv3+}.

Typically one $1\times1$ convolution, three $3\times3$ dilated convolutions of atrous rates equal to (8, 12 and 18), and a global average pooling layer are applied in parallel. The resulting representations are then concatenated together and pooled with a $1\times1$ convolution as shown in Figure~{\ref{asppfig}} \textit{(a)}. Applying three different and separate dilated convolutions allows the model to extract spatial information at three different scales, with a maximum receptive field size equal to $37$ pixels.

\subsection{Dense ASPP}
Dense ASPP~\cite{dense_aspp_paper} applies dilated convolutions with increasing atrous rates in a cascade manner. The input to each dilated convolution block is the initially extracted feature maps concatenated with all the representation from previous dilated convolutions of lower rates as shown in Figure~\ref{asppfig} \textit{(b)}. Typically, four atrous convolutions are applied with rates equal to (3, 6, 12 and 18). When two convolutions of different receptive fields are stacked together, the resulting receptive field increases in a linear manner, as shown in Equation \ref{receptive_field_eq}:

\begin{equation}
R_{new} = R_1 + R_2 - 1.
\label{receptive_field_eq}
\end{equation}

where $R_1$ and $R_2$ denote receptive fields size in pixels of the \nth{1} and \nth{2} convolutional layers, and $R_{new}$ is the size of the new receptive field after stacking the two convolutional layers together.

Usage of Dense ASPP would lead to $16$ receptive field scales and a maximum value equal to $79$ pixels, which means that more pixels are involved in the convolution resulting in a denser feature pyramid than ASPP. Thus, based on the above analysis, Dense ASPP is integrated into the proposed Sci-Net model.

\subsection{Sci-Net Architecture}
\label{framework}
In this subsection, we provide a detailed description of the proposed Sci-Net model architecture and illustrate the role of the modifications that we apply to deliver ultimate performance. The proposed Sci-Net model shown in Figure~{\ref{modelfig}} adapts conventional UNet encoder-decoder architecture~\cite{unet} with the following modifications:

\begin{enumerate}[label=(\alph*)]
    \item Replacing the encoder with a more powerful yet light-weight feature extractor from the RegNet Family ($RegNetY-1.6GF$). RegNets have similar performance to their Efficient-Net counterparts while being $3$x to $5$x times faster.

    \item Integration of a Dense ASPP block to extract multi-scale representation from the features of the last encoder stage. The output features of Dense ASPP are the input to the first decoder block. We used the following rates ($3,6,12,18$) for the dilated blocks, and we set the output channels to $256$ for each block. When concatenated, the multi-scale representations alone are a total of $256 \times 4 = 1024$ channels, and the initial feature maps contain $888$ channels.

    \item Substitution of the last $3\times3$ convolution in the $5$-th encoder that has a kernel stride $= 2$, with a dilated convolution of a low atrous rate $= 2$ and kernel stride $= 1$ to avoid down-sampling of  stage $5$ features. And thus, the output stride becomes equal to $16$ instead of $32$. This modification preserves a sufficiently good spatial resolution at the Dense ASPP input.
    \item No up-sampling is applied at the first decoder block as both stages $4$ and $5$ feature maps have the exact same spatial resolution.
\end{enumerate}

Each decoder block comprises two 3x3 convolutions with a stride equal to 1 followed by a 2x up-sampling bilinear interpolation.

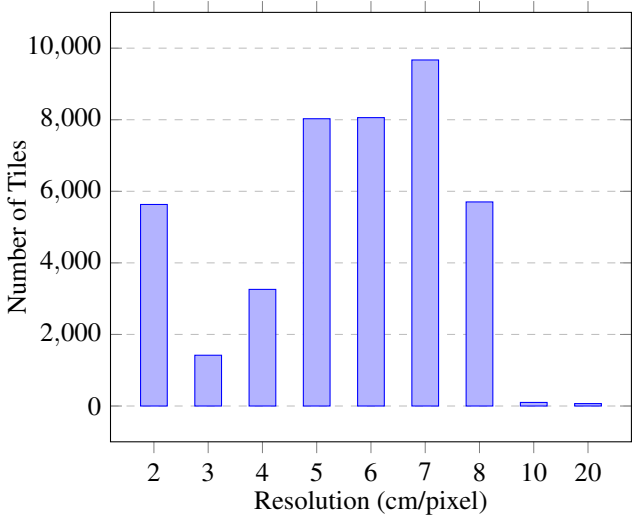
\begin{figure}[h]
    \begin{center}
        \begin{tikzpicture}
            \begin{axis}[
                        ybar ,
                        enlargelimits=0.1,
                        xlabel={Resolution (cm/pixel)},
                        ylabel={Number of Tiles},
                        symbolic x coords={2,3,4,5,6,7,8,10,20},
                        xtick=data,
                        ytick = {0,2000,4000,6000,8000,10000},
                        ymajorgrids=true,
                        ymin=0,
                        ymax=10000,
                        scaled y ticks = false,
                        grid style=dashed,
                        ]
            \addplot coordinates {(2,5632) (3,1419) (4,3259) (5,8028) (6,8059) (7,9670) (8,5703) (10,100) (20,68)};
            \end{axis}
        \end{tikzpicture}
    \end{center}
    \caption{Open Cities AI dataset tiles distribution per spatial resolution in cm/pixel.}
\label{stats_res}
\end{figure}

\section{Dataset \& Training}
\label{datasets}

In this section, we present both datasets used in this work, the training pipeline and implementation details, in addition to the evaluation metrics used.

\subsection{Datasets}
\label{datasetdescription}
As far as we know, Open Cities AI Challenge (OCAIC) and the Multi-Scale Building (MSB) datasets are the only suitable datasets to assess and evaluate the performance of the proposed Sci-Net model over images with multi-scale spatial resolution:

\paragraph{OCAIC \cite{open_cities}:}
Open Cities AI Challenge dataset is also known as Segmenting buildings for disaster resilience dataset. The majority of the data are collected
across different African cities, where the images and labels quality varies from one region to another. Open Cities AI dataset is split into two tiers.
For the scope of this work, tier 1 images are used, where imagery and labels are distributed under the {\bf CC-BY-4} and {\bf ODbL-1.0} licenses, respectively.
Tier 1 data is made of $31$ GeoTiff images of different spatial resolution and size. Resolution varies from very high (2cm/pixel) up to medium (20cm/pixel) resolutions as shown in Figure~{\ref{samples_res}}. Moreover, Figure~{\ref{stats_res}} shows images distribution across different resolutions.

The resulting dataset contains 40,000 tiles with their corresponding buildings' masks.

\paragraph{MSB \cite{msb_dataset}:}
Multi-Scale Building dataset is a relatively small dataset made of 2,900 images. The images are collected from different public dataset after undergoing horizontally/vertically flip augmentation and saturation transform. The training set is based on the Indiana dataset with 0.3m/pixel spatial resolution and the WHU satellite dataset I (global cities) dataset with spatial resolution varying from 0.3m/pixel to 2.5m/pixel. While the testset relies on the Inria dataset with 0.3m/pixel spatial resolution and the ISPRS Potsdam dataset with with 5cm/pixel spatial resolution. The resultant multisensor and multiscale MSB dataset covers 16 different cities with RGB images of 512 x 512 pixels size.

\subsection{Training details}
\label{train_pipeline}
For fair comparison, we ensured that all compared methods are using mostly the same hyperparameter choices. Specifically, the encoder is replaced by $RegNetY-1.6GF$ for all methods, except for HRNet which uses $HRNetV2-W32$ backbone. The decoder channels are fixed to (256, 128, 64, 32, 16) except for Deeplabv3+ where the number of channels is fixed to 256, and PSPNet where the number of output channels is 512 (number of filters in Spatial Pyramid). The number of channels for the PAB module in MANet is 64. All encoders are initialized with ImageNet pre-trained weights\cite{imagenet}.

\begin{table*}[h]
\begin{center}
\begin{tabular}{c|cccc|cc} \toprule
     {Model} & {micro-IoU} & {micro-F1} & {macro-IoU} & {macro-F1} & {Params. (M)} & {GFLOPs} \\ \midrule
     {PSPNet \cite{psp}} & {78.47} & {87.93} & {85.04} &{89.55} &{12.0} &{8.7} \\
     {DeepLabV3+ \cite{deeplabv3+}} & {79.83} & {88.78} & {86.28} &{90.50} &{12.0} &{13.5} \\
     {MANet \cite{manet}} & {80.08} & {88.93} & {86.20} &{90.38} &{35.5} &{25.0} \\
     {HRNet \cite{hrnet}} & {80.30} & {89.07} & {82.85} &{90.46} &{29.5} &{45.0} \\
     {\bf Sci-Net} & {\bf 82.25} & {\bf 91.04} & {\bf 88.42} &{\bf 92.62} &{24.2} &{33.1} \\
     \bottomrule

\end{tabular}
\end{center}
\caption{Performance metrics of Sci-Net architecture compared to existing SoA models on the OCAIC testset where Sci-Net outperforms all benchmarked models in terms of IoU and F1 scores.}
\label{results1}
\end{table*}

In all our experiments, models are trained until convergence using Adam optimizer~\cite{adam} and polynomial learning rate policy~\cite{psp} where the learning rate is decayed from the initial value of $0.0001$ till zero following Equation \ref{learning_rate_eq}:

\begin{equation}
    lr_{t+1} = lr_t * (1 - \frac{epoch_{t+1}}{epoch_{max}})^{0.9}
\label{learning_rate_eq}
\end{equation}

A weighted combination of Dice loss and Binary Cross-Entropy ($BCE$) loss is used as defined in Equation \ref{loss_func_eq}:

\begin{equation}
  \text{Loss} = \Gamma_{1} \cdot BCE + \Gamma_{2} \cdot Dice
\label{loss_func_eq}
\end{equation}

where $\Gamma_{1}=\Gamma_{2}=0.5$ is considered for simplicity.\\

During training on the OCAIC dataset, we use a batch-size equals to $12$ and randomly crop $512\times512$ chips of the original $1024\times1024$ tiles. Then, we apply, with an $80\%$ probability, positional augmentations such as horizontal and vertical flipping, in addition to \ang{180} rotation. These augmentations help to introduce some randomness at every training iteration and prevent the model from over-fitting.

Moreover, the training is performed over a Titan-XP GPU card with $12$ GB of VRAM. In addition to the proposed Sci-Net model, we trained some well-known SoA models for comparison. Training these models on the OCAIC dataset took between $20$ and $120$ hours, depending on the model complexity. Training framework is done in PyTorch using mixed precision functionality.

\begin{figure}[ht]
\begin{center}
\scalebox{0.5}{%
\begin{subfigure}[b]{\columnwidth}

\begin{tikzpicture}
\begin{axis}[
    xlabel={resolution(cm/pixel)},
    ylabel={micro-IoU Score},
    ymin=55, ymax=95,
    symbolic x coords={2,3,4,5,6,7,8,10,20},
    xtick=data,
    legend pos=north east,
    ymajorgrids=true,
    grid style=dashed,
]

\addplot[
    color=blue,
    mark=*,
    dotted
    ]
    coordinates {
    (2,85.621)(3,89.023)(4,86.032)(5,74.572)(6,79.731)(7,76.141)(8,83.505)(10,75.774)(20,75.152)
    };

\addplot[
    color=red,
    mark=*,
    dotted
    ]
    coordinates {
    (2,86.294)(3,89.09)(4,86.169)(5,74.962)(6,79.986)(7,75.943)(8,83.381)(10,80.392)(20,73.736)
    };
\addplot[
    color=green,
    mark=*,
    dotted
    ]
    coordinates {
    (2,87.441)(3,90.956)(4,87.281)(5,78.045)(6,81.691)(7,79.292)(8,86.324)(10,85.2)(20,76.478)
    };
\addplot[
    color=violet,
    mark=*,
    dotted
    ]
    coordinates {
    (2,85.708)(3,89.468)(4,86.328)(5,74.307)(6,78.914)(7,75.439)(8,83.58)(10,79.968)(20,75.419)
    };
\addplot[
    color=cyan,
    mark=*,
    dotted
    ]
    coordinates {
    (2,85.992)(3,88.841)(4,86.061)(5,74.809)(6,79.755)(7,75.895)(8,82.856)(10,75.369)(20,73.738)
    };
\addplot[
    color=magenta,
    mark=*,
    dotted
    ]
    coordinates {
    (2,84.425)(3,88.676)(4,85.881)(5,74.67)(6,79.746)(7,75.74)(8,83.014)(10,81.65)(20,71.04)
    };
\addplot[
    color=brown,
    mark=*,
    dotted
    ]
    coordinates {
    (2,84.838)(3,86.996)(4,84.313)(5,73.35)(6,68.816)(7,73.483)(8,79.465)(10,69.974)(20,61.248)
    };
\end{axis}
\end{tikzpicture}

\end{subfigure}
}%
\scalebox{0.5}{%
\begin{subfigure}[b]{\columnwidth}

\begin{tikzpicture}
\begin{axis}[
    xlabel={resolution(cm/pixel)},
    ylabel={micro-F1 Score},
    ymin=65, ymax=100,
    symbolic x coords={2,3,4,5,6,7,8,10,20},
    xtick=data,
    legend pos=north east,
    ymajorgrids=true,
    grid style=dashed,
]

\addplot[
    color=blue,
    mark=*,
    dotted
    ]
    coordinates {
    (2,92.253)(3,94.193)(4,92.492)(5,85.434)(6,88.723)(7,86.455)(8,91.011)(10,86.217)(20,85.813)
    };

\addplot[
    color=red,
    mark=*,
    dotted
    ]
    coordinates {
    (2,92.643)(3,94.23)(4,92.571)(5,85.689)(6,88.88)(7,86.327)(8,90.937)(10,89.131)(20,84.883)
    };
\addplot[
    color=green,
    mark=*,
    dotted
    ]
    coordinates {
    (2,94.149)(3,96.155)(4,95.056)(5,87.744)(6,90.698)(7,88.552)(8,92.903)(10,91.133)(20,87.372)
    };
\addplot[
    color=violet,
    mark=*,
    dotted
    ]
    coordinates {
    (2,92.304)(3,94.441)(4,92.662)(5,85.26)(6,88.214)(7,86.0)(8,91.056)(10,88.869)(20,85.987)
    };
\addplot[
    color=cyan,
    mark=*,
    dotted
    ]
    coordinates {
    (2,92.469)(3,94.091)(4,92.508)(5,85.589)(6,88.737)(7,86.296)(8,90.624)(10,85.955)(20,84.884)
    };
\addplot[
    color=magenta,
    mark=*,
    dotted
    ]
    coordinates {
    (2,91.555)(3,93.998)(4,92.404)(5,85.498)(6,88.732)(7,86.195)(8,90.719)(10,89.898)(20,83.069)
    };
\addplot[
    color=brown,
    mark=*,
    dotted
    ]
    coordinates {
    (2,91.797)(3,93.046)(4,91.489)(5,84.627)(6,81.527)(7,84.715)(8,88.558)(10,82.335)(20,75.968)
    };
\end{axis}
\end{tikzpicture}

\end{subfigure}
}

\scalebox{0.5}{%
\begin{subfigure}[b]{\columnwidth}

\begin{tikzpicture}
\begin{axis}[
    xlabel={resolution(cm/pixel)},
    ylabel={macro-IoU Score},
    ymin=60, ymax=100,
    symbolic x coords={2,3,4,5,6,7,8,10,20},
    xtick=data,
    legend pos=north east,
    ymajorgrids=true,
    grid style=dashed,
]

\addplot[
    color=blue,
    mark=*,
    dotted
    ]
    coordinates {
    (2,82.953)(3,87.043)(4,83.043)(5,71.054)(6,95.153)(7,90.285)(8,91.602)(10,77.974)(20,78.397)
    };

\addplot[
    color=red,
    mark=*,
    dotted
    ]
    coordinates {
    (2,84.074)(3,86.618)(4,84.279)(5,71.768)(6,95.645)(7,90.537)(8,92.451)(10,79.527)(20,77.2)
    };
\addplot[
    color=green,
    mark=*,
    dotted
    ]
    coordinates {
    (2,85.148)(3,88.427)(4,84.88)(5,74.864)(6,97.407)(7,92.654)(8,94.29)(10,90.867)(20,79.851)
    };
\addplot[
    color=violet,
    mark=*,
    dotted
    ]
    coordinates {
    (2,82.652)(3,87.763)(4,84.406)(5,70.959)(6,95.85)(7,89.976)(8,91.932)(10,80.618)(20,78.582)
    };
\addplot[
    color=cyan,
    mark=*,
    dotted
    ]
    coordinates {
    (2,82.852)(3,87.851)(4,83.765)(5,71.563)(6,96.056)(7,90.246)(8,91.27)(10,77.933)(20,77.071)
    };
\addplot[
    color=magenta,
    mark=*,
    dotted
    ]
    coordinates {
    (2,82.631)(3,87.388)(4,82.602)(5,71.332)(6,95.515)(7,90.687)(8,92.35)(10,87.437)(20,74.811)
    };
\addplot[
    color=brown,
    mark=*,
    dotted
    ]
    coordinates {
    (2,81.671)(3,84.283)(4,81.975)(5,69.833)(6,94.673)(7,89.487)(8,90.934)(10,74.709)(20,66.325)
    };
\end{axis}
\end{tikzpicture}

\end{subfigure}
}%
\scalebox{0.5}{%
\begin{subfigure}[b]{\columnwidth}

\begin{tikzpicture}
\begin{axis}[
    xlabel={resolution(cm/pixel)},
    ylabel={macro-F1 Score},
    ymin=65, ymax=100,
    symbolic x coords={2,3,4,5,6,7,8,10,20},
    xtick=data,
    legend pos=north east,
    ymajorgrids=true,
    grid style=dashed,
]

\addplot[
    color=blue,
    mark=*,
    dotted
    ]
    coordinates {
    (2,88.014)(3,92.157)(4,88.413)(5,80.142)(6,96.139)(7,93.132)(8,93.602)(10,82.13)(20,87.579)
    };

\addplot[
    color=red,
    mark=*,
    dotted
    ]
    coordinates {
    (2,88.962)(3,91.814)(4,89.644)(5,80.869)(6,96.615)(7,93.388)(8,94.5)(10,83.138)(20,86.801)
    };
\addplot[
    color=green,
    mark=*,
    dotted
    ]
    coordinates {
    (2,90.284)(3,94.658)(4,90.549)(5,83.088)(6,96.272)(7,95.484)(8,96.331)(10,94.123)(20,89.229)
    };
\addplot[
    color=violet,
    mark=*,
    dotted
    ]
    coordinates {
    (2,87.694)(3,92.859)(4,89.867)(5,80.279)(6,96.81)(7,92.789)(8,93.911)(10,85.709)(20,87.698)
    };
\addplot[
    color=cyan,
    mark=*,
    dotted
    ]
    coordinates {
    (2,87.839)(3,93.047)(4,89.437)(5,80.764)(6,97.028)(7,93.043)(8,93.196)(10,82.128)(20,86.69)
    };
\addplot[
    color=magenta,
    mark=*,
    dotted
    ]
    coordinates {
    (2,87.733)(3,92.708)(4,88.038)(5,80.563)(6,96.448)(7,93.606)(8,94.496)(10,92.913)(20,85.177)
    };
\addplot[
    color=brown,
    mark=*,
    dotted
    ]
    coordinates {
    (2,86.97)(3,90.296)(4,88.262)(5,79.463)(6,95.804)(7,92.56)(8,93.191)(10,80.103)(20,78.988)
    };
\end{axis}
\end{tikzpicture}

\end{subfigure}
}
\scalebox{0.5}{%
\begin{subfigure}[b]{\columnwidth}
\center
\frame{\includegraphics[width = 2.5in, height = 2in]{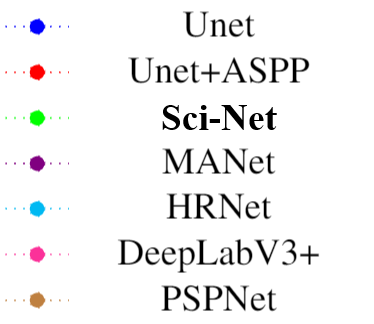}}
\end{subfigure}
}
\end{center}
\caption{Performance comparison of Sci-Net (in green) versus SoA models in terms of micro- and macro- IoU and F1 scores using the OCAIC testset.}
\label{resultsgraphs}
\end{figure}
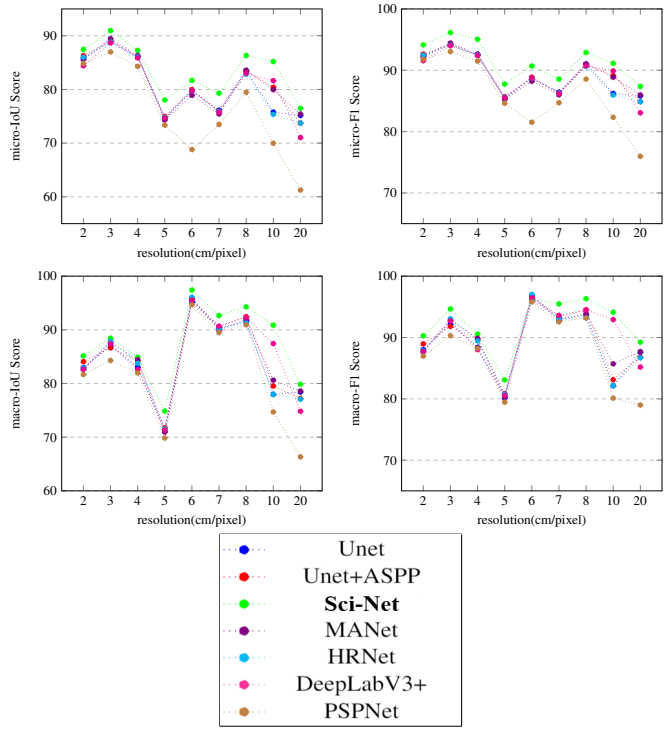

\subsection{Metrics}
Performance evaluation of the proposed Sci-Net and SoA models is measured using macro, and micro Intersection over Union (IoU) and F1-score defined in Equations \ref{iou_eq} and \ref{f1score_eq}:

\begin{equation}
IoU  = \frac{ TP + \epsilon}{ TP + FP + FN + \epsilon}
\label{iou_eq}
\end{equation}

\begin{equation}
F1-score = \frac{(1+\beta^2) \cdot TP + \epsilon}{(1+\beta^2) \cdot TP + \beta{^2} \cdot FN + FP + \epsilon}
\label{f1score_eq}
\end{equation}

where $\beta = 1$ and $\epsilon=0.0001$. $TP$, $FP$, and $FN$ are True Positive, False Positive, and False Negative, respectively.

The employed metrics are used according to the following two types of averaging across the whole test dataset:

\begin{enumerate}
    \item  \textbf{Macro} scores are calculated per prediction according to $TP$, $FP$, and $FN$ for each prediction mask and then averaged afterward.
    \item  \textbf{Micro} scores are calculated using the total number of $TP$, $FP$, and $FN$ across all prediction masks, and then the final score is computed accordingly.
\end{enumerate}

Macro scoring helps in assessing the average performance per image, while micro scoring assesses the overall performance. Correctly classified images with blank ground truth masks (True Negative) provide a boost in macro scoring. However, such images do not affect micro scoring, as it treats the whole dataset as having one large ground truth mask.

It is worth noting that simulation results are validated every epoch, and the model's weights that maximize micro-IoU score over the validation set are saved.

\begin{figure*}[h]
\setlength\tabcolsep{0pt}
\begin{center}
\begin{tabular}{@{{\makebox[2.5em][r]{\rownumber\space}}} | ccccccccc}
 \textbf{Image} &
 \textbf{GT} &
 \textbf{PSPNet} &
 \textbf{DeepLabV3+} &
 \textbf{UNet} &
 \textbf{MANet} &
 \textbf{HRNet} &
 \textbf{UNet+ASPP} &
 \textbf{Sci-Net}

 \gdef\rownumber{\stepcounter{magicrownumbers}\arabic{magicrownumbers}}\\

 \includegraphics[width=0.7in]{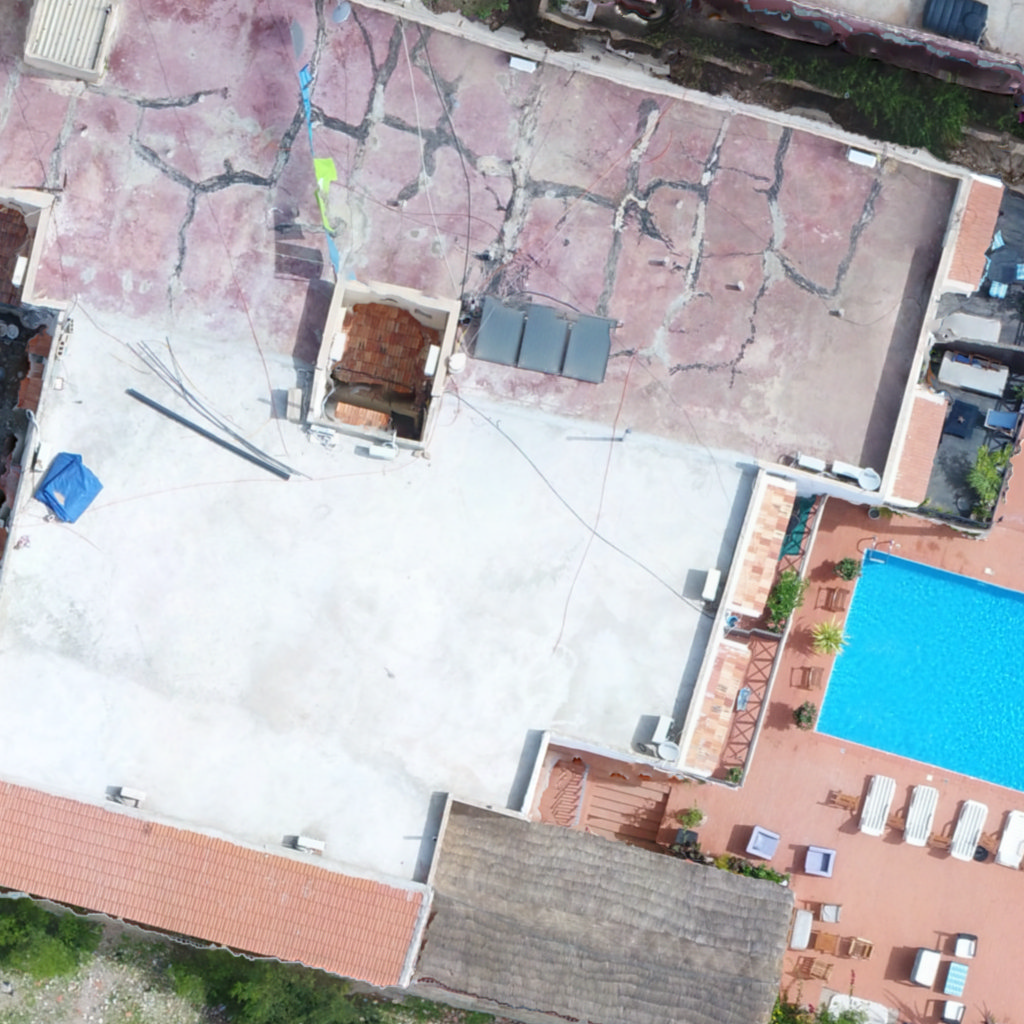} &
 \includegraphics[width=0.7in]{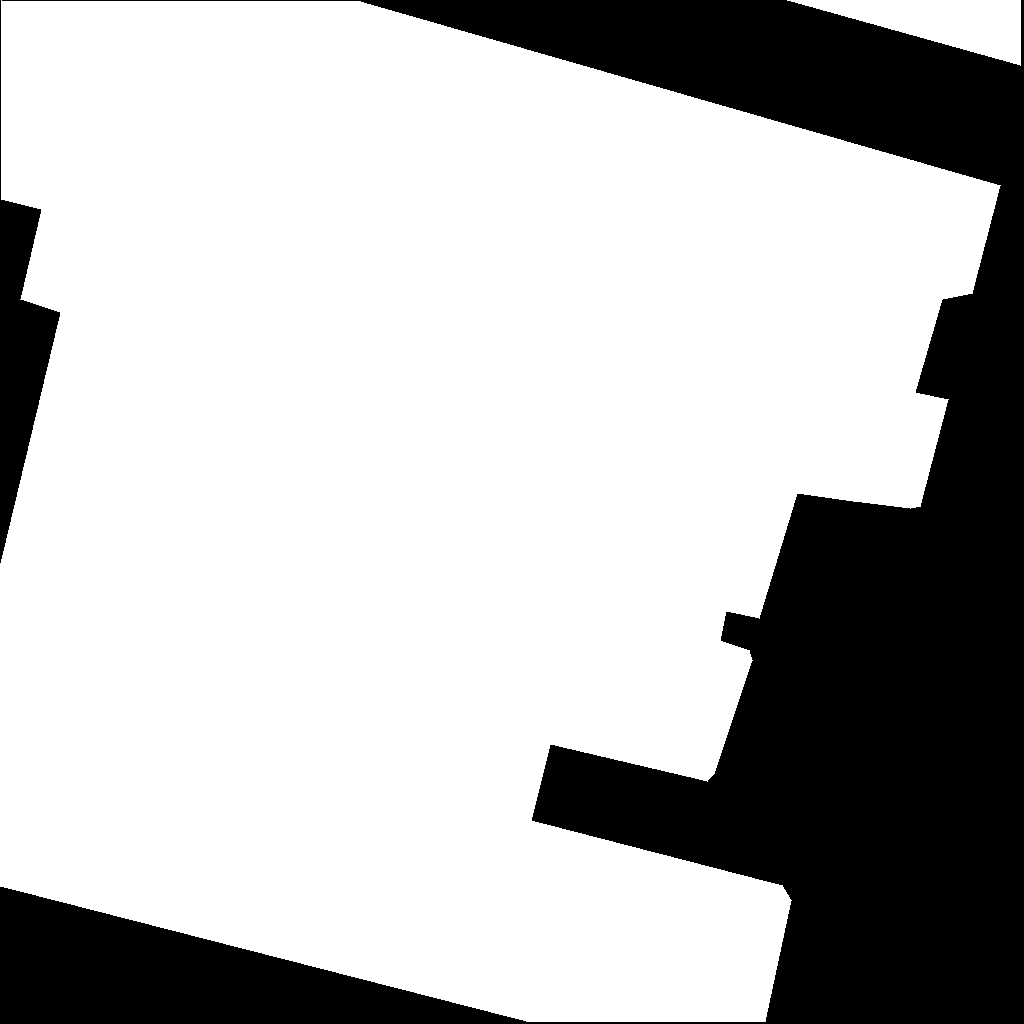} &
 \includegraphics[width=0.7in]{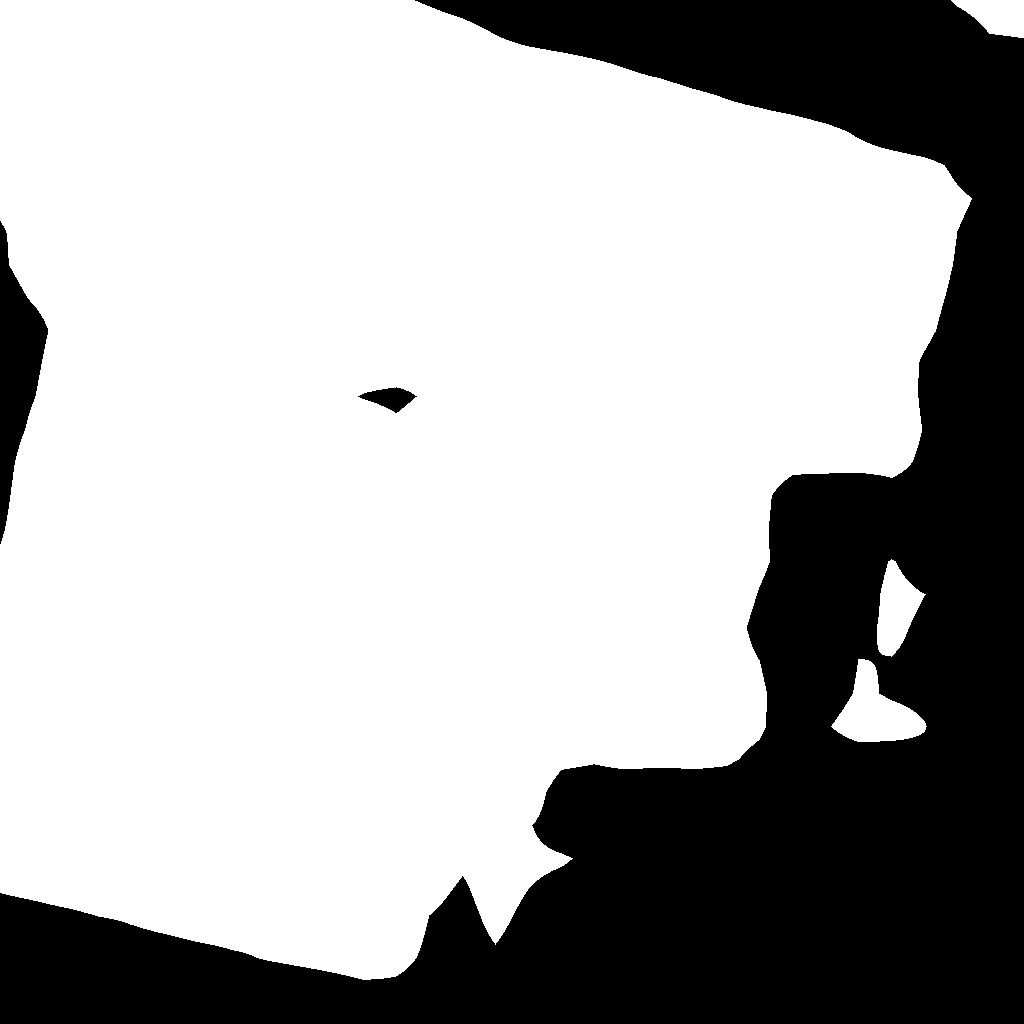} &
 \includegraphics[width=0.7in]{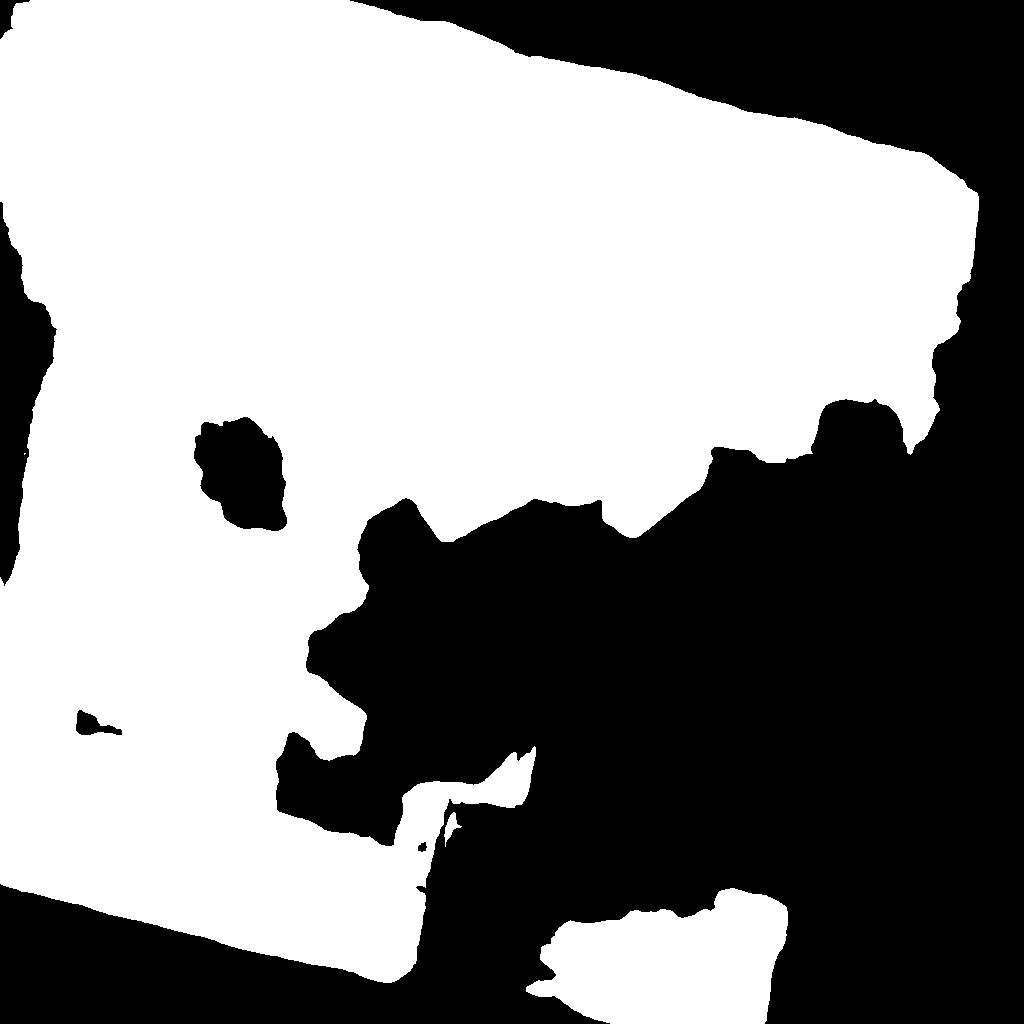} &
 \includegraphics[width=0.7in]{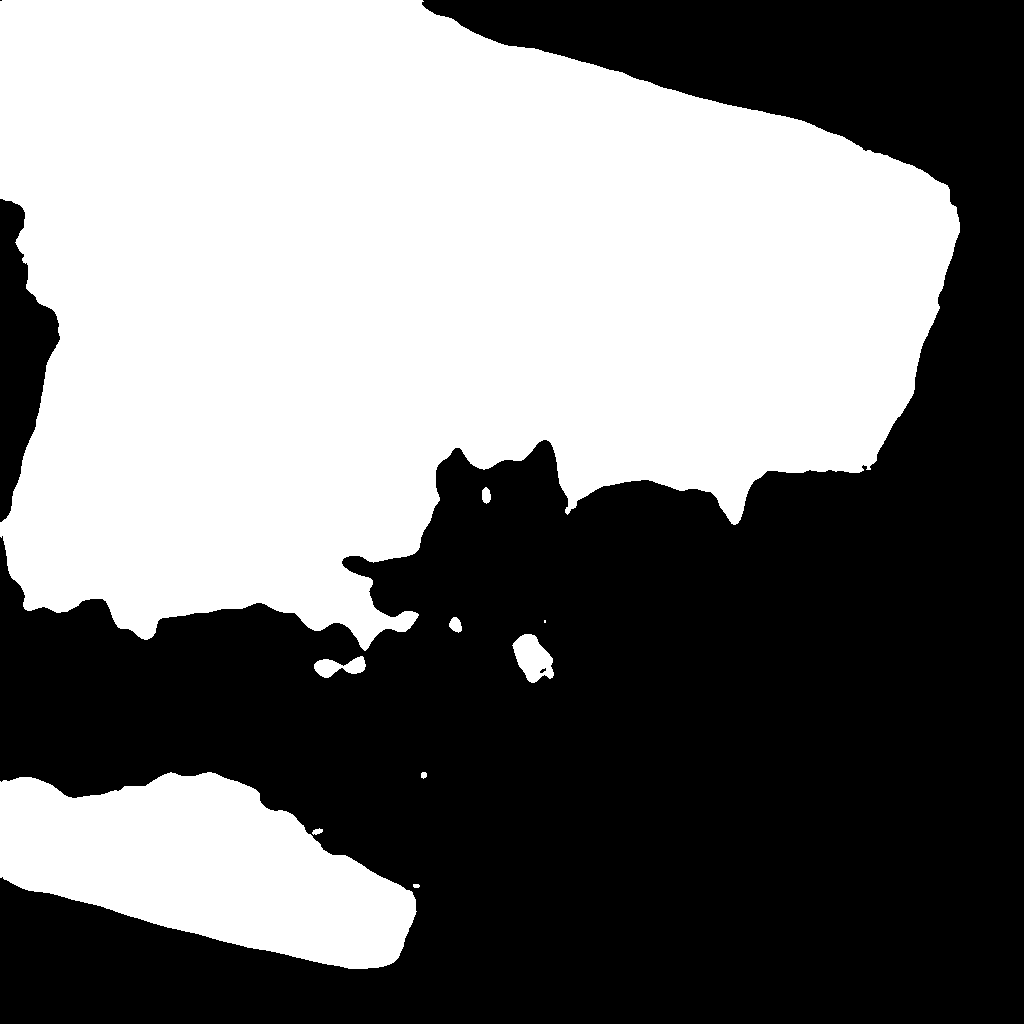} &
 \includegraphics[width=0.7in]{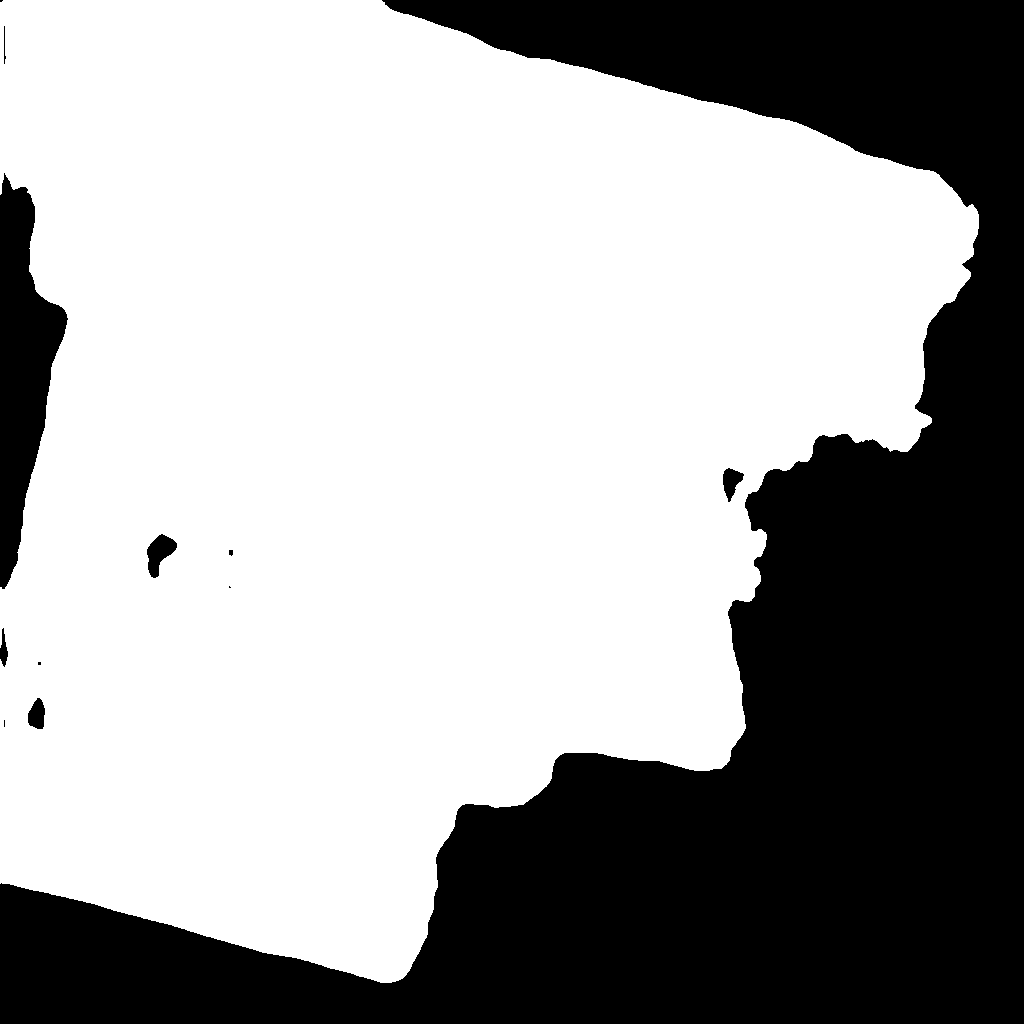} &
 \includegraphics[width=0.7in]{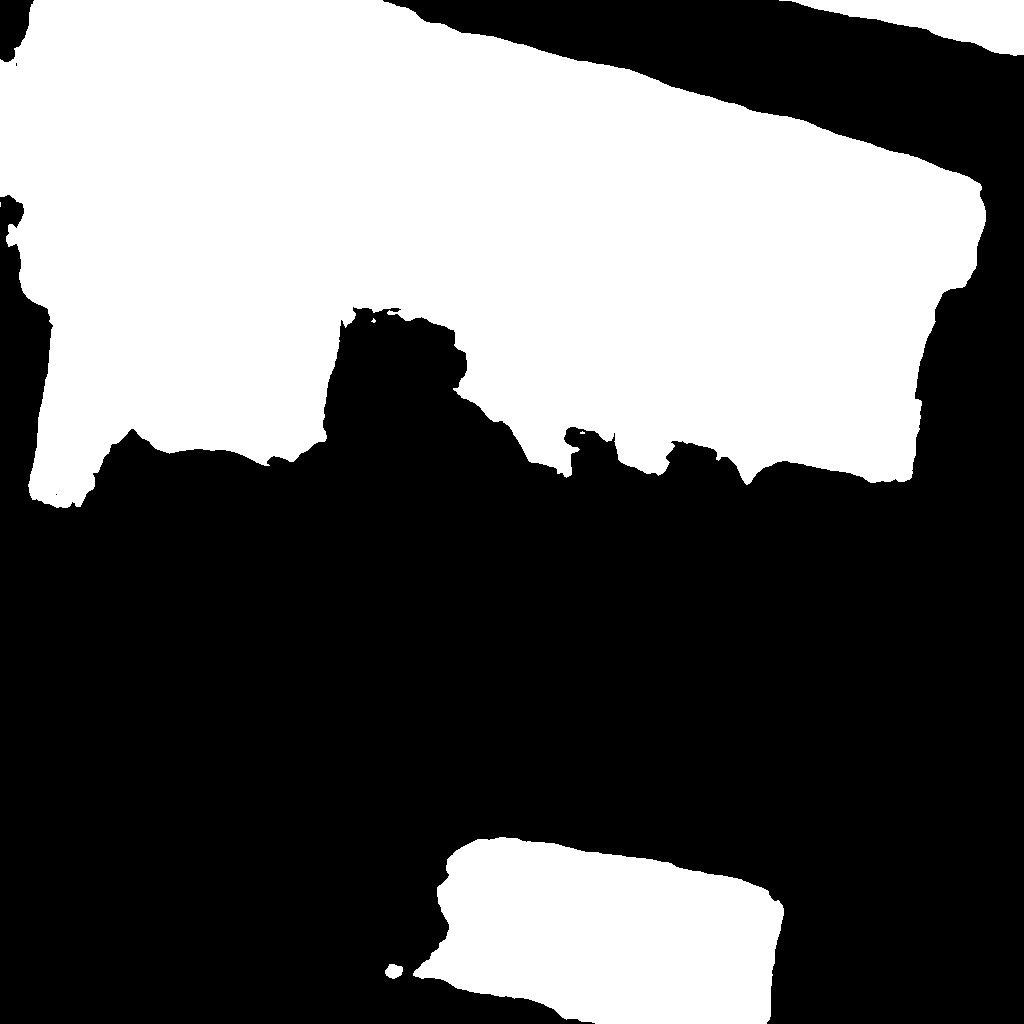} &
 \includegraphics[width=0.7in]{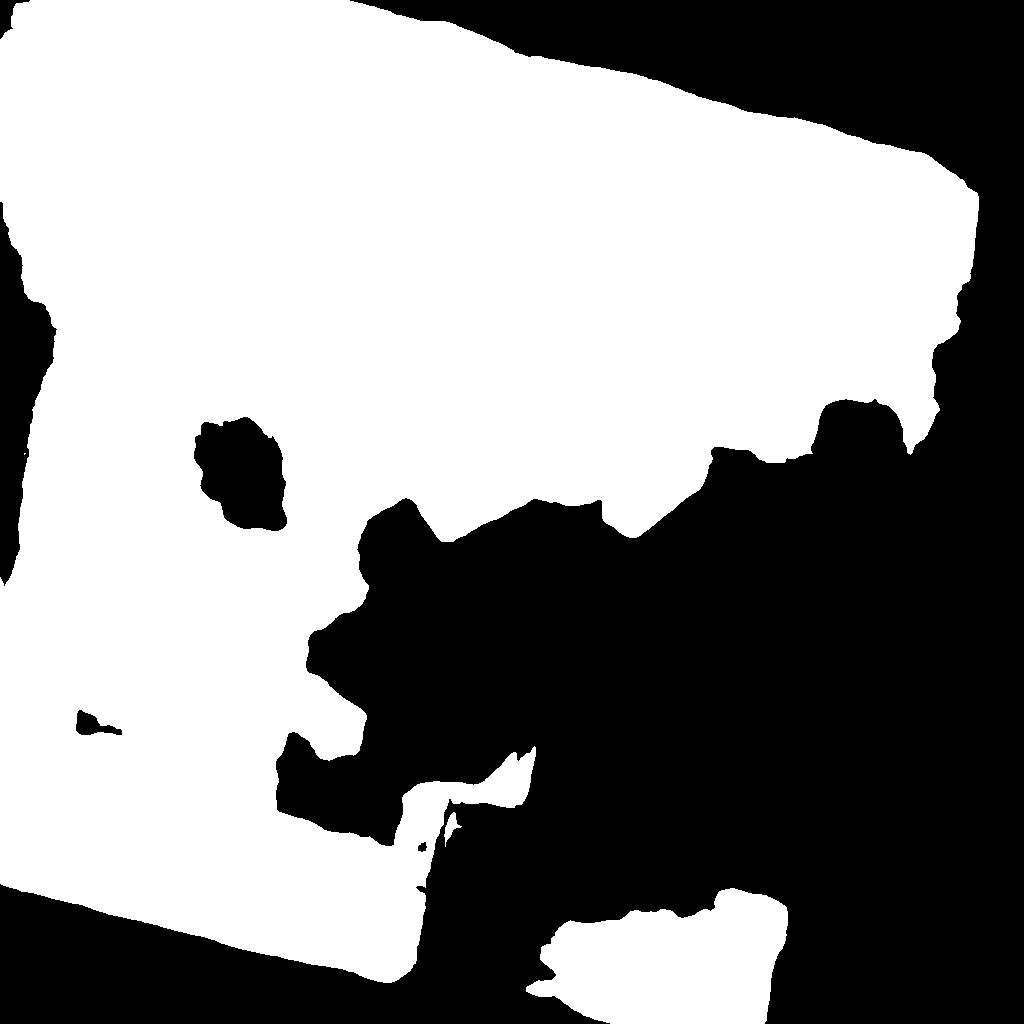} &
 \includegraphics[width=0.7in]{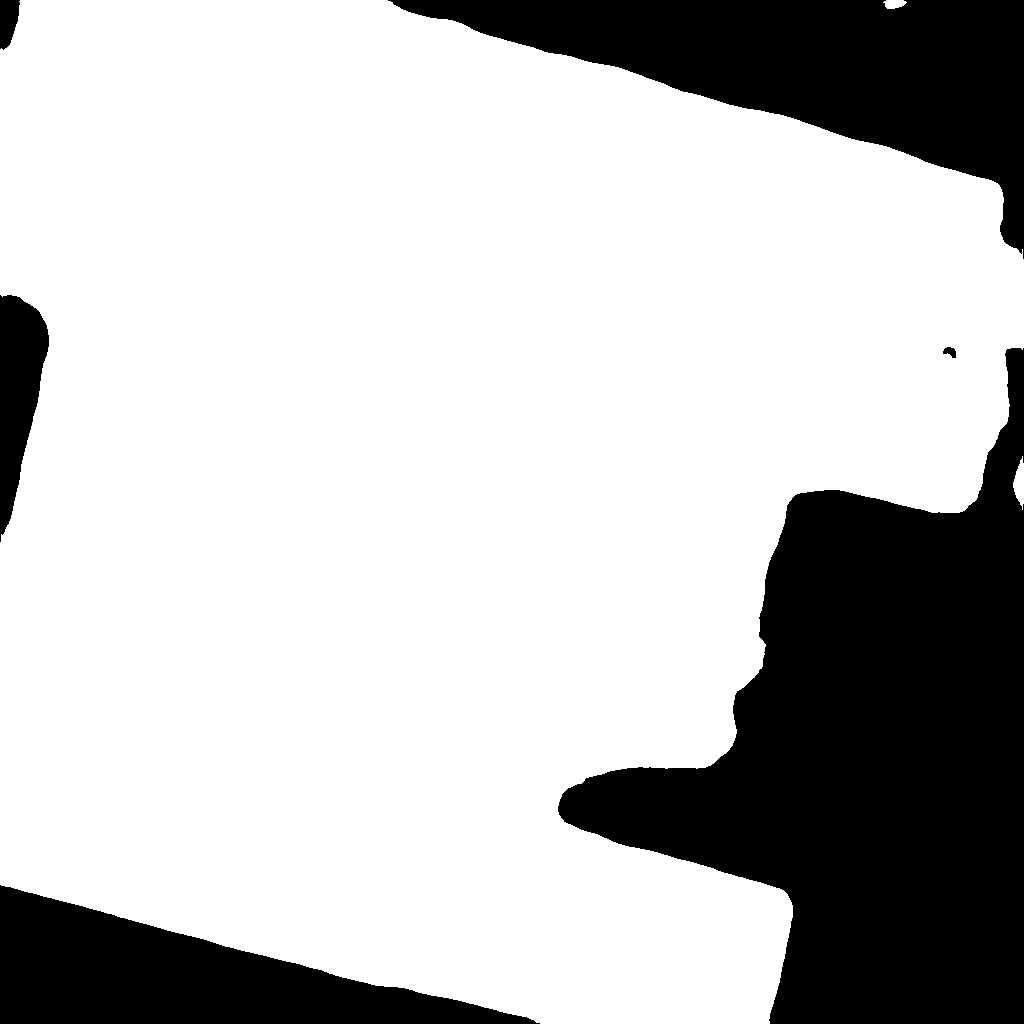} \\

 \includegraphics[width=0.7in]{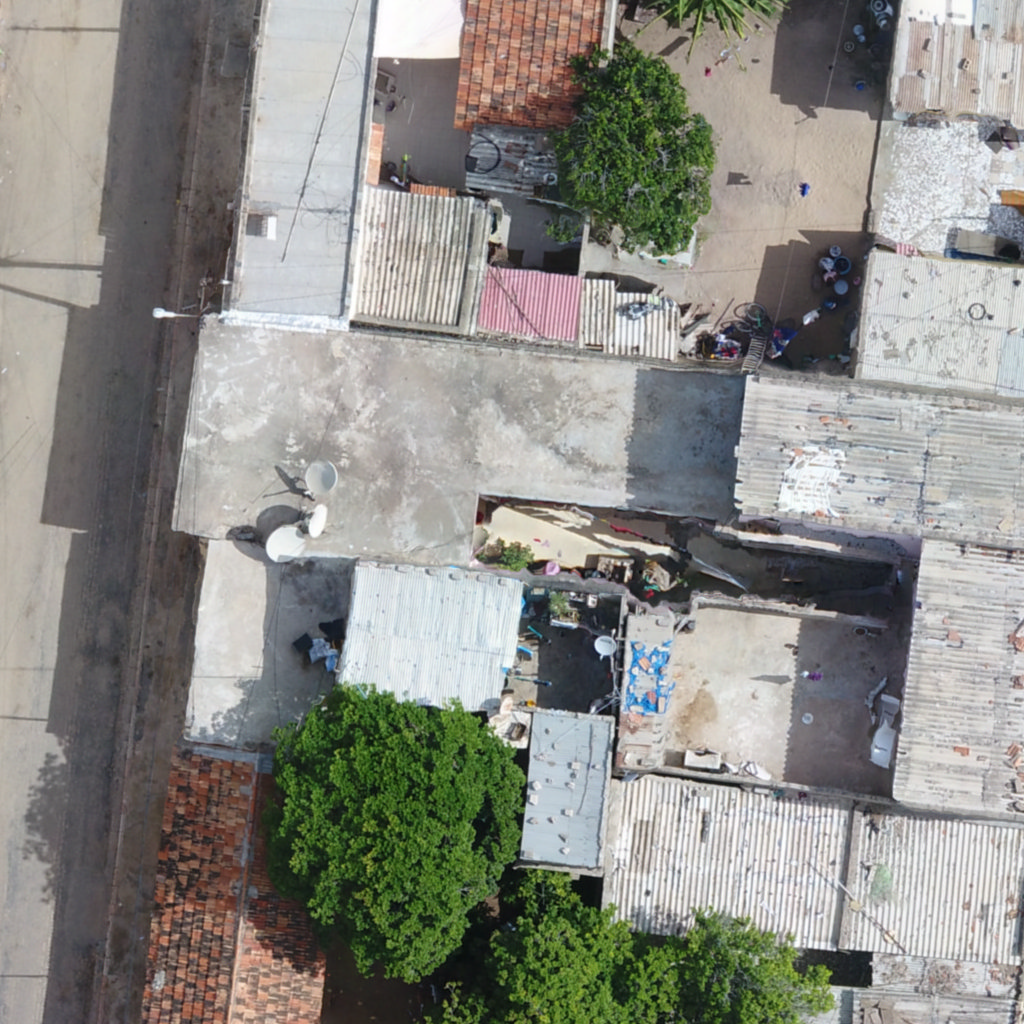} &
 \includegraphics[width=0.7in]{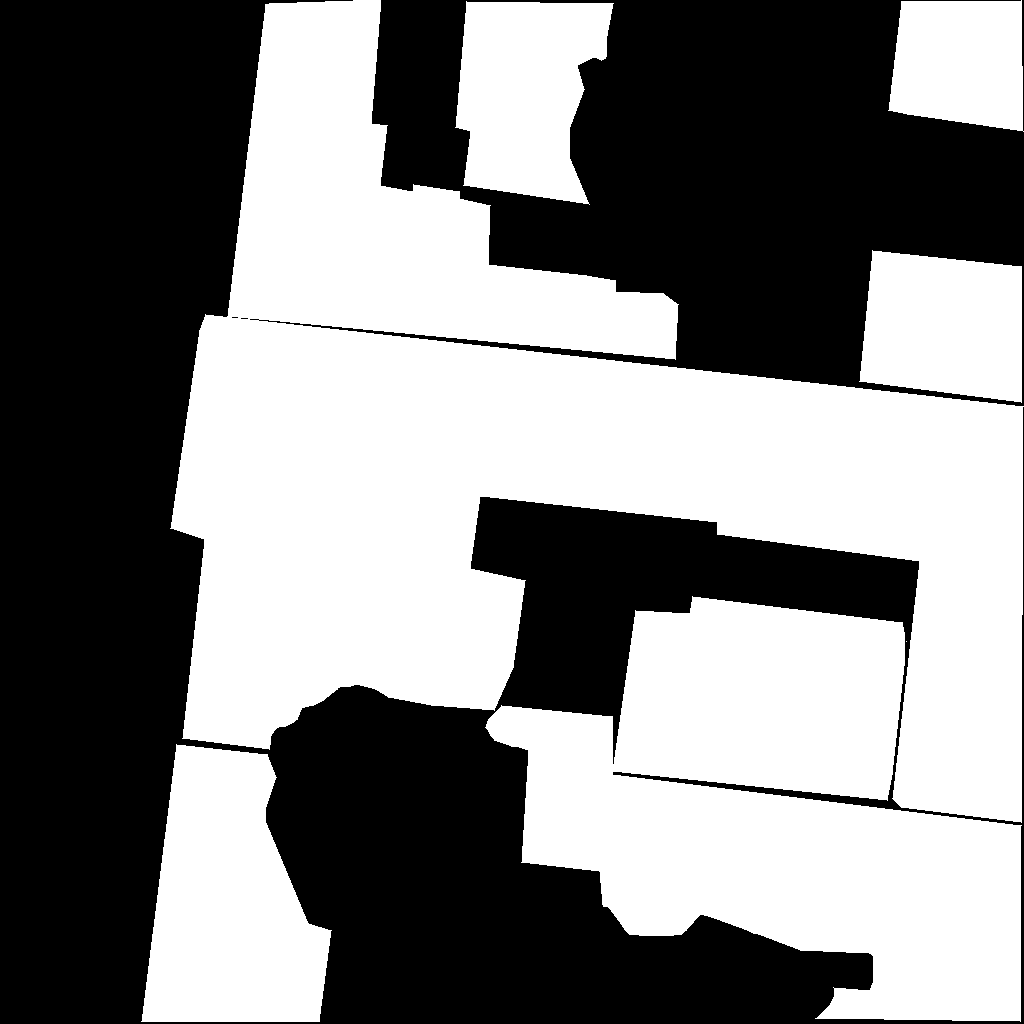} &
 \includegraphics[width=0.7in]{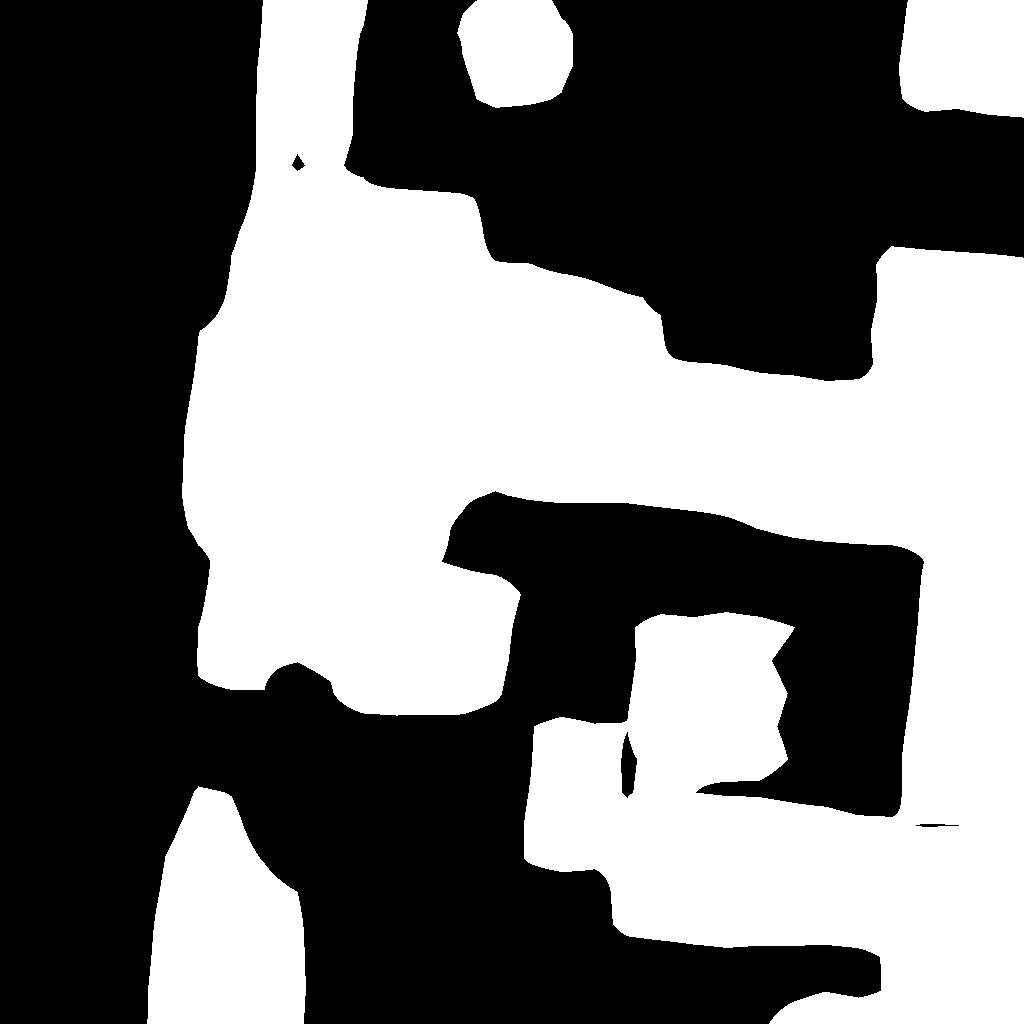} &
 \includegraphics[width=0.7in]{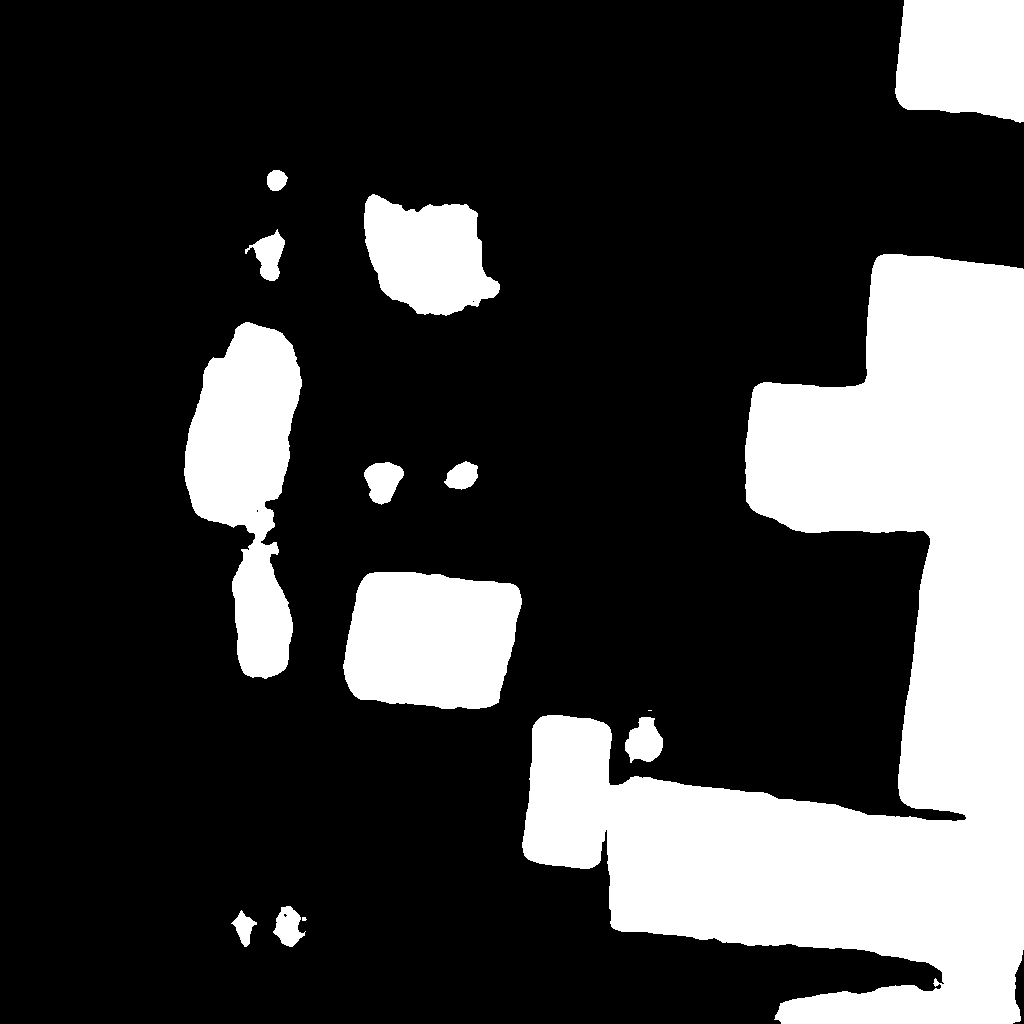} &
 \includegraphics[width=0.7in]{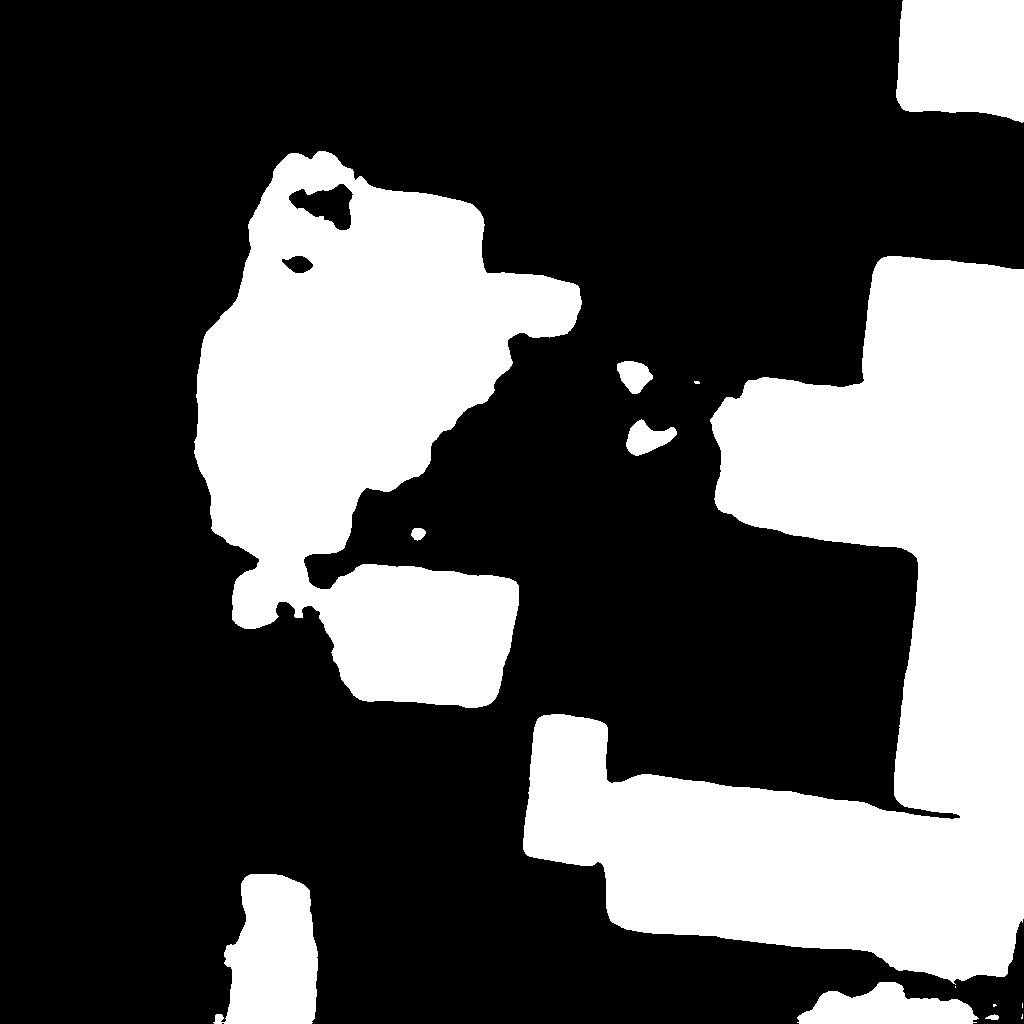} &
 \includegraphics[width=0.7in]{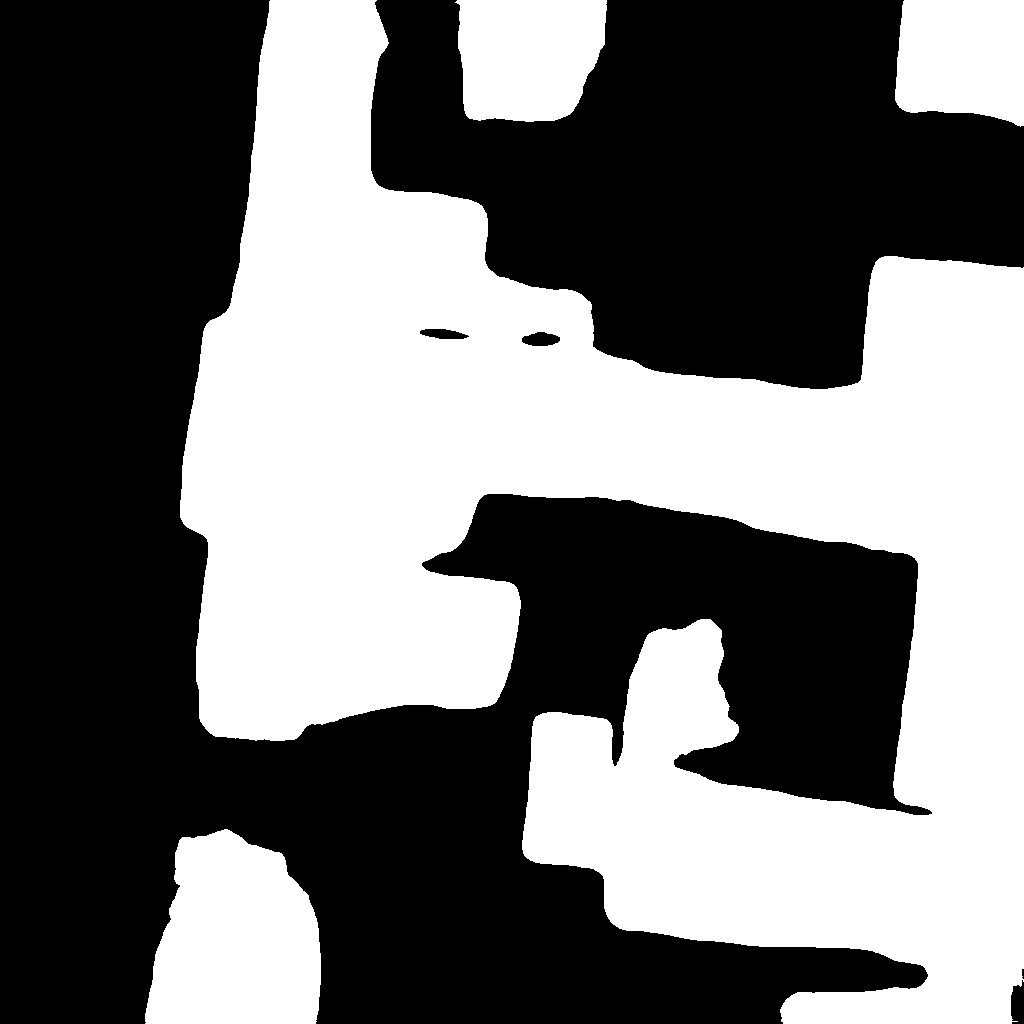} &
 \includegraphics[width=0.7in]{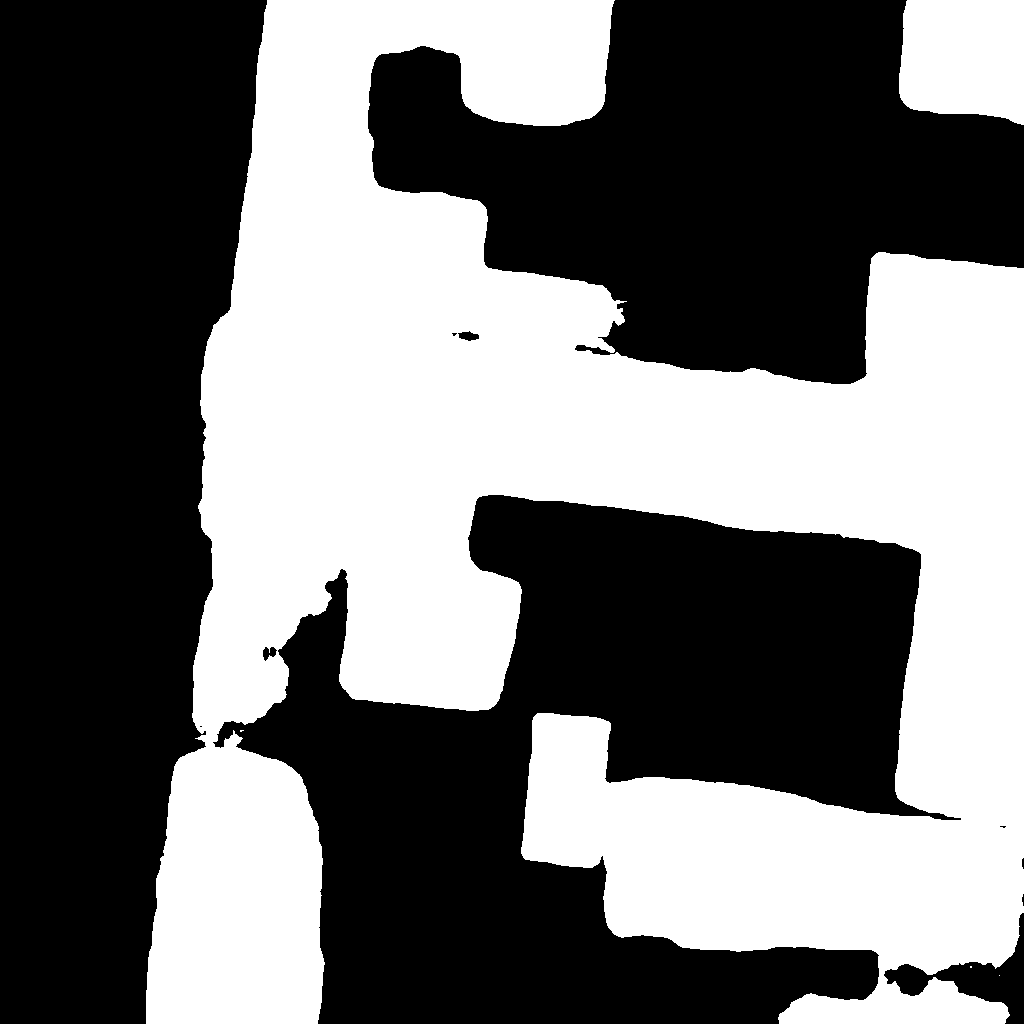} &
 \includegraphics[width=0.7in]{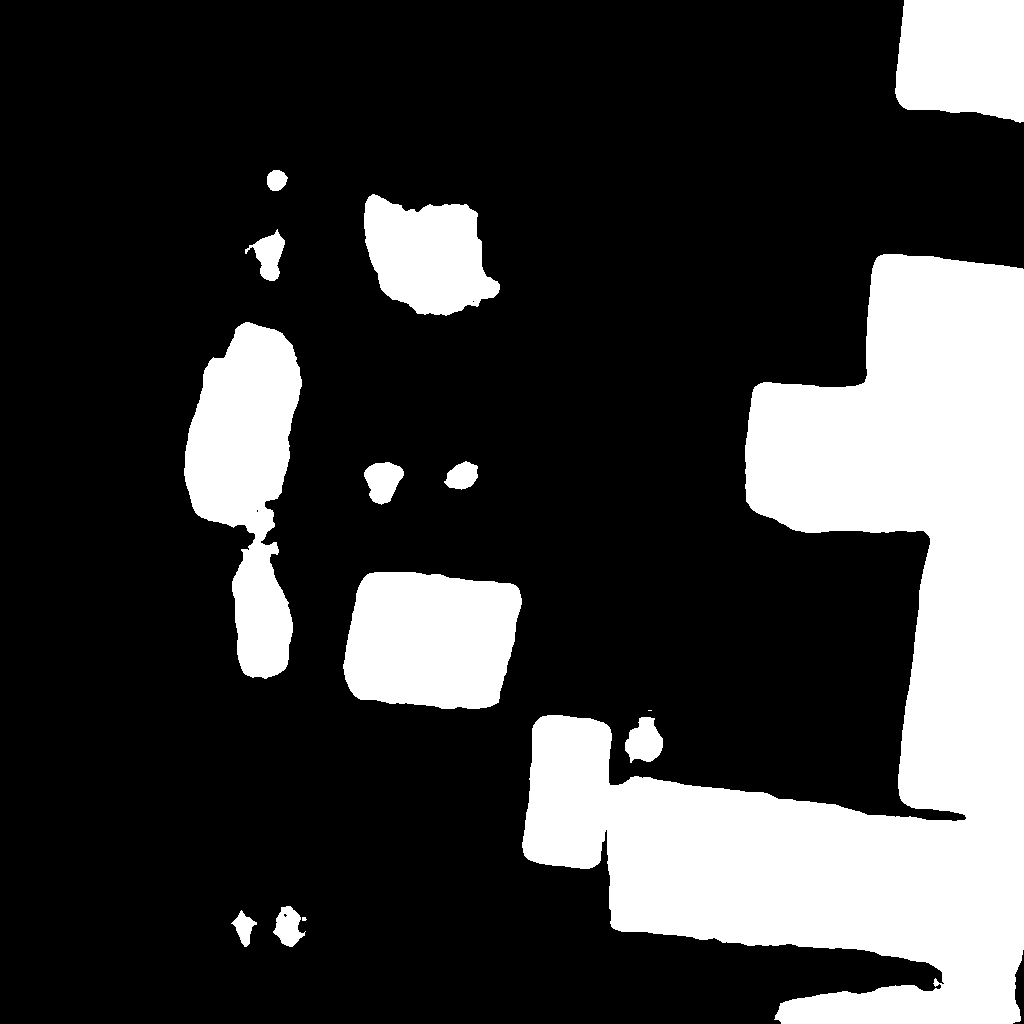} &
 \includegraphics[width=0.7in]{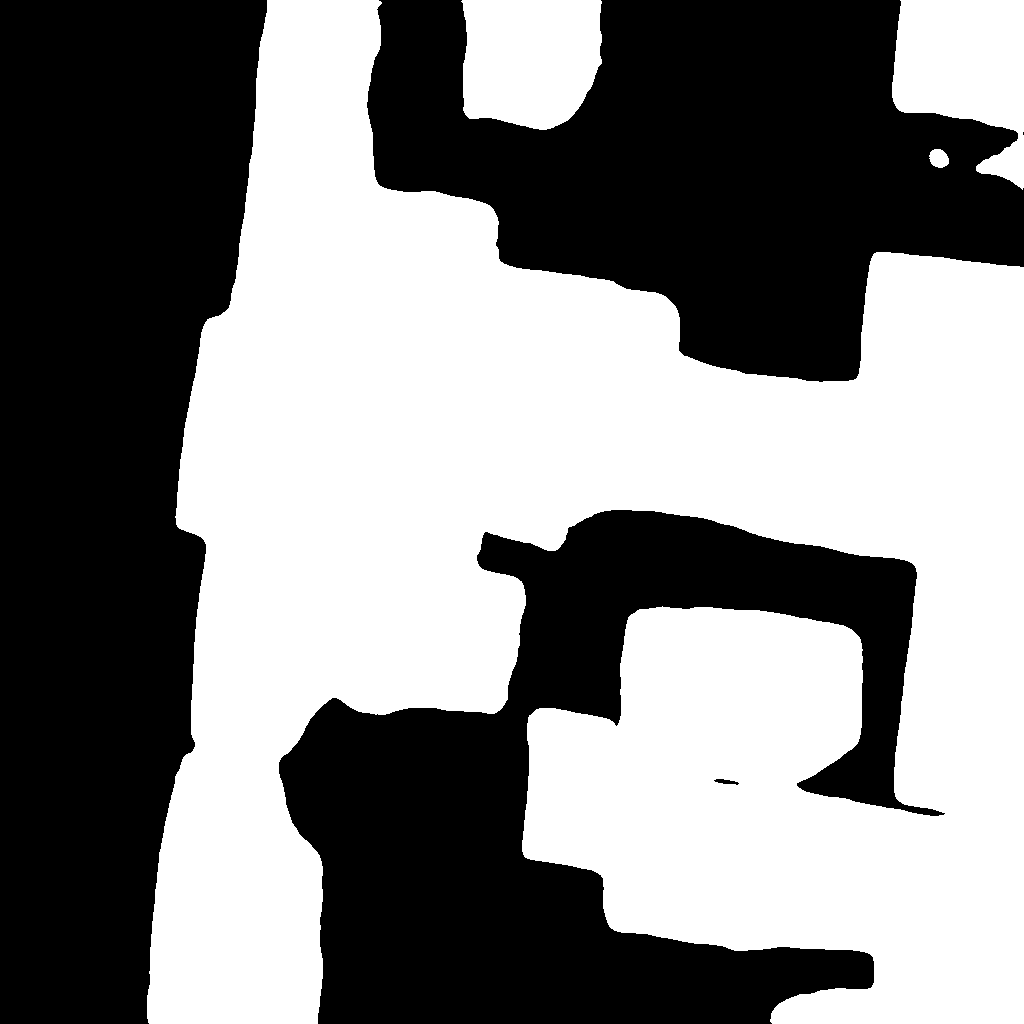} \\

 \includegraphics[width=0.7in]{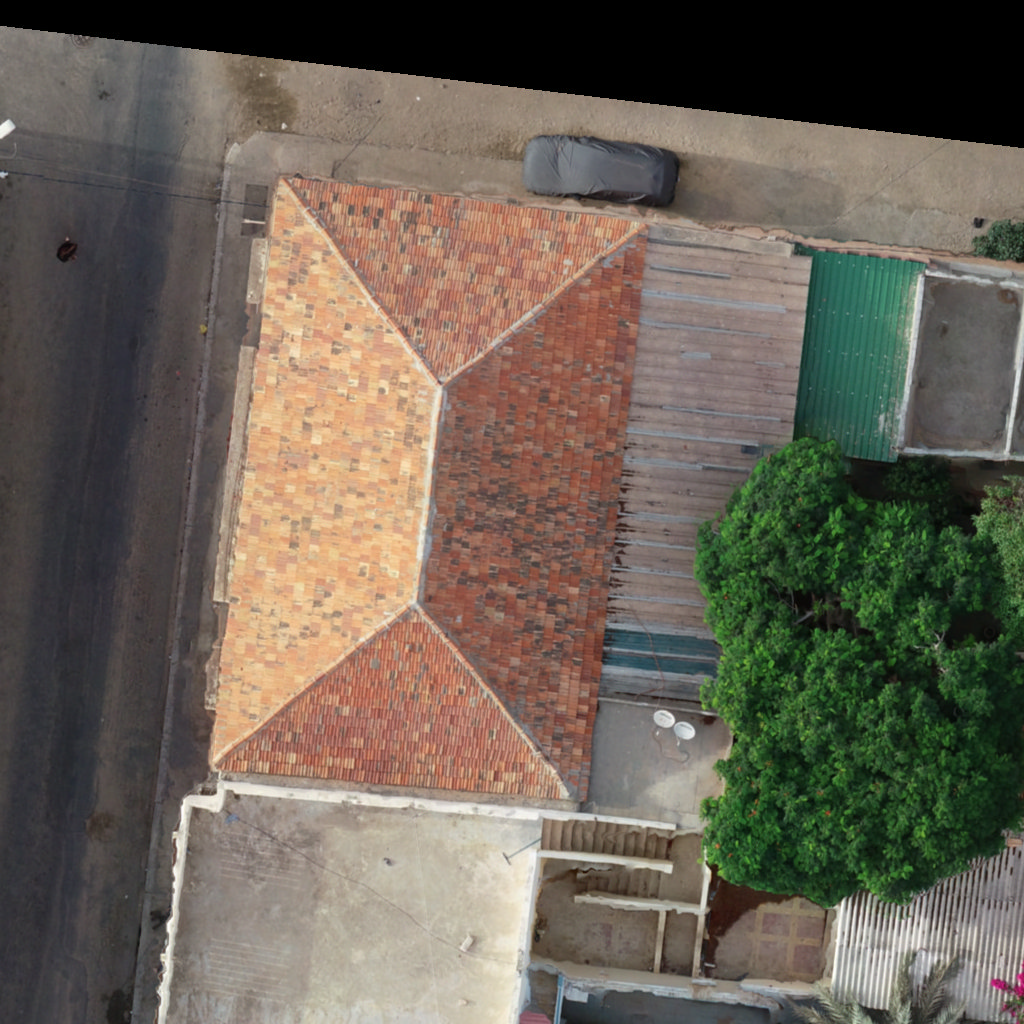} &
 \includegraphics[width=0.7in]{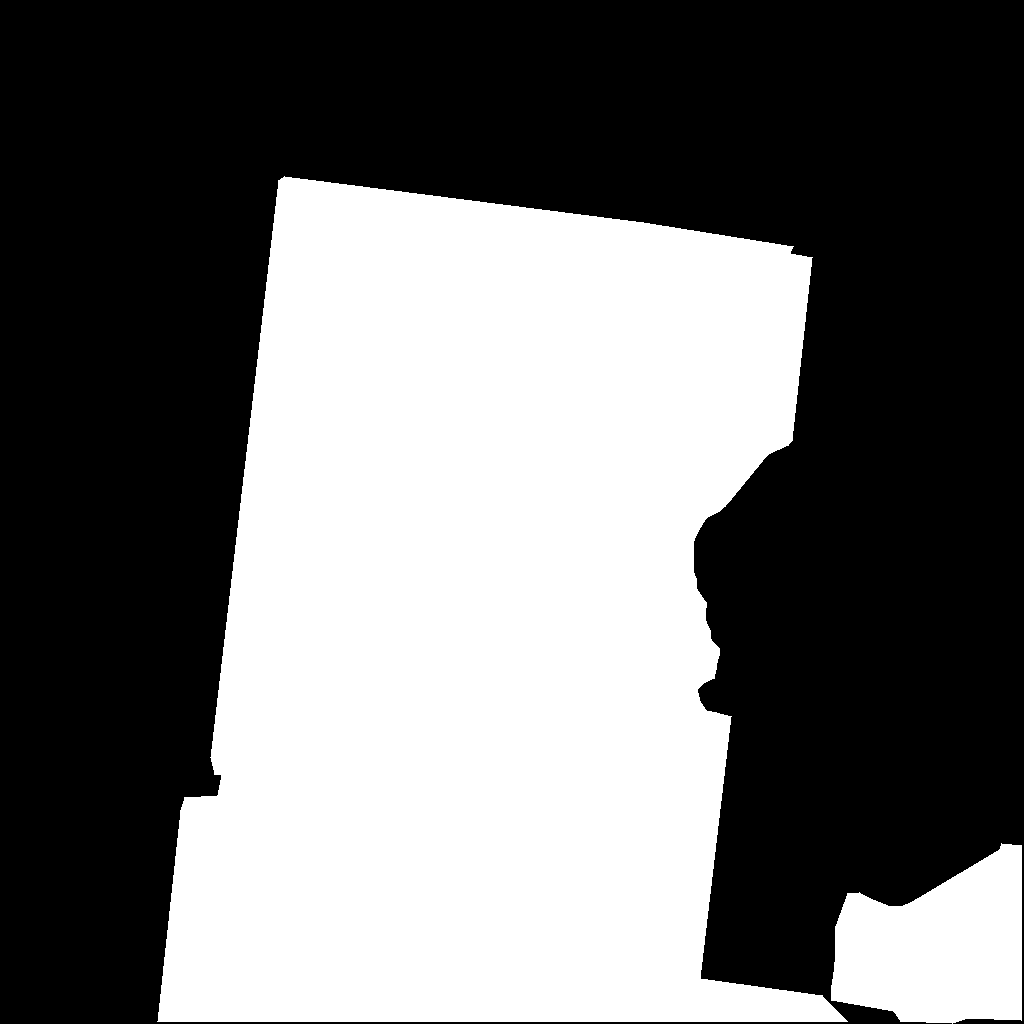} &
 \includegraphics[width=0.7in]{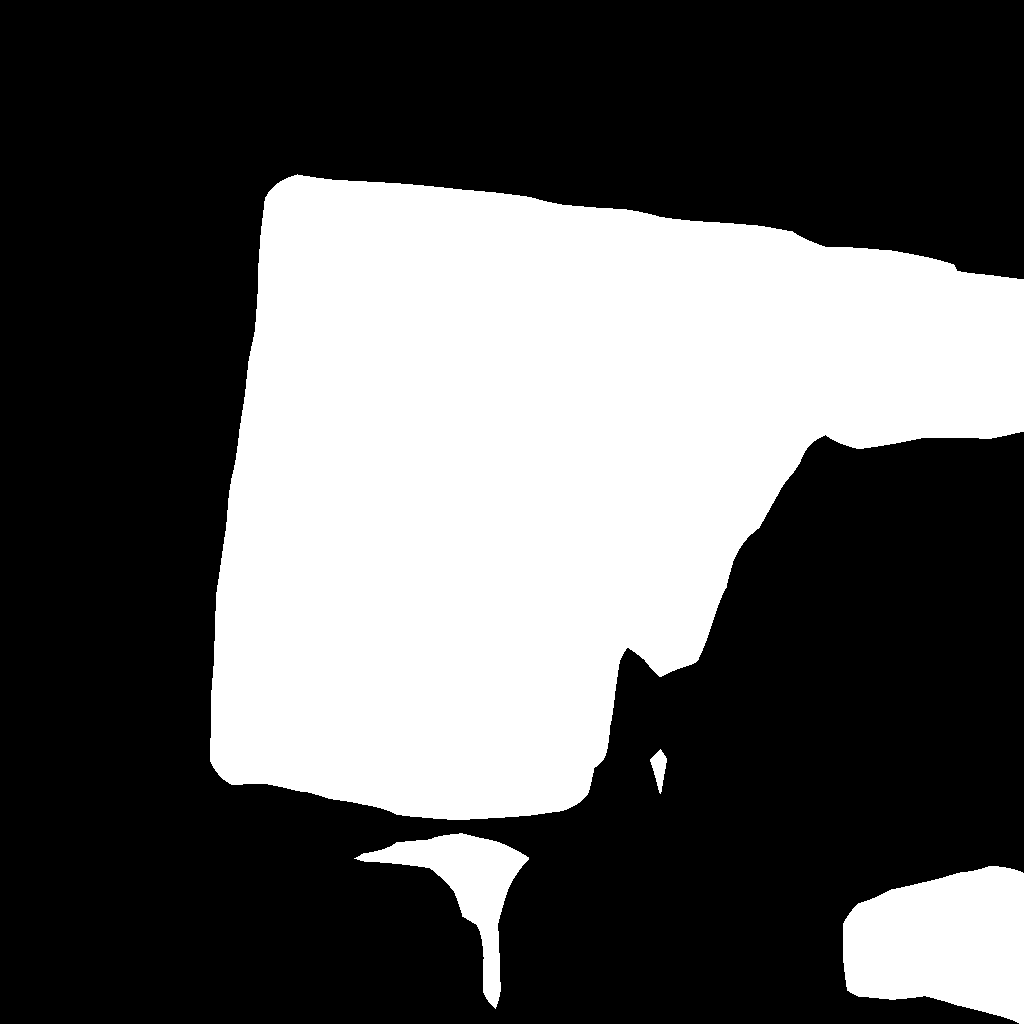} &
 \includegraphics[width=0.7in]{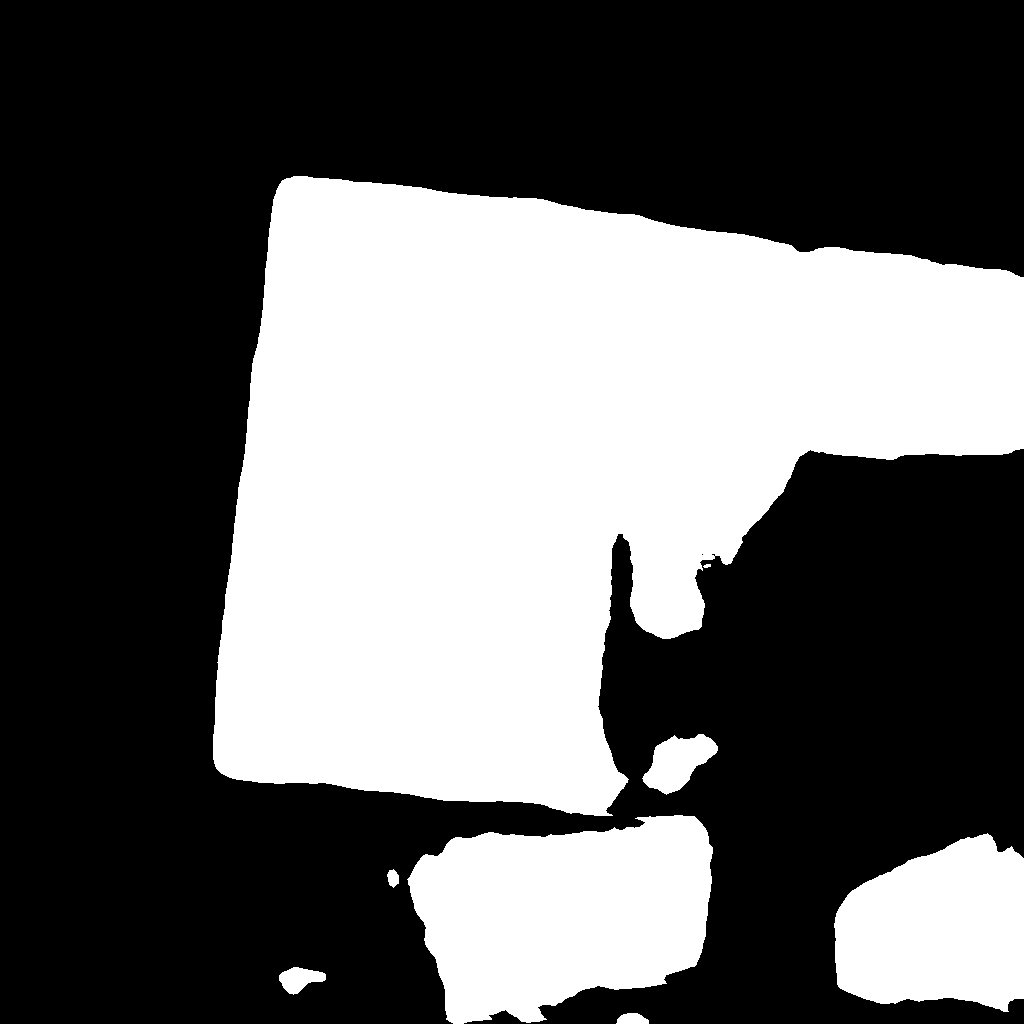} &
 \includegraphics[width=0.7in]{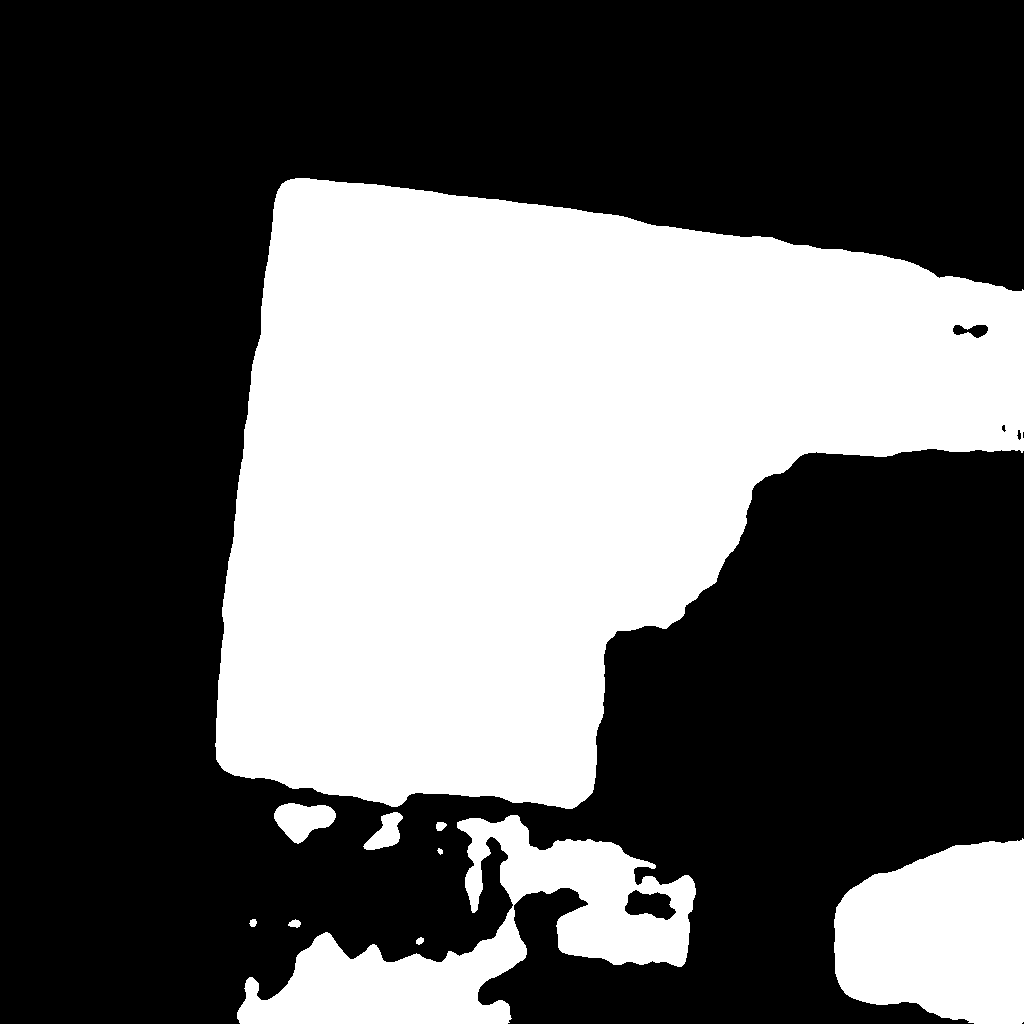} &
 \includegraphics[width=0.7in]{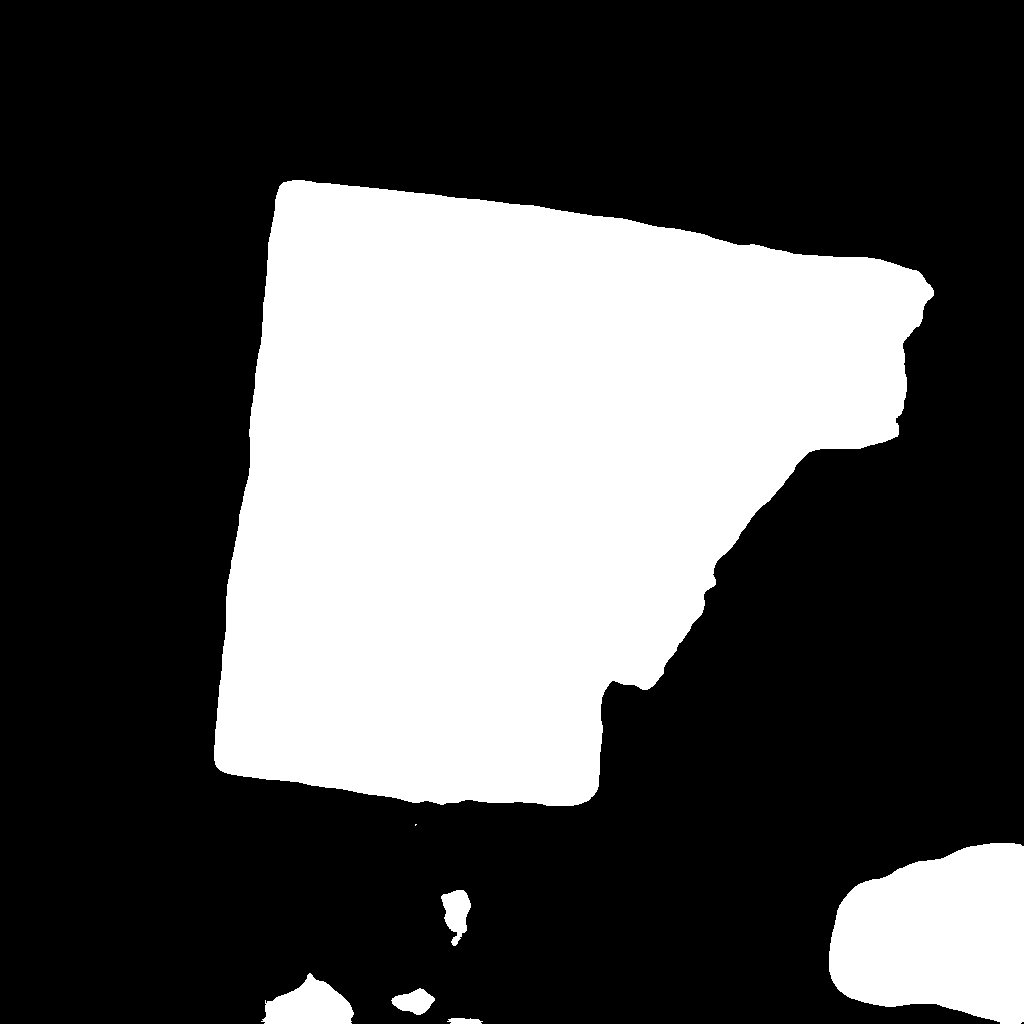} &
 \includegraphics[width=0.7in]{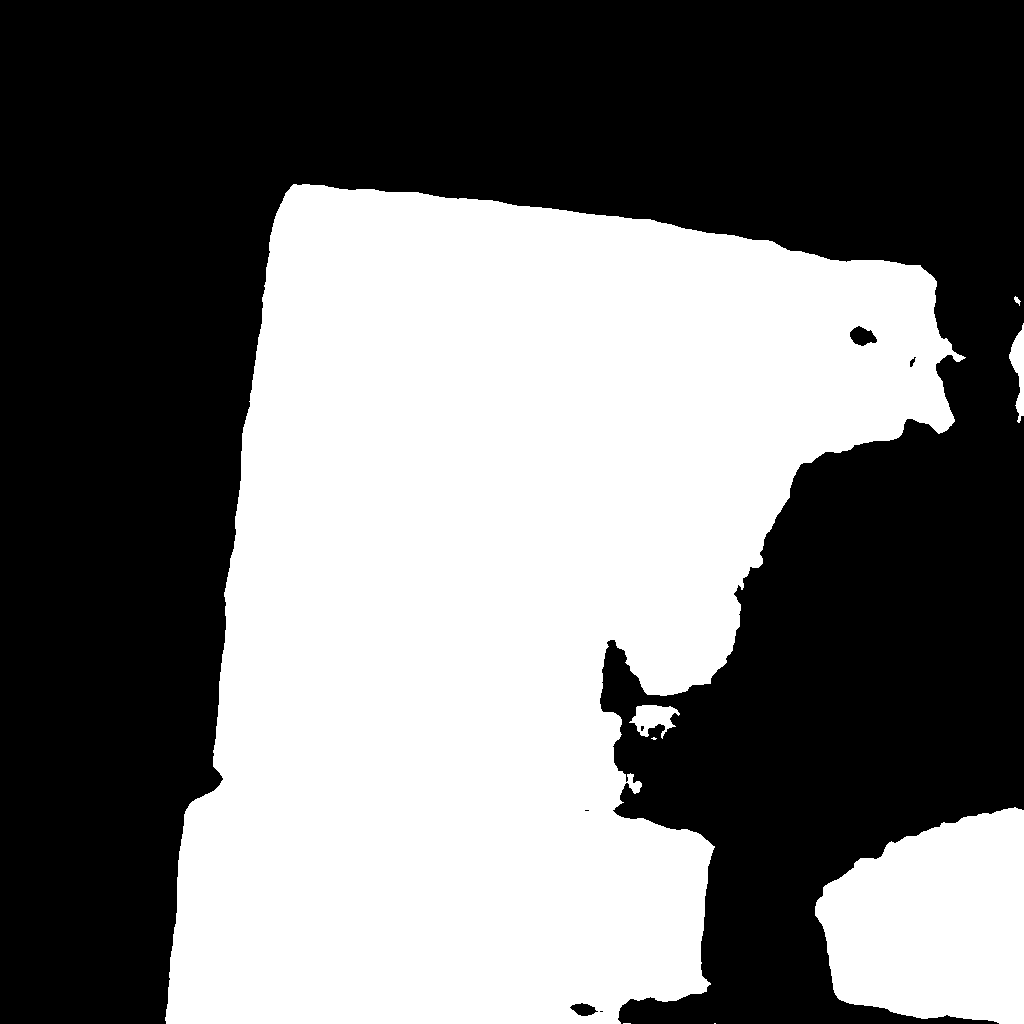} &
 \includegraphics[width=0.7in]{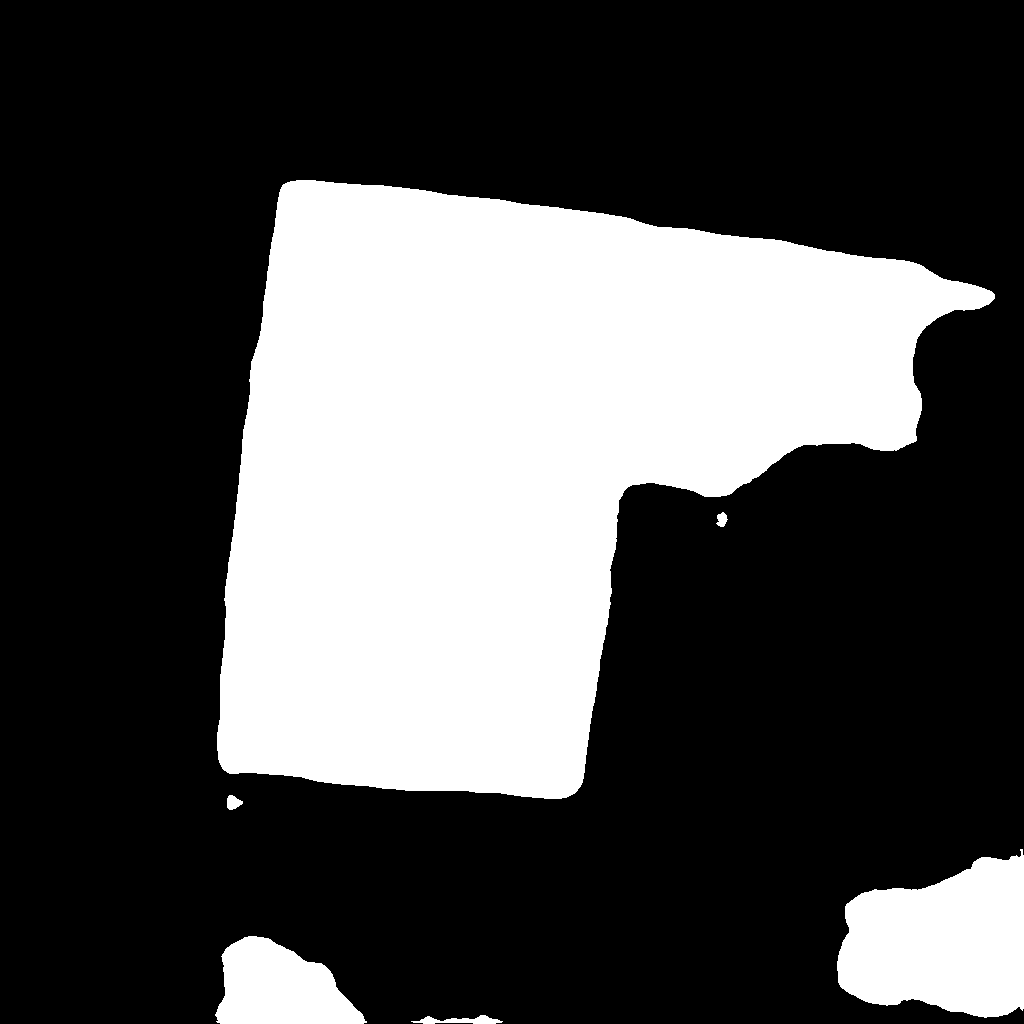} &
 \includegraphics[width=0.7in]{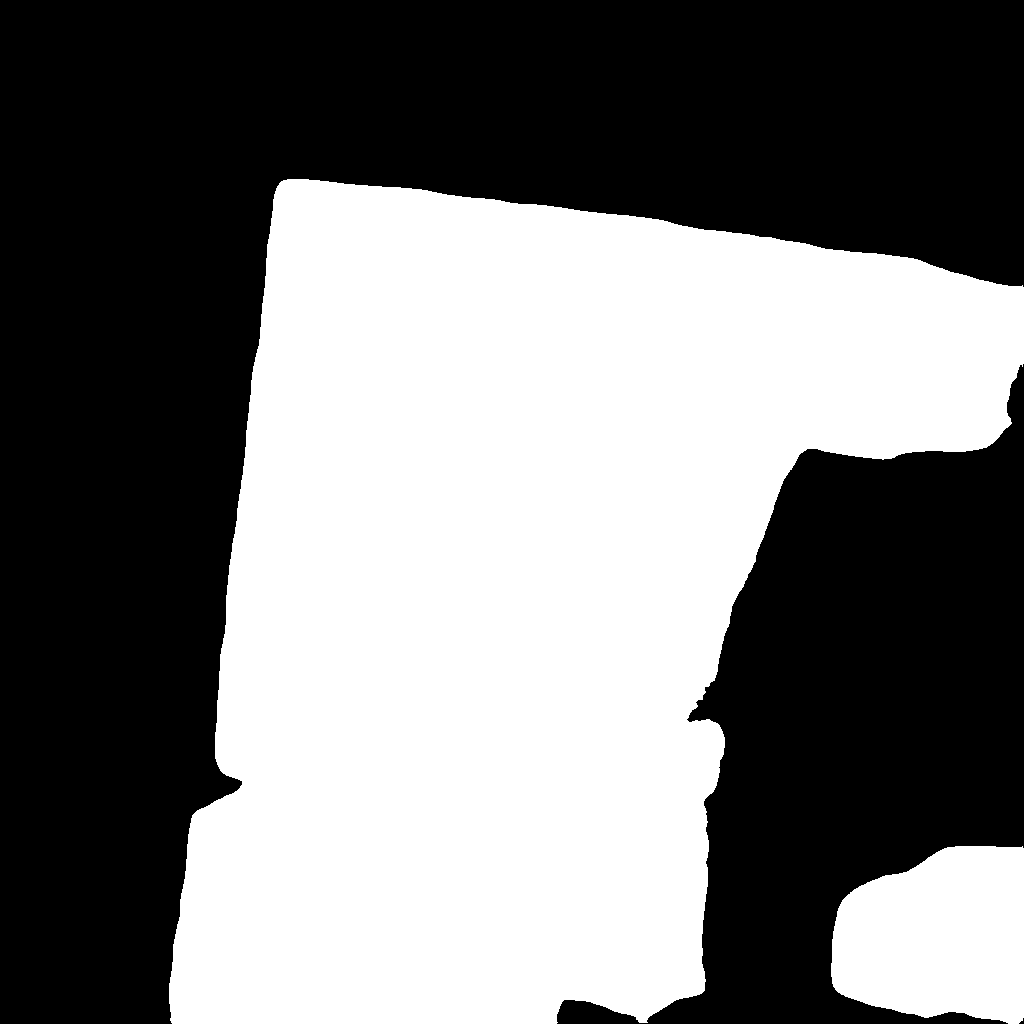} \\

 \includegraphics[width=0.7in]{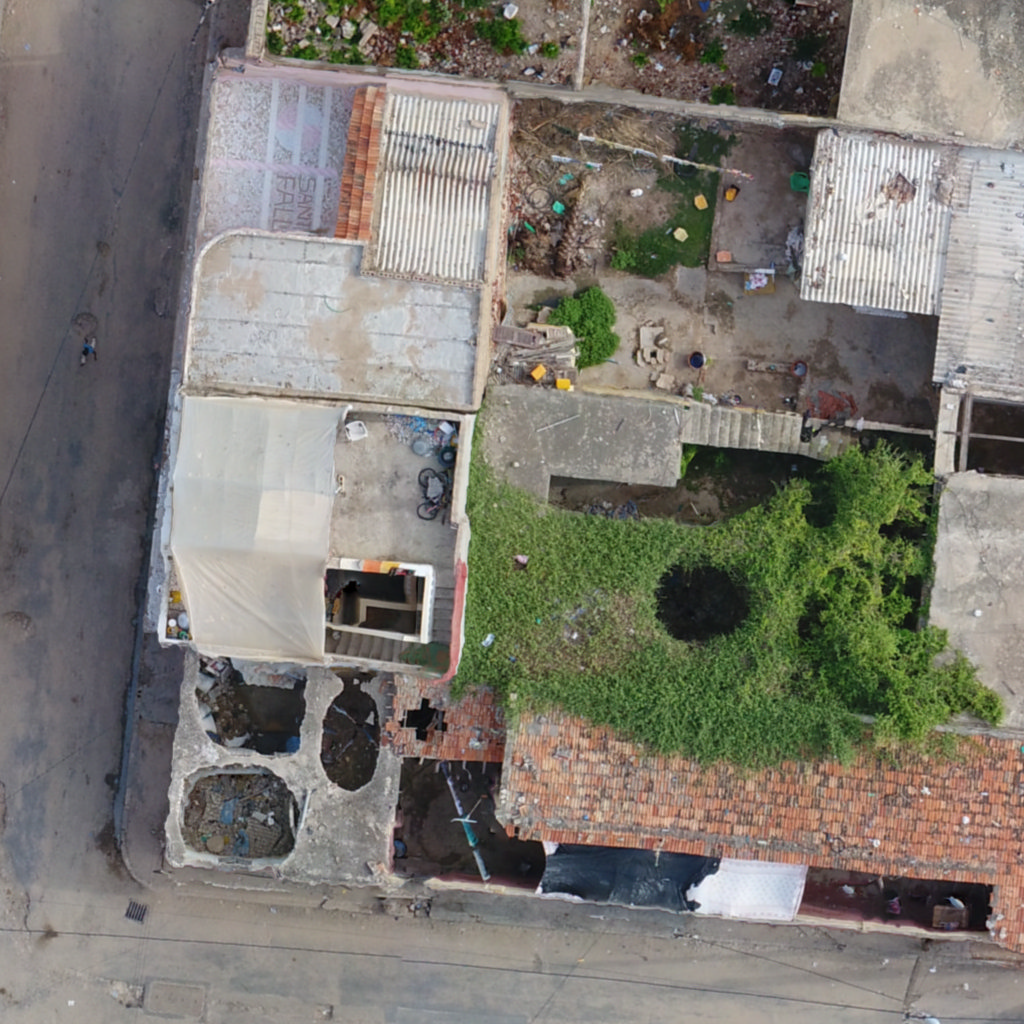} &
 \includegraphics[width=0.7in]{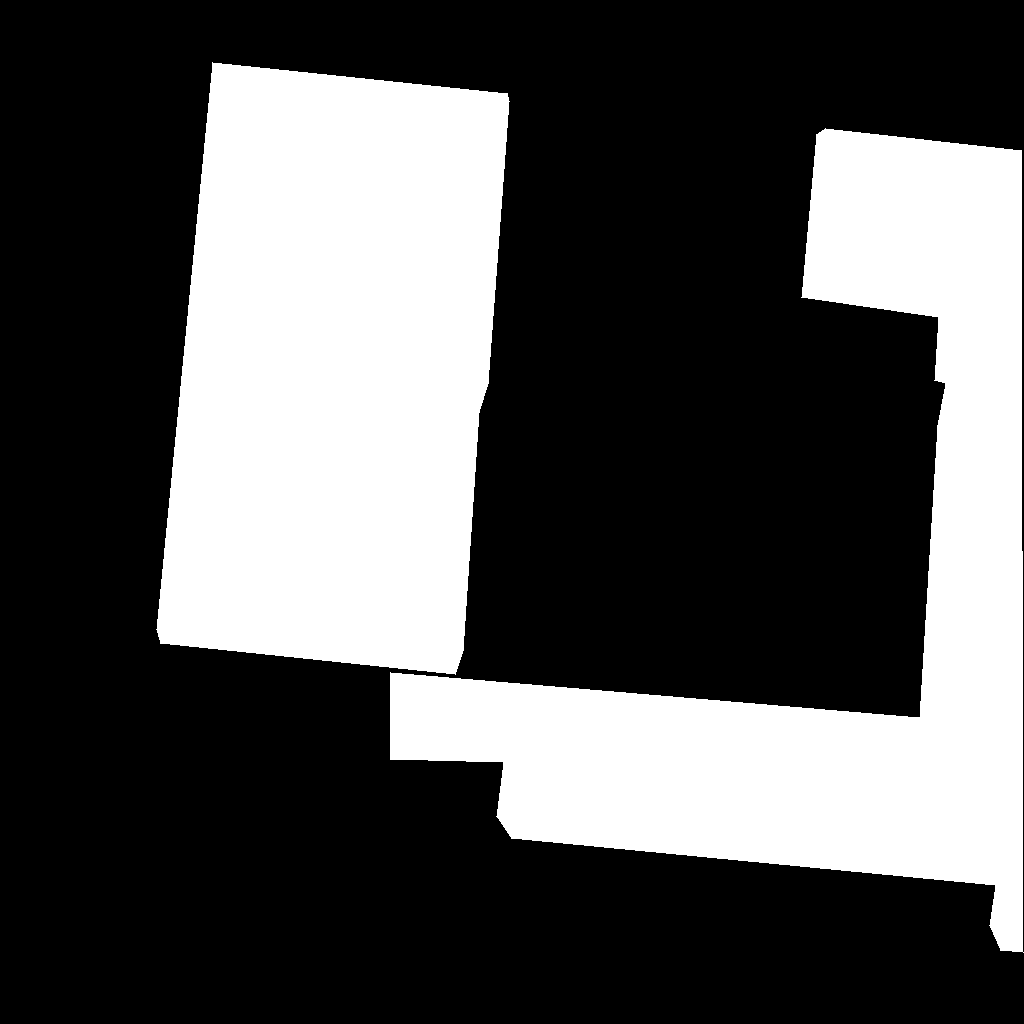} &
 \includegraphics[width=0.7in]{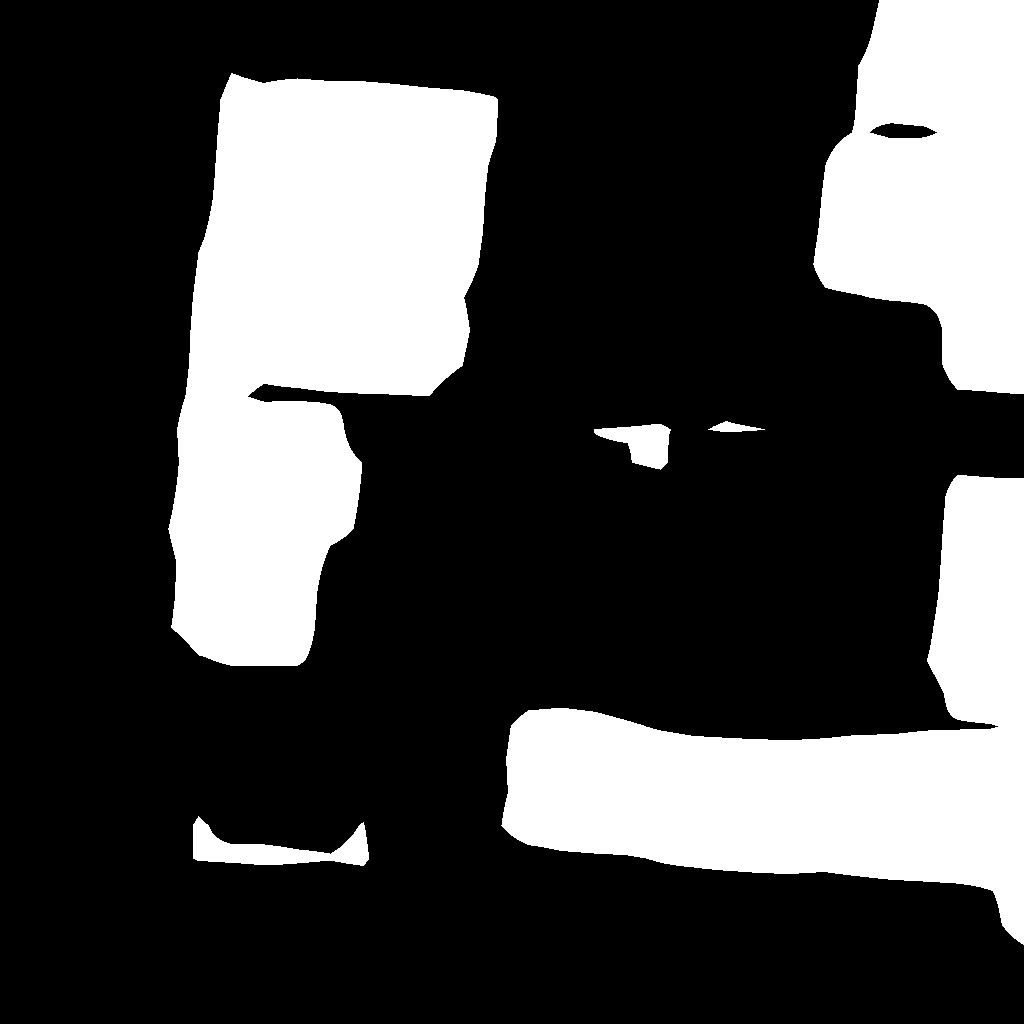} &
 \includegraphics[width=0.7in]{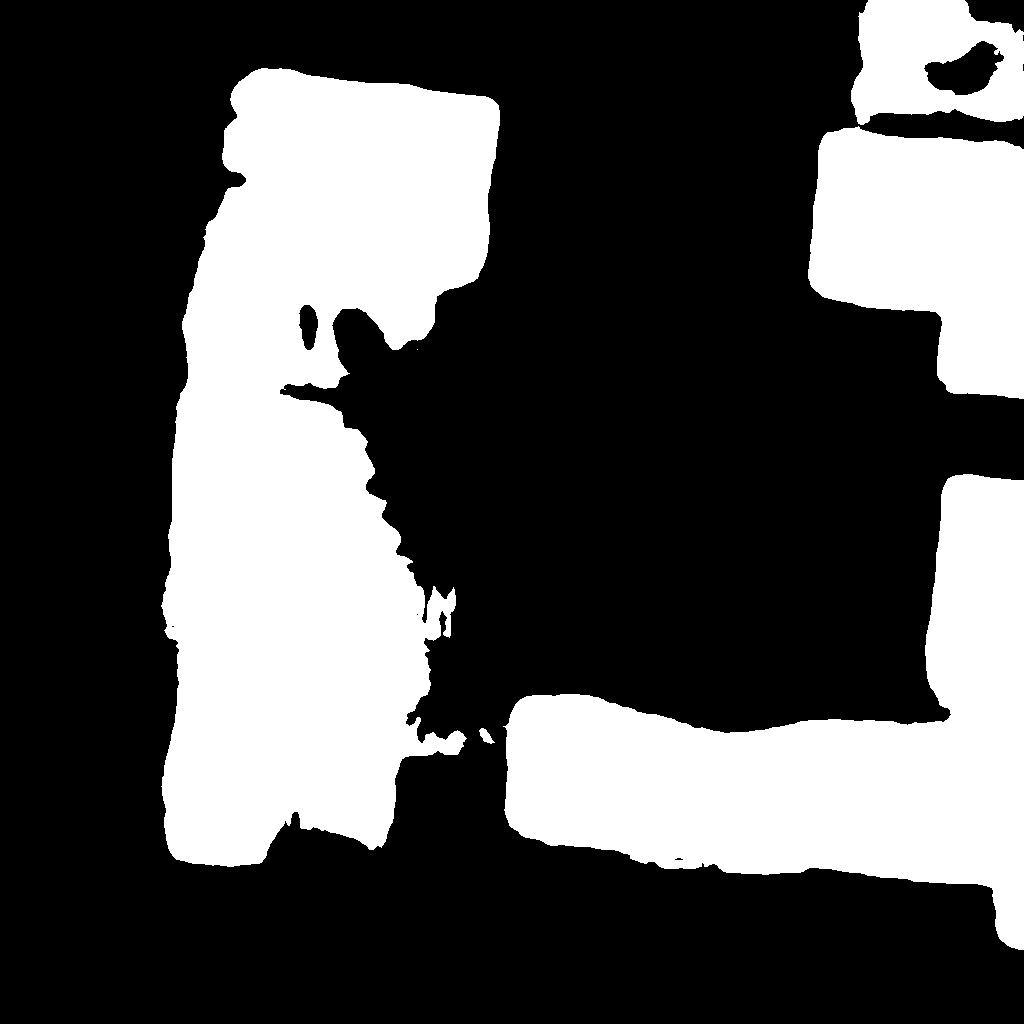} &
 \includegraphics[width=0.7in]{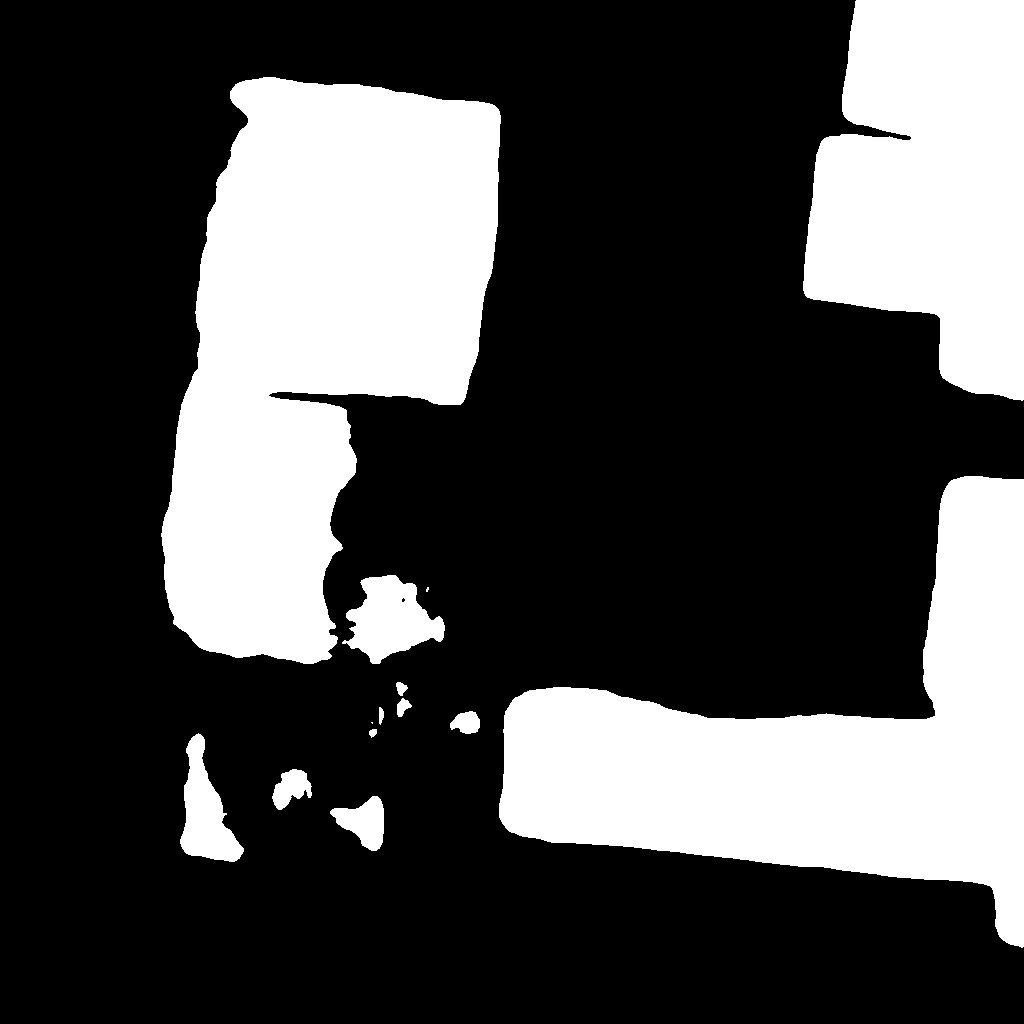} &
 \includegraphics[width=0.7in]{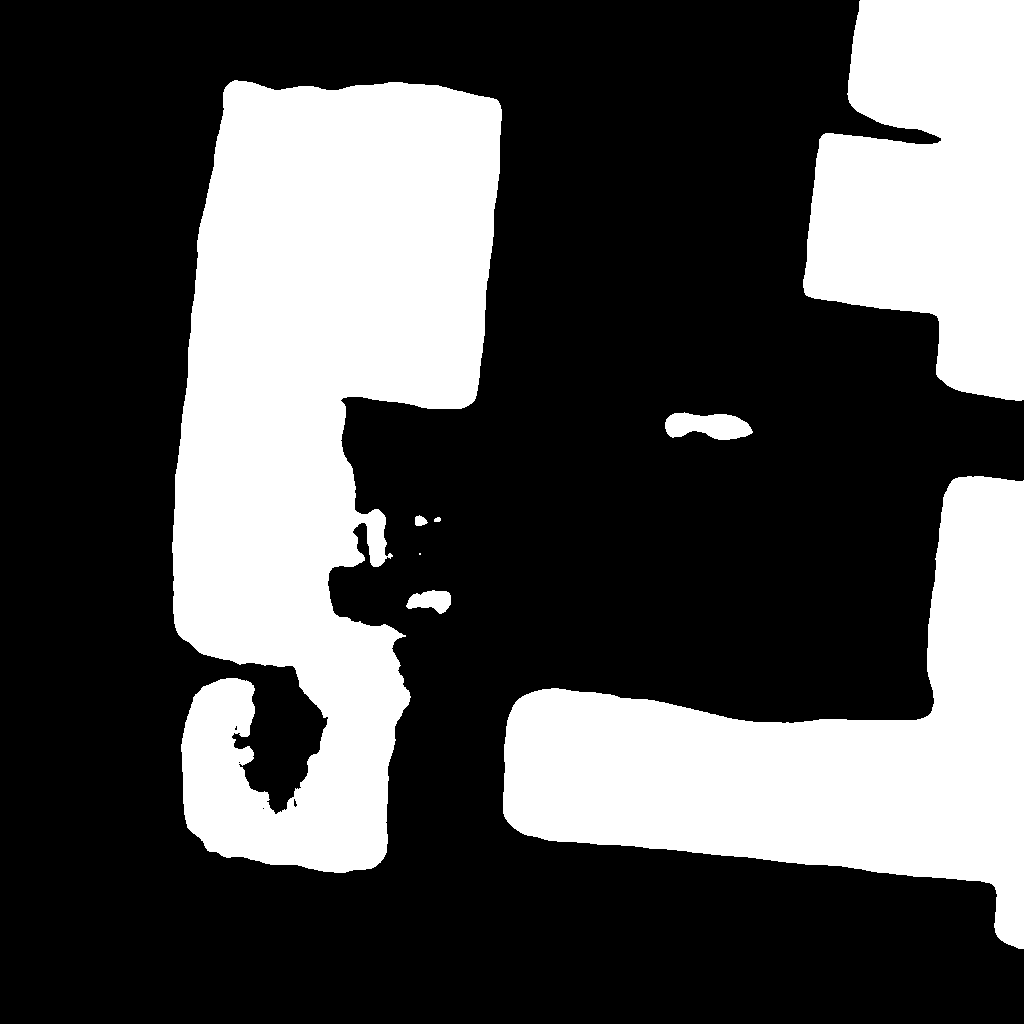} &
 \includegraphics[width=0.7in]{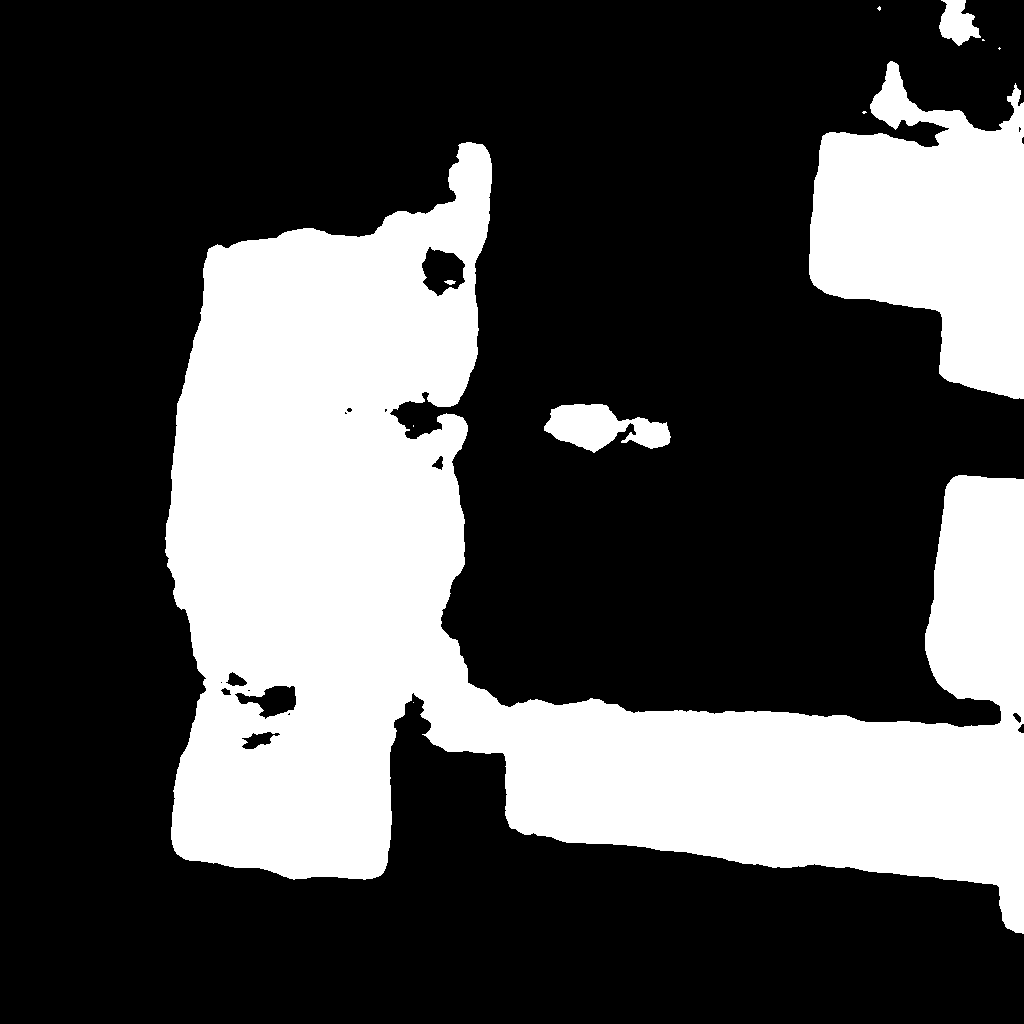} &
 \includegraphics[width=0.7in]{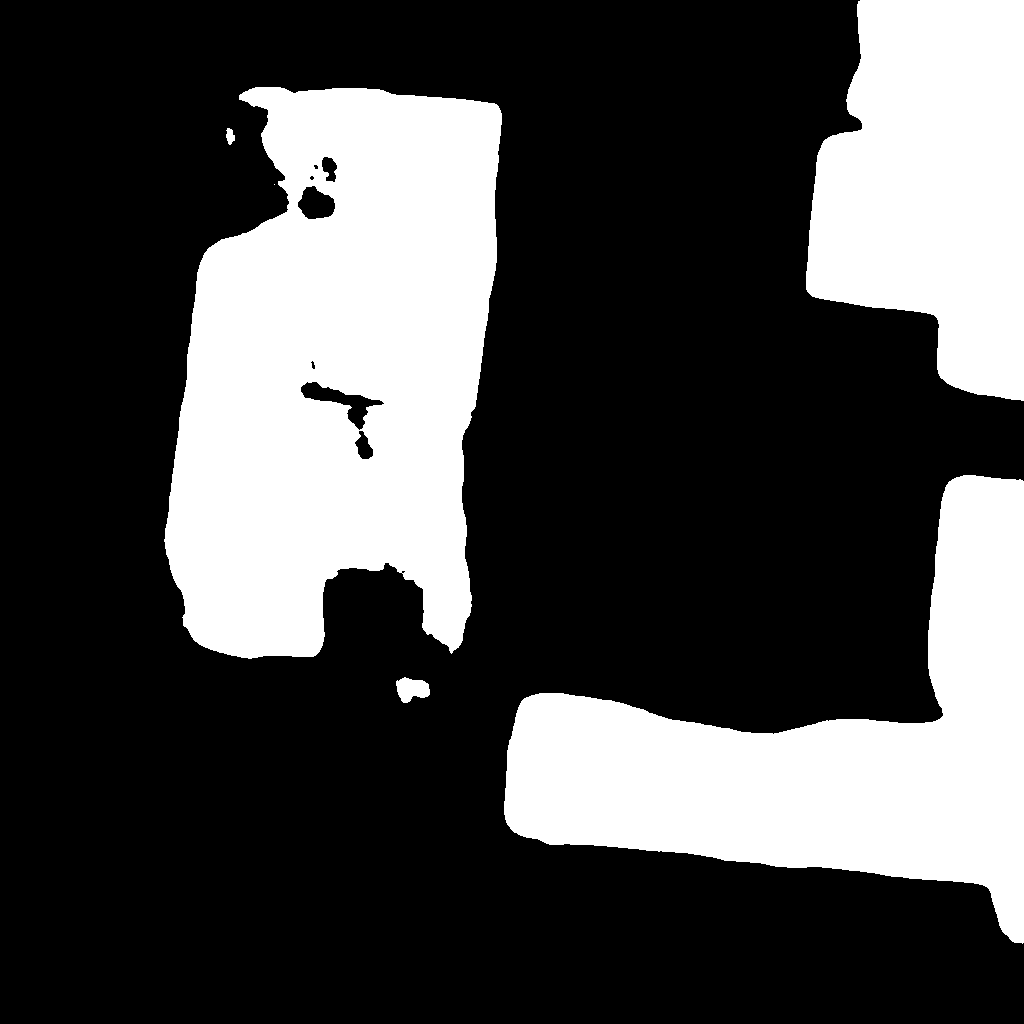} &
 \includegraphics[width=0.7in]{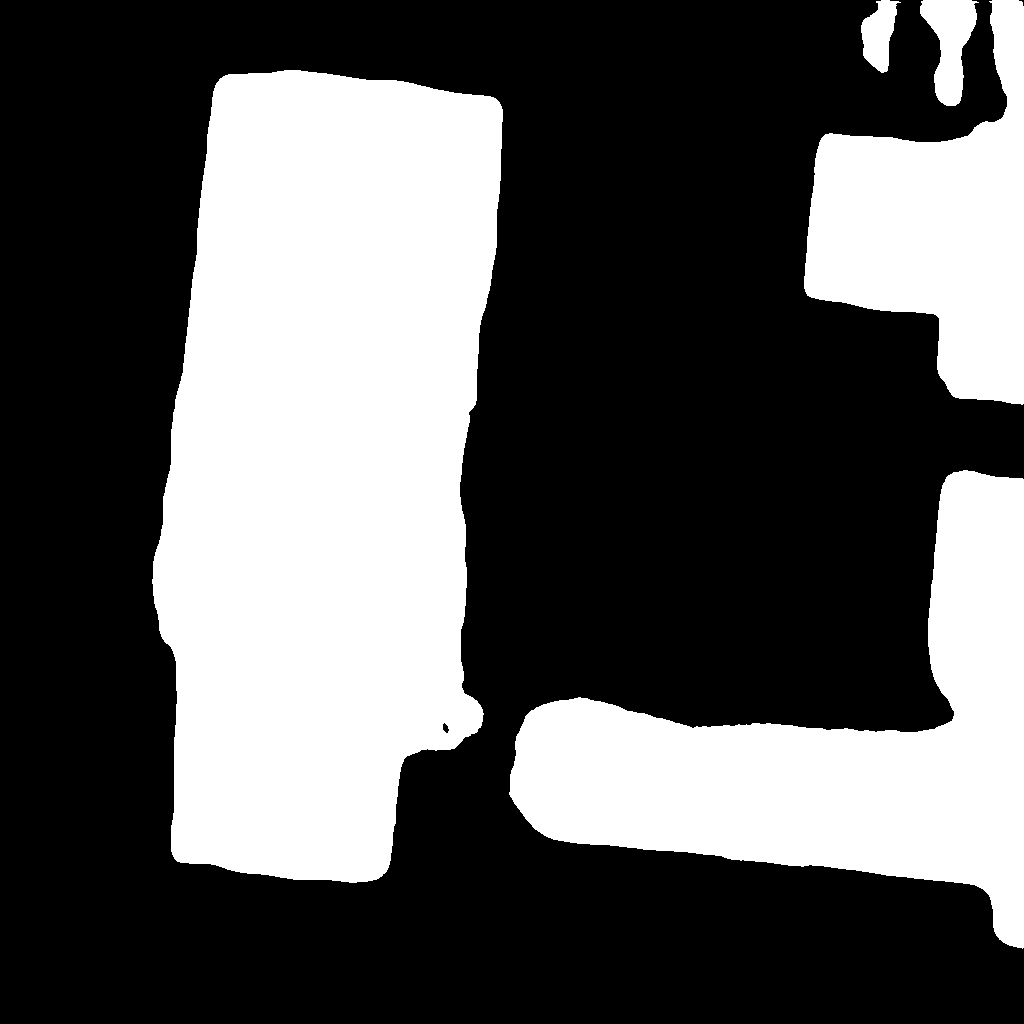} \\

 \includegraphics[width=0.7in]{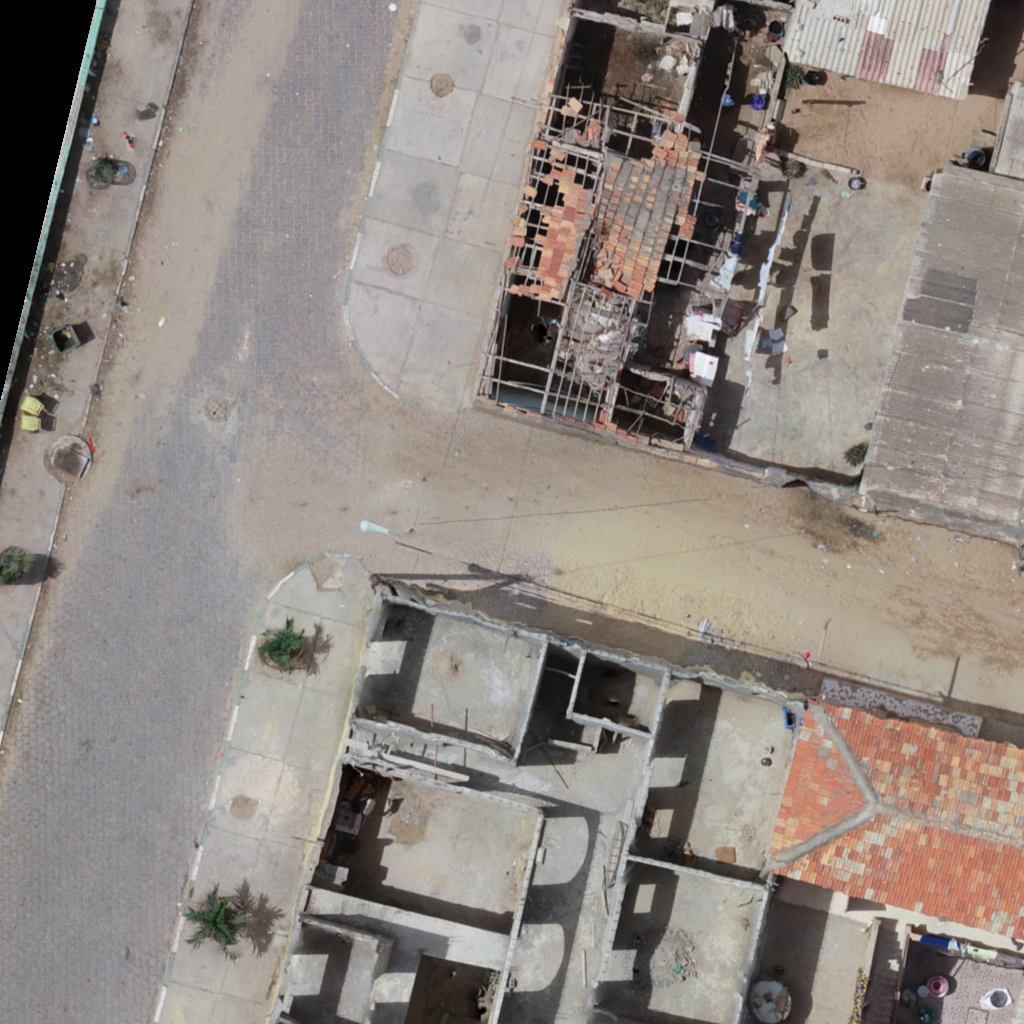} &
 \includegraphics[width=0.7in]{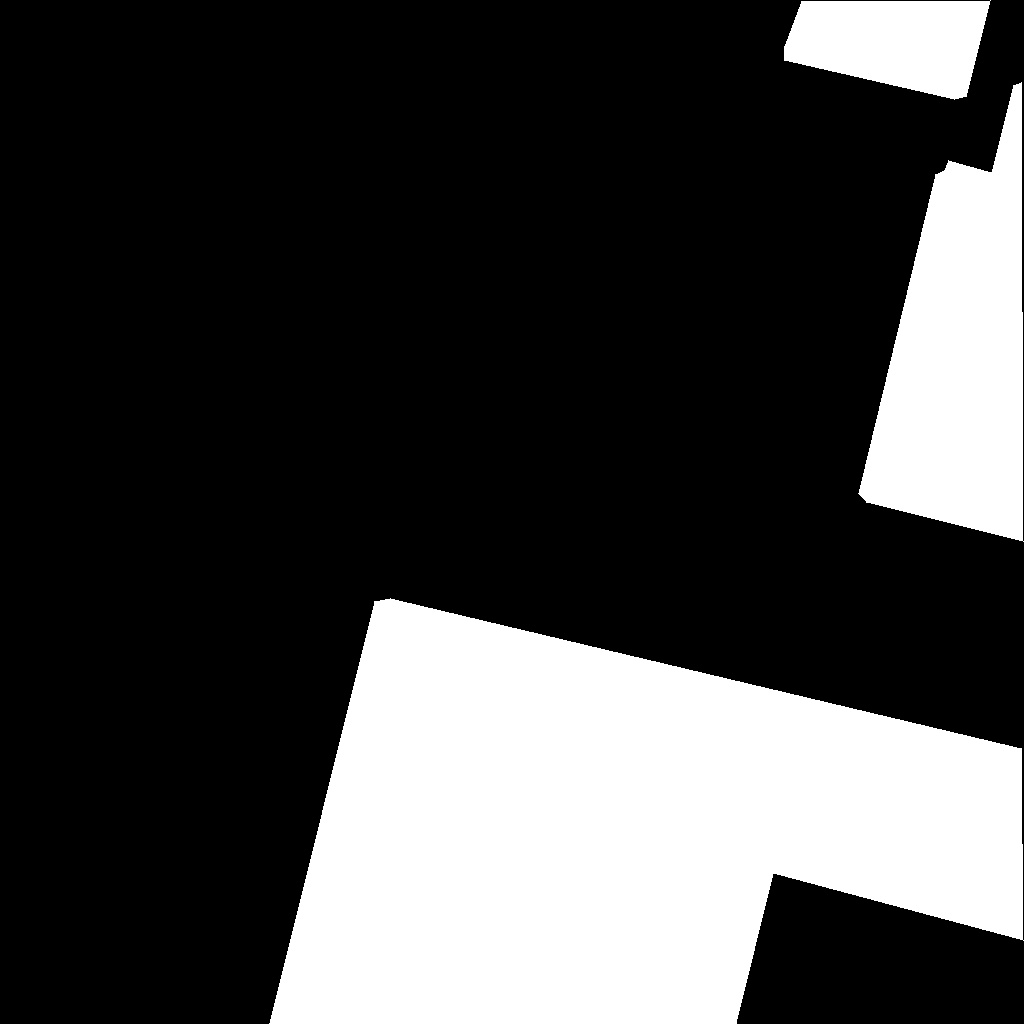} &
 \includegraphics[width=0.7in]{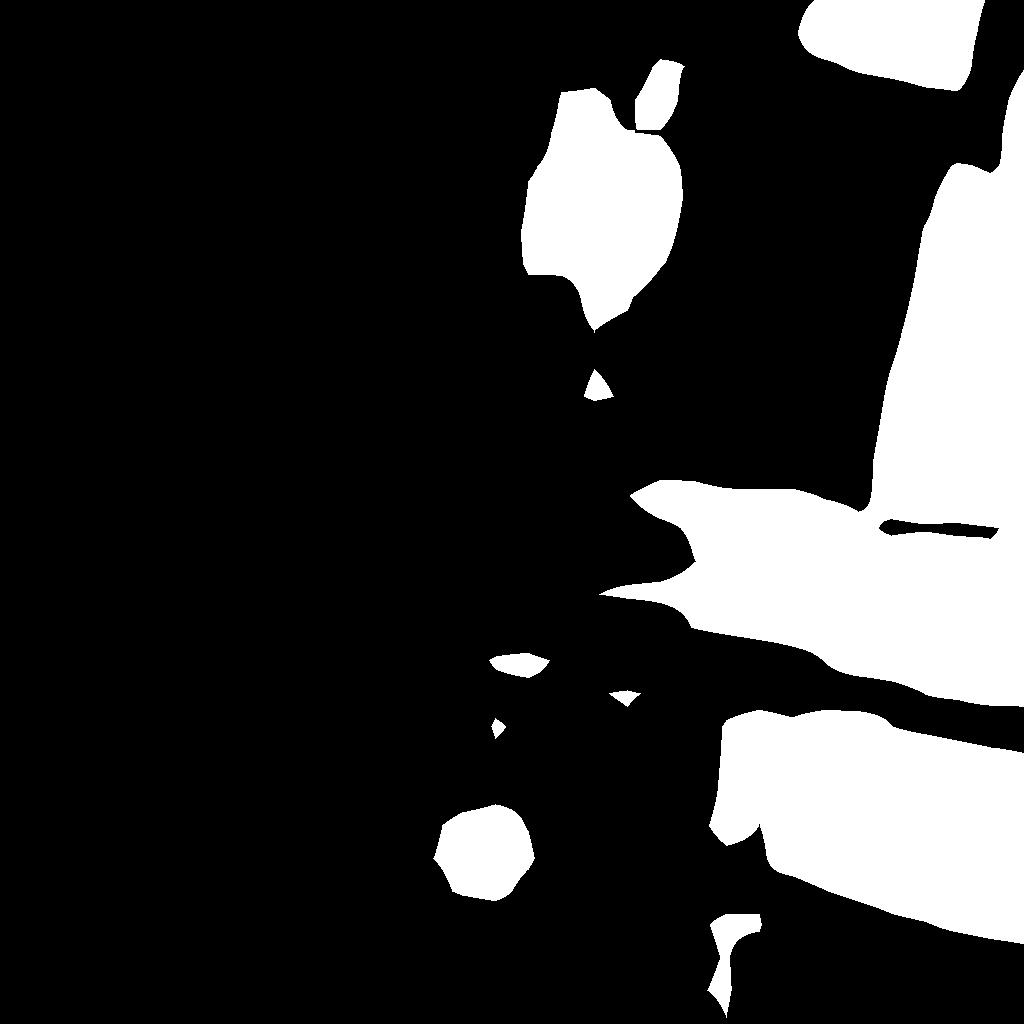} &
 \includegraphics[width=0.7in]{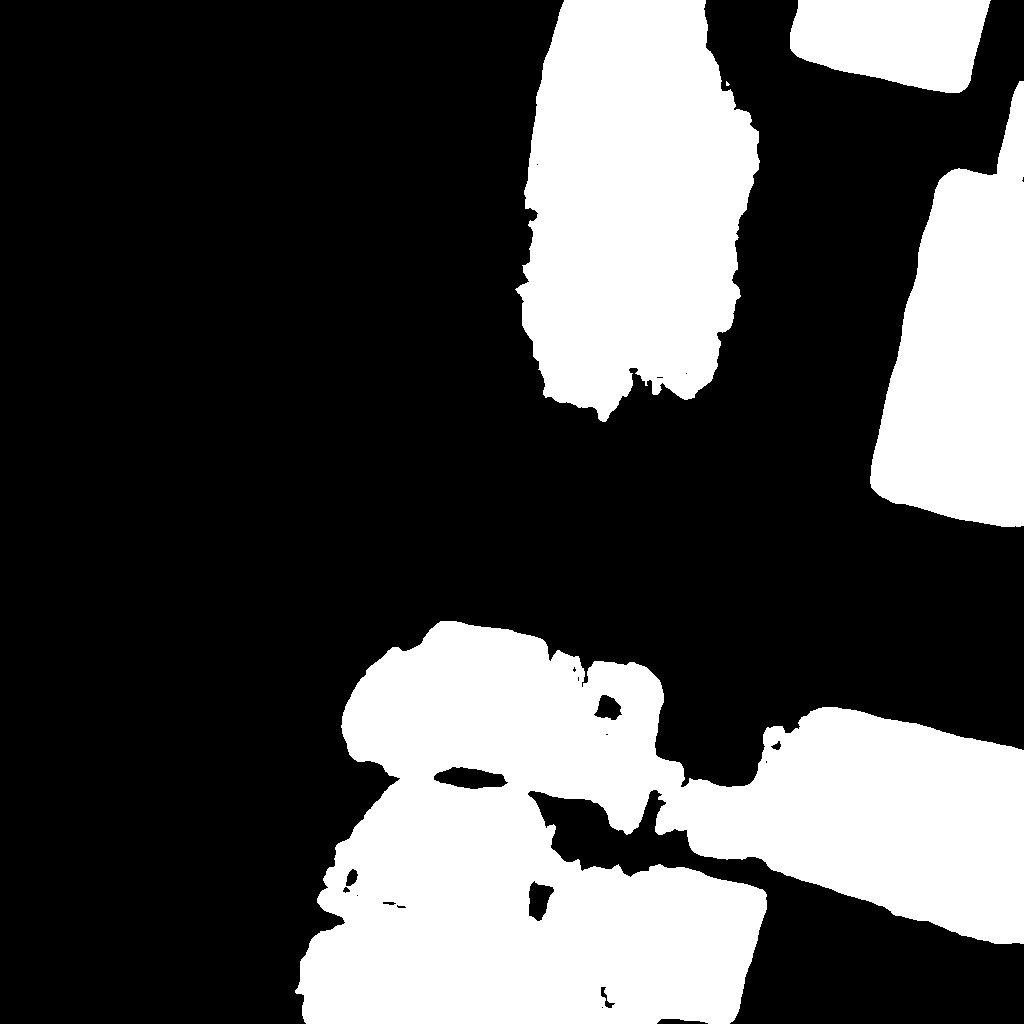} &
 \includegraphics[width=0.7in]{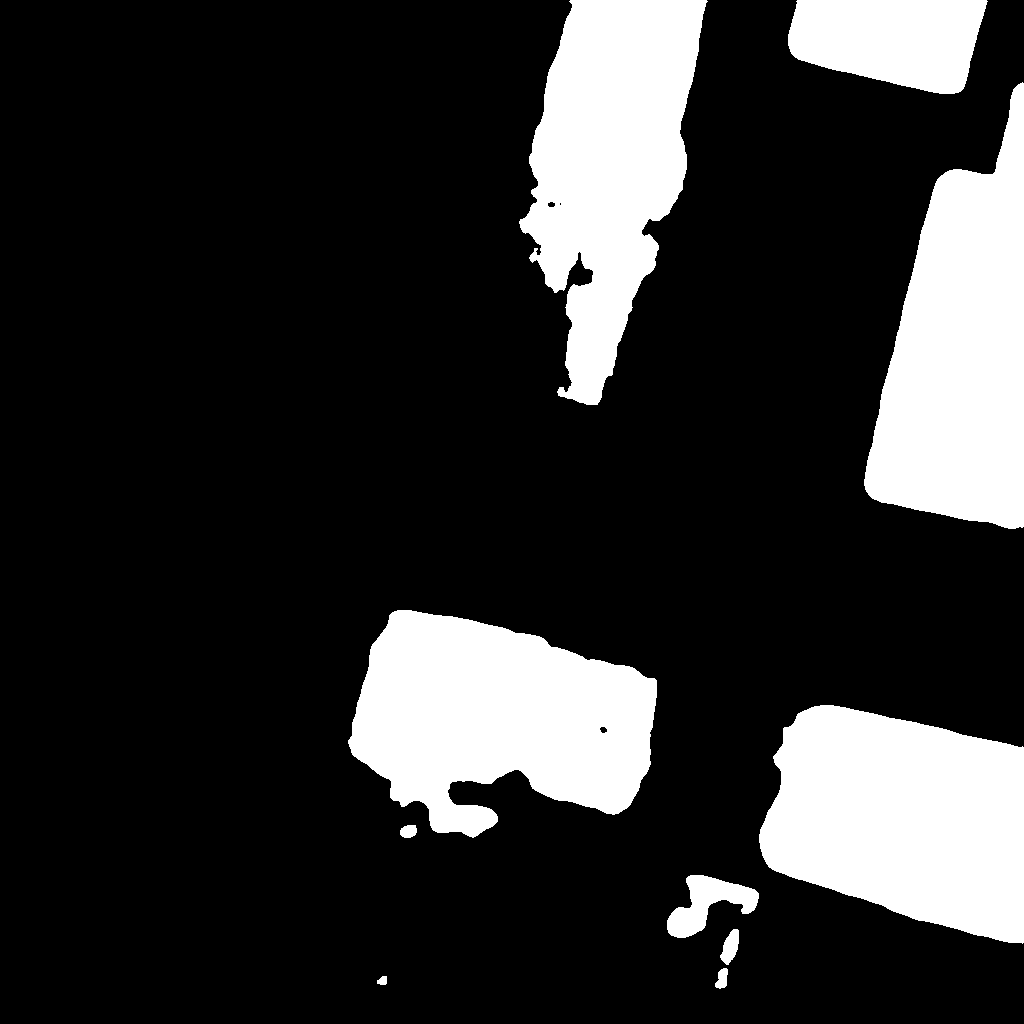} &
 \includegraphics[width=0.7in]{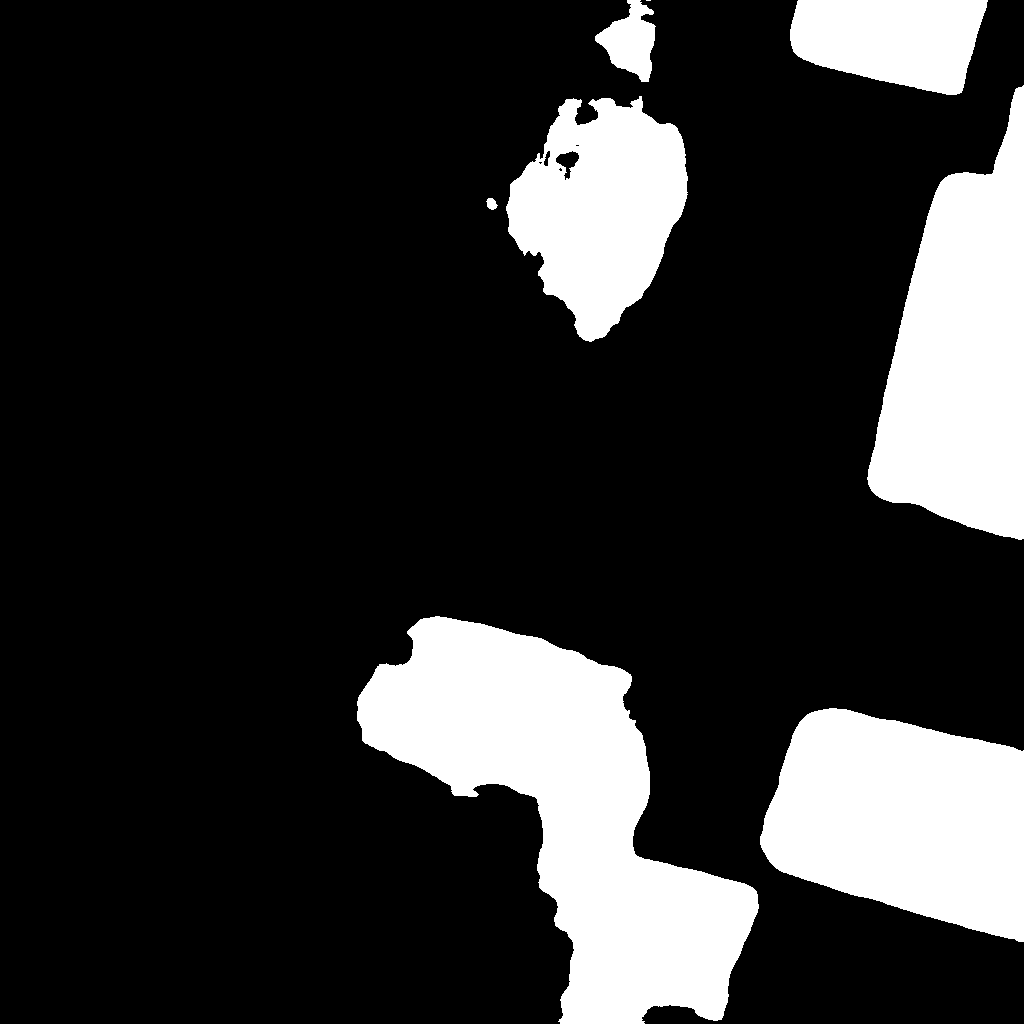} &
 \includegraphics[width=0.7in]{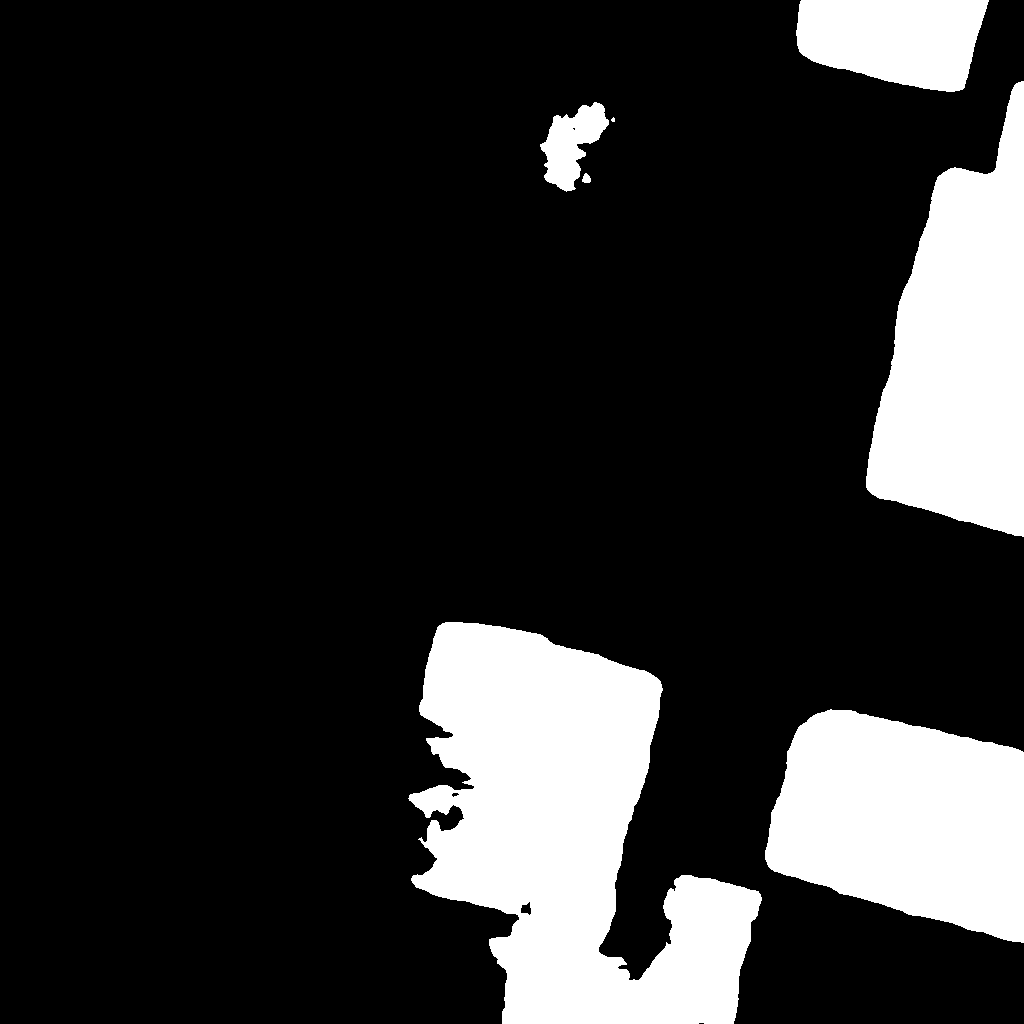} &
 \includegraphics[width=0.7in]{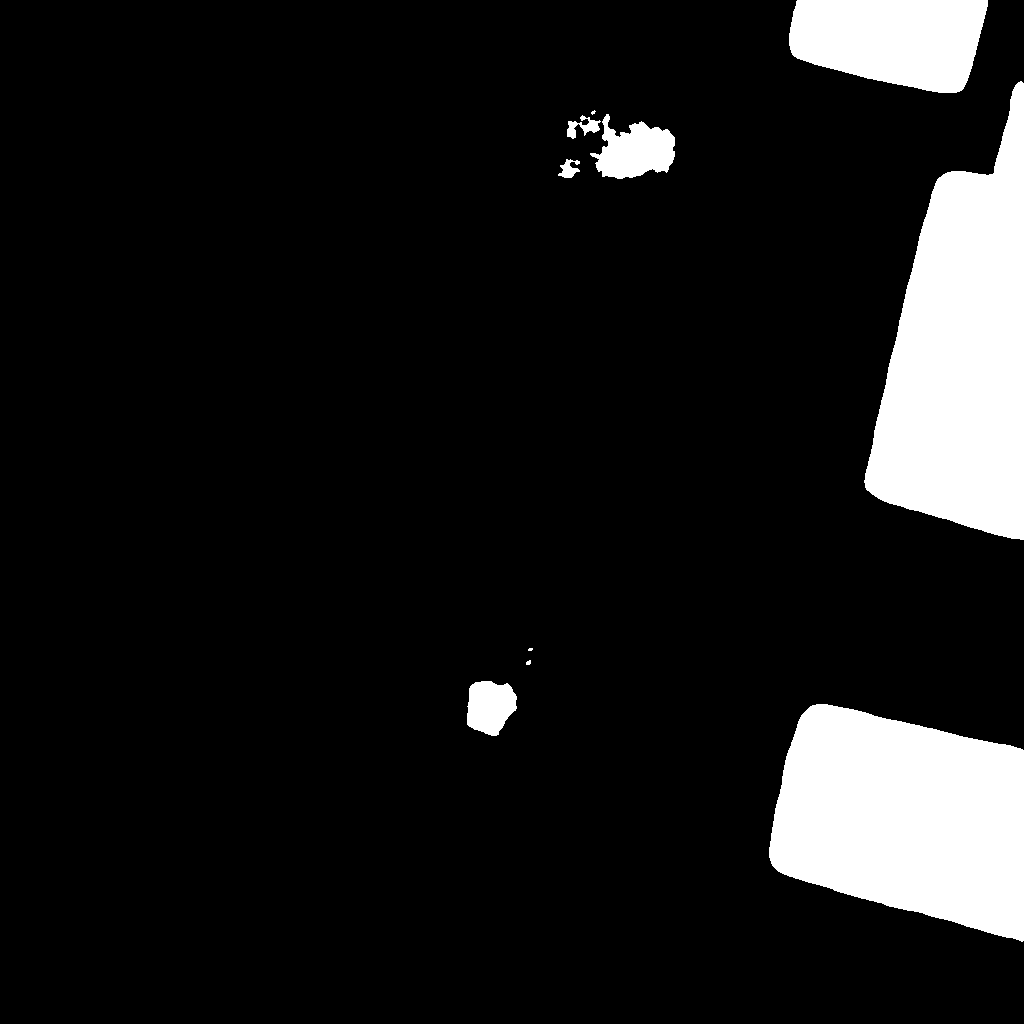} &
 \includegraphics[width=0.7in]{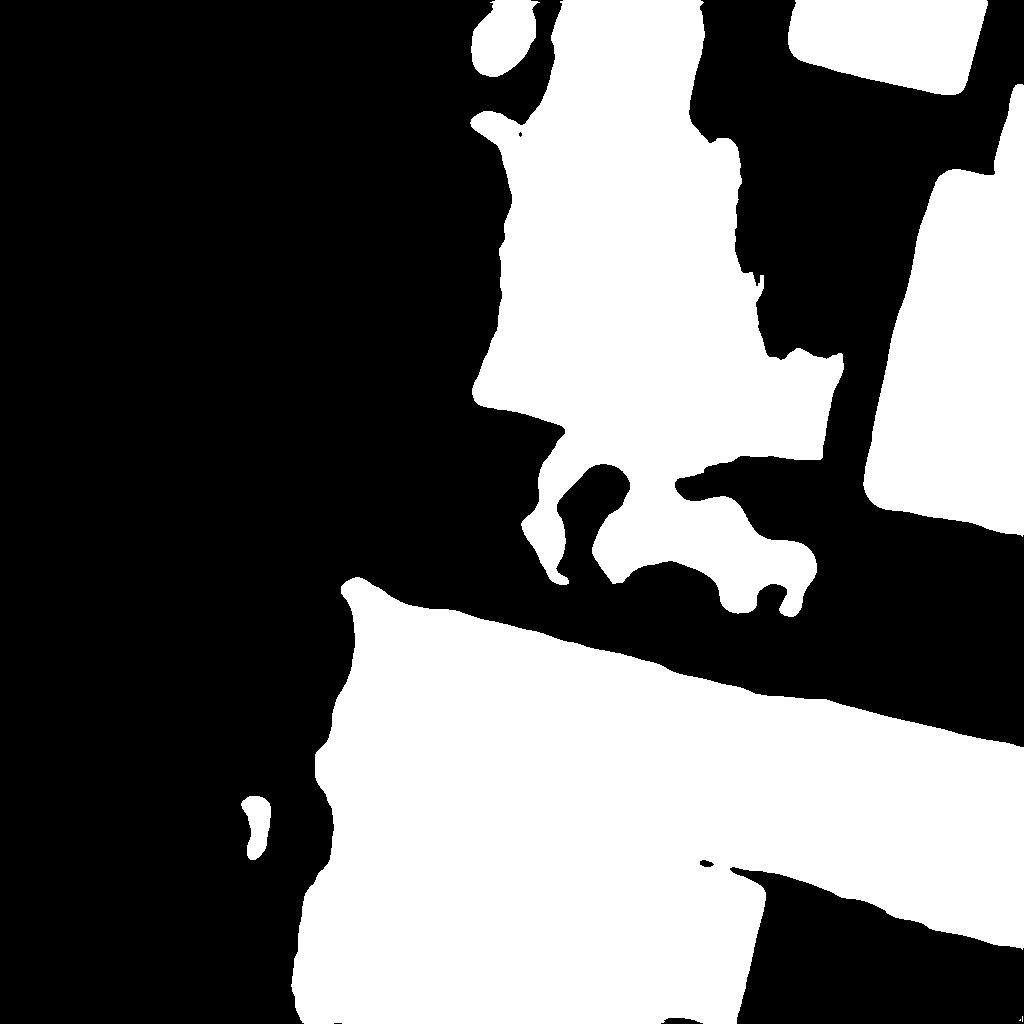} \\

 \includegraphics[width=0.7in]{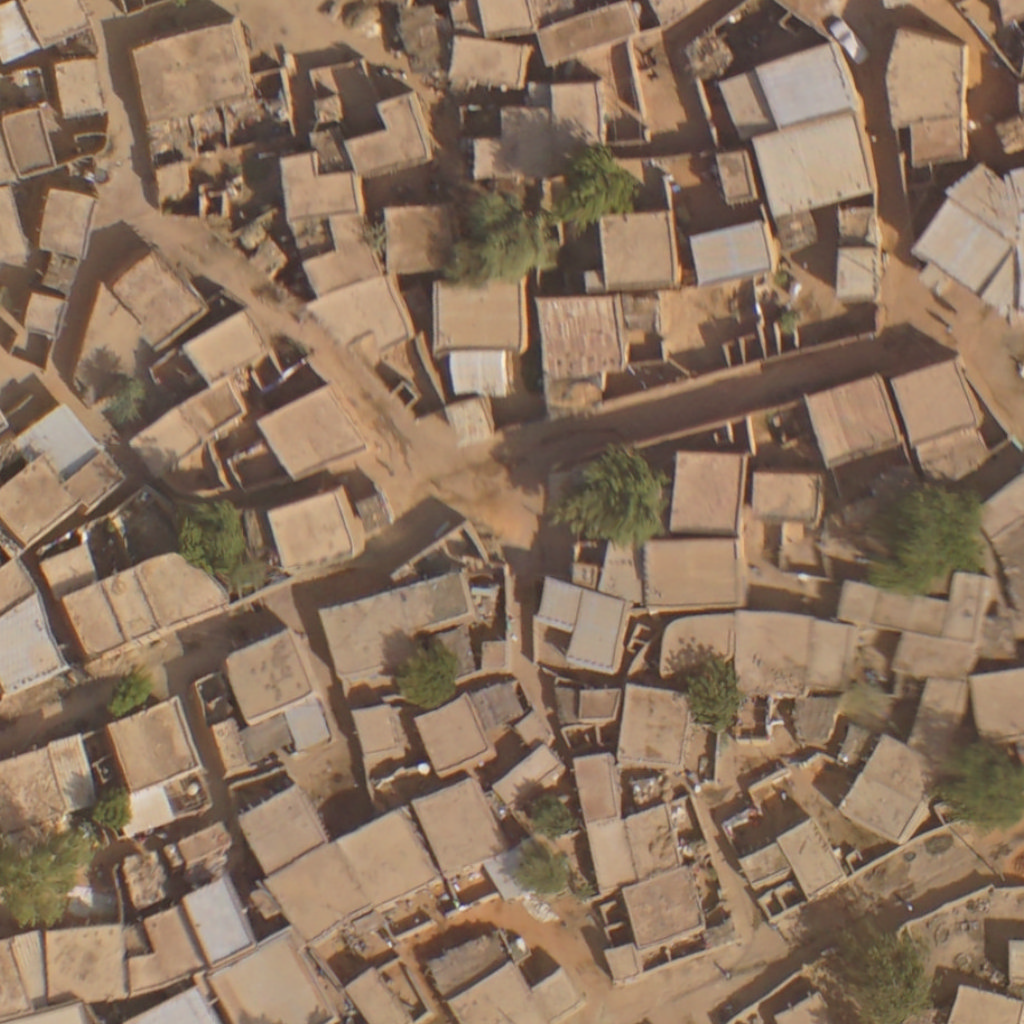} &
 \includegraphics[width=0.7in]{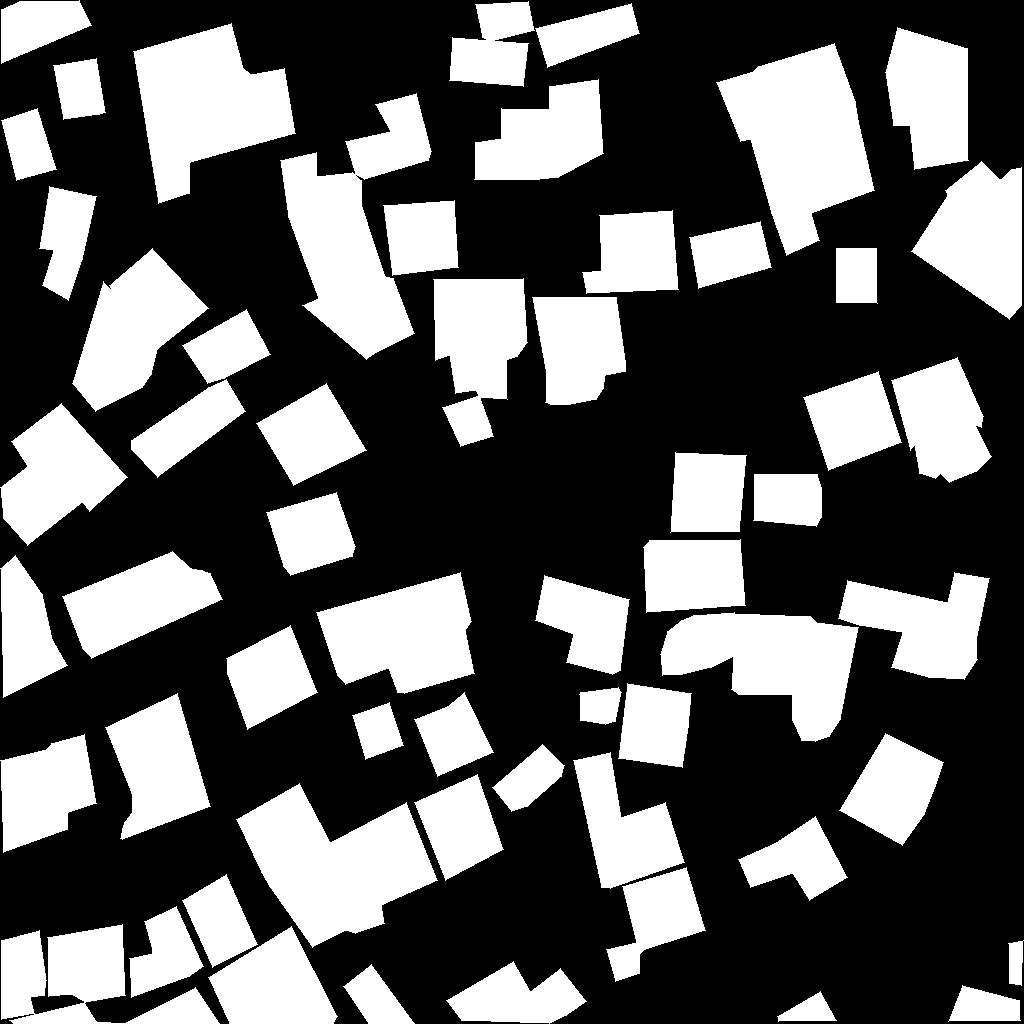} &
 \includegraphics[width=0.7in]{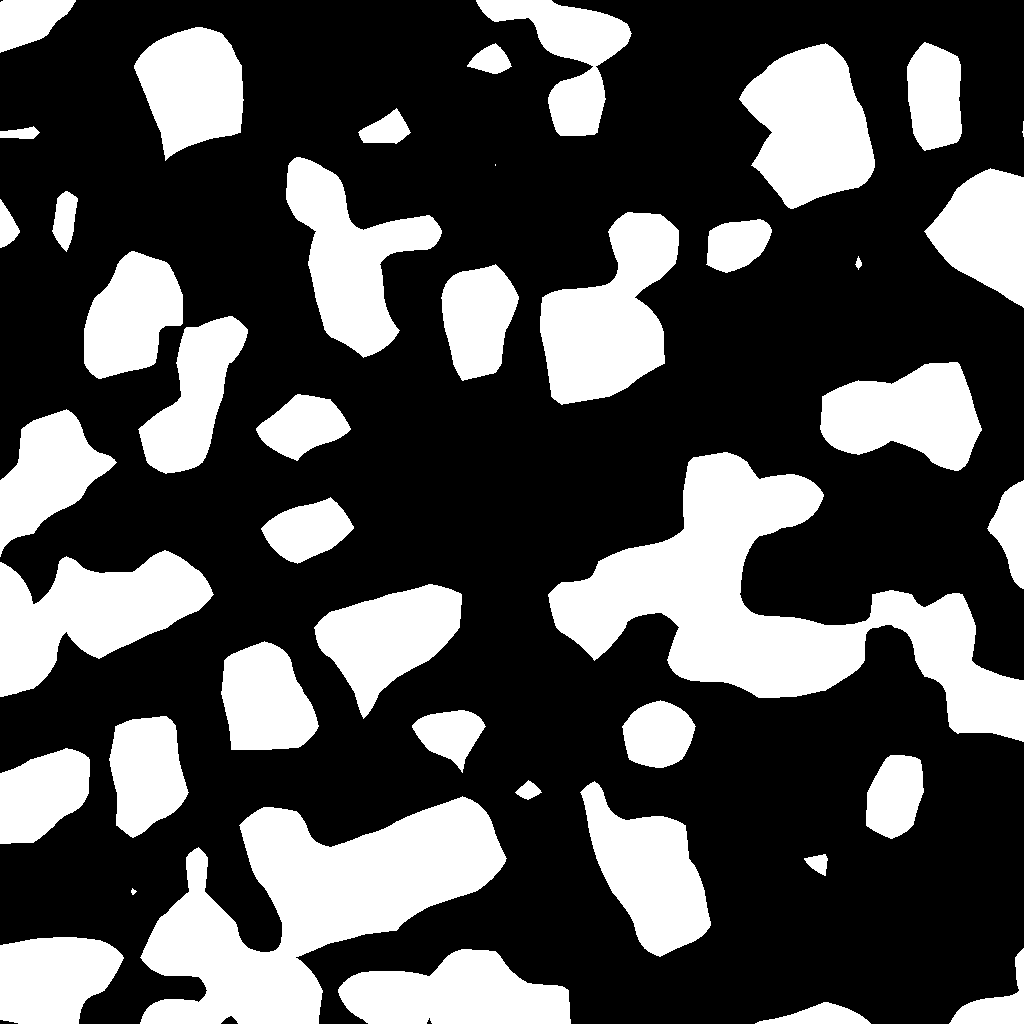} &
 \includegraphics[width=0.7in]{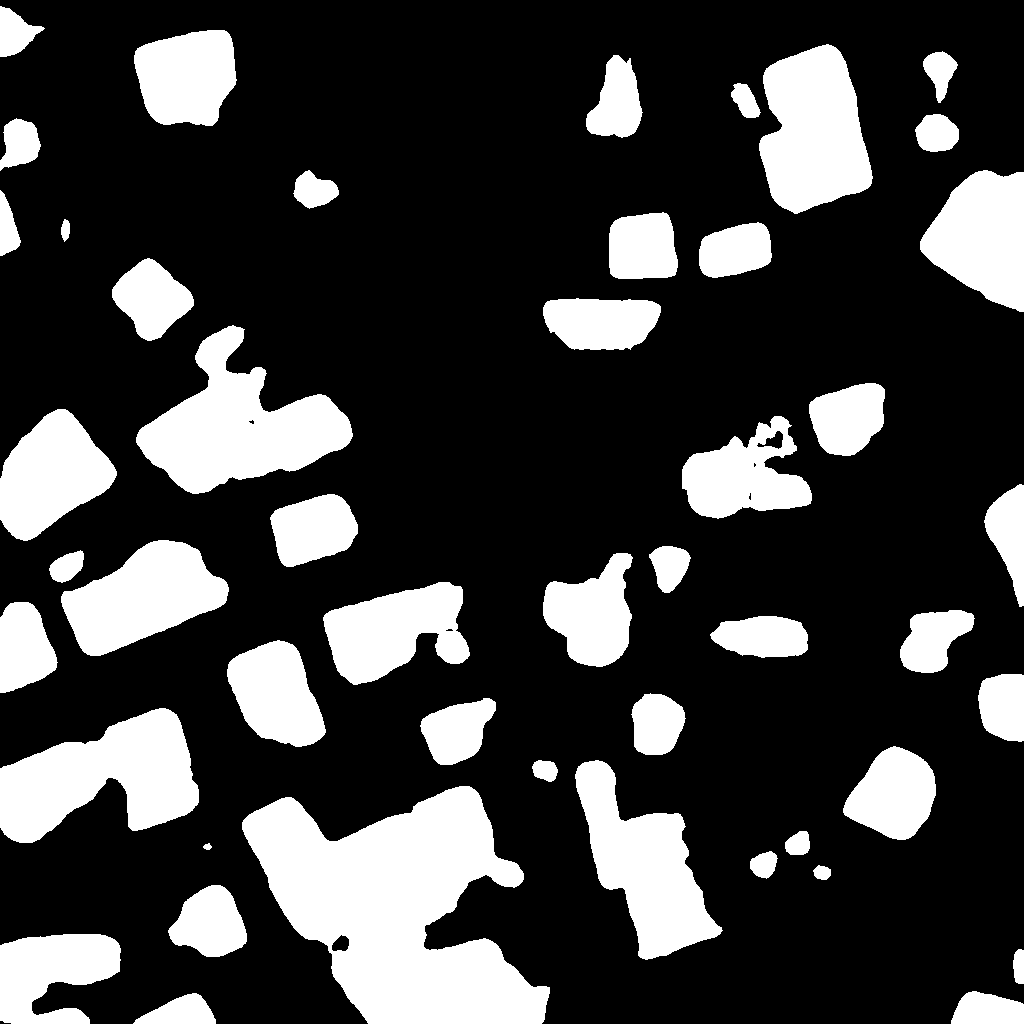} &
 \includegraphics[width=0.7in]{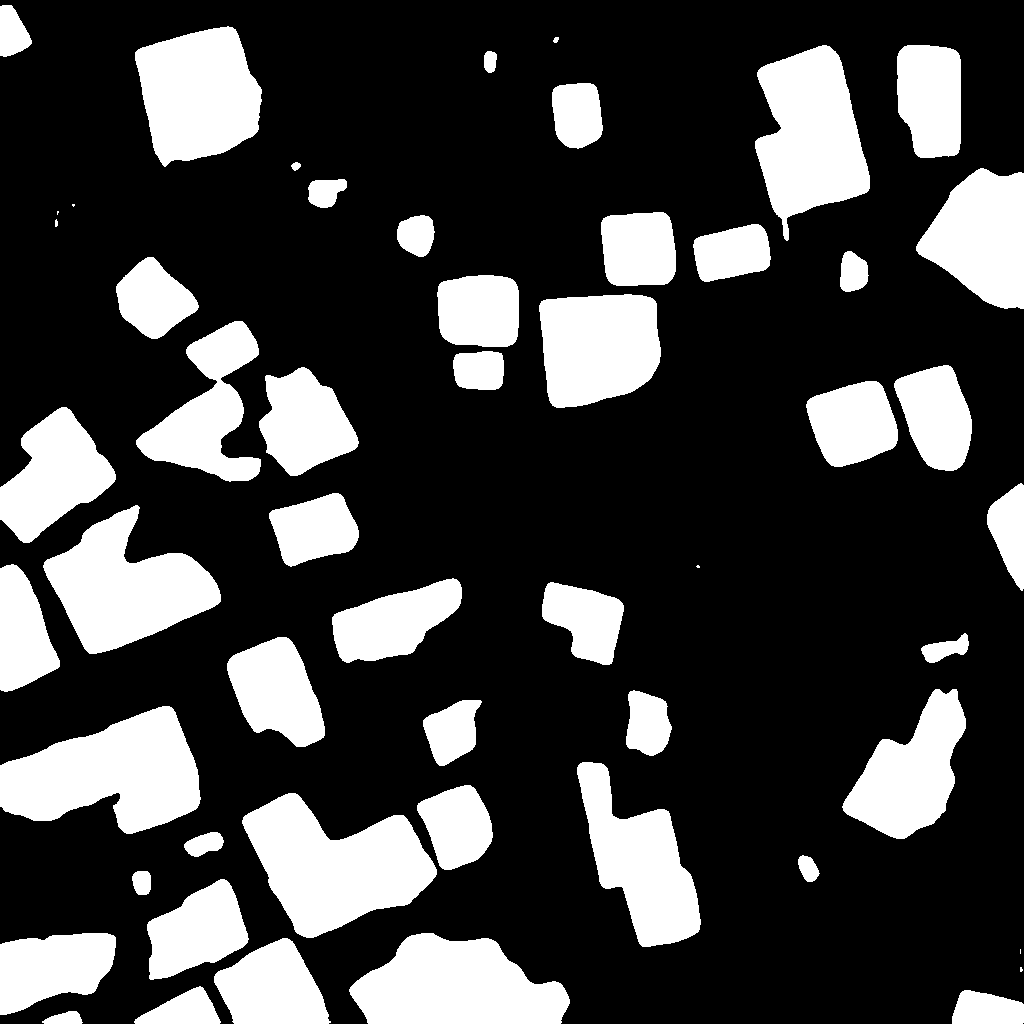} &
 \includegraphics[width=0.7in]{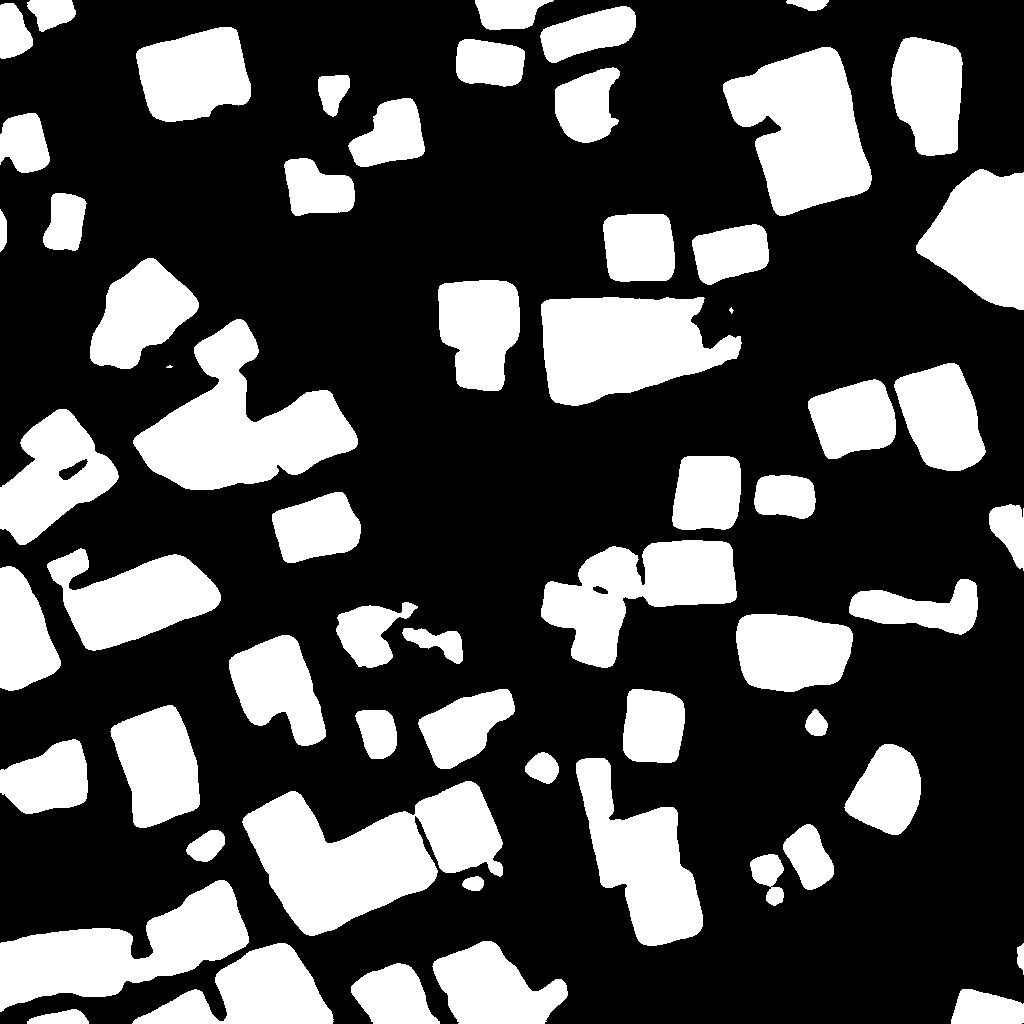} &
 \includegraphics[width=0.7in]{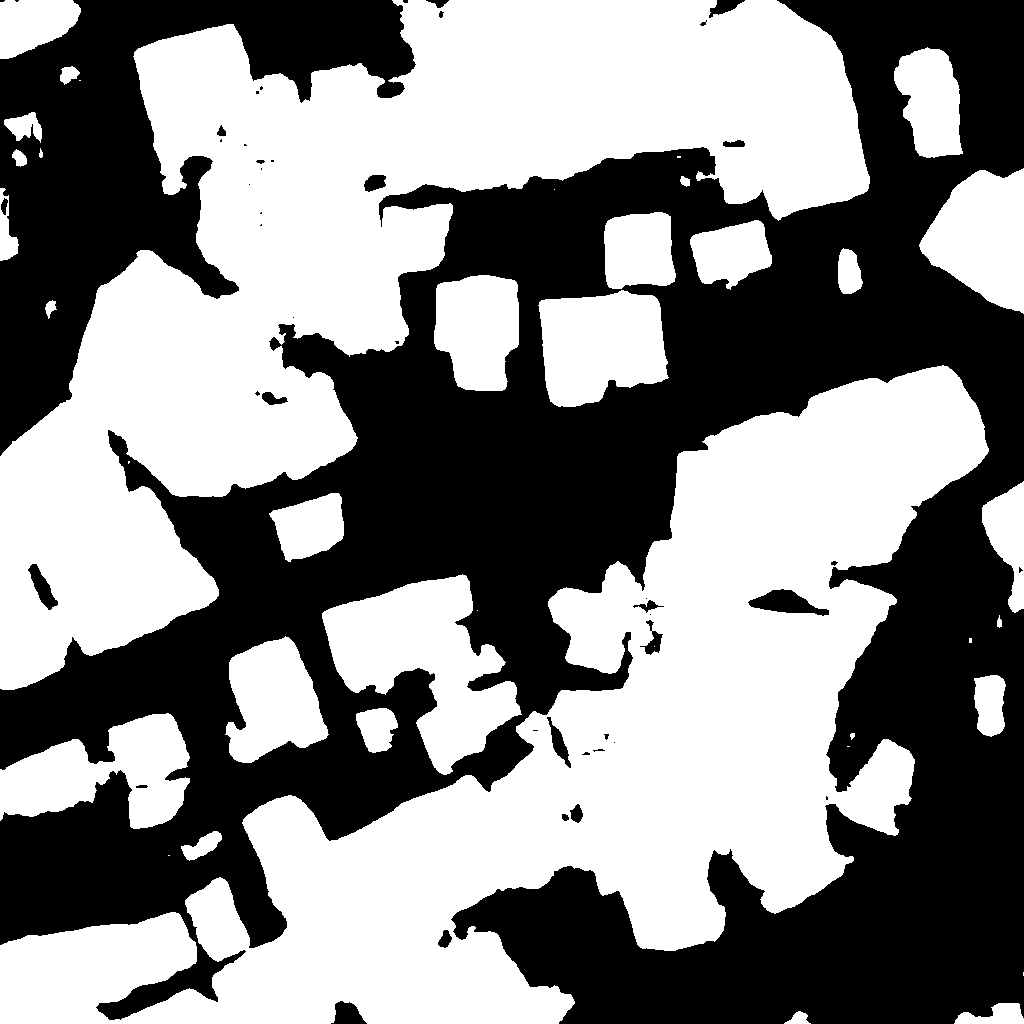} &
 \includegraphics[width=0.7in]{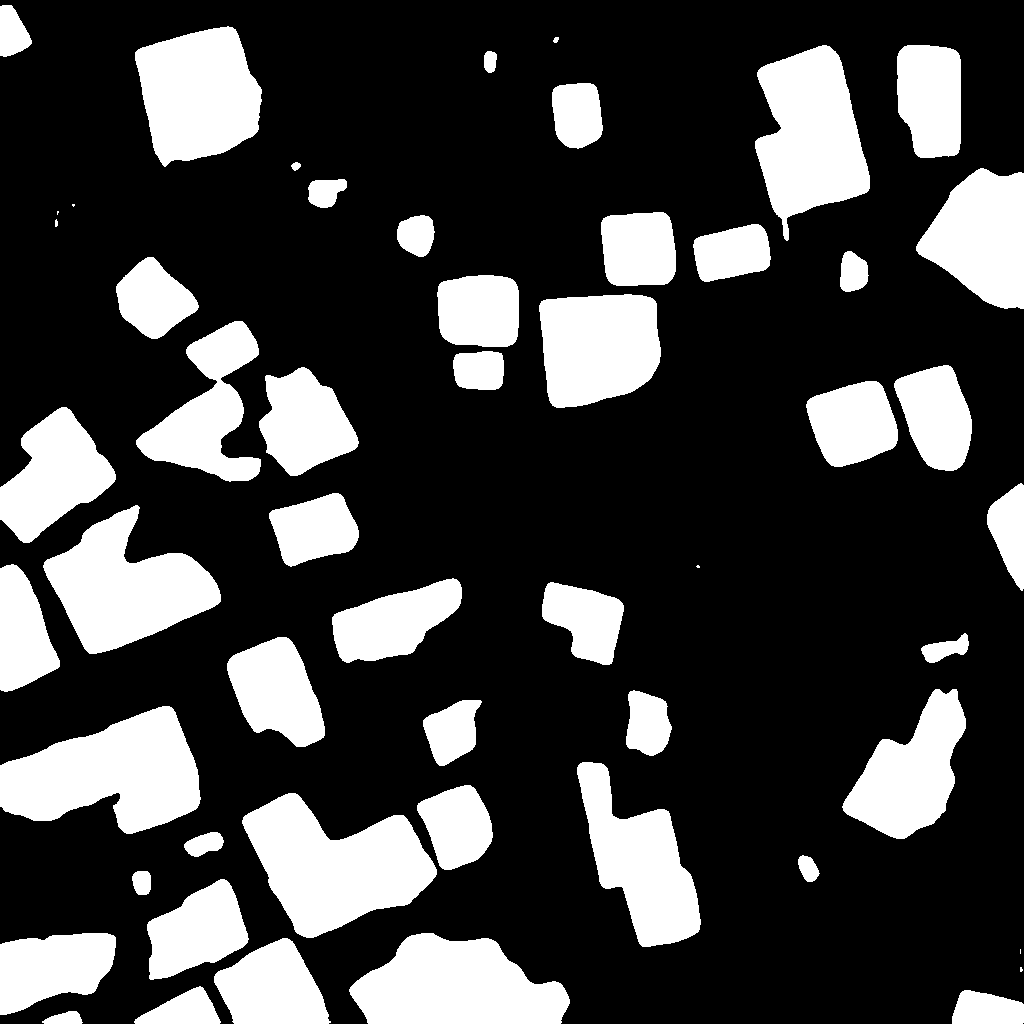} &
 \includegraphics[width=0.7in]{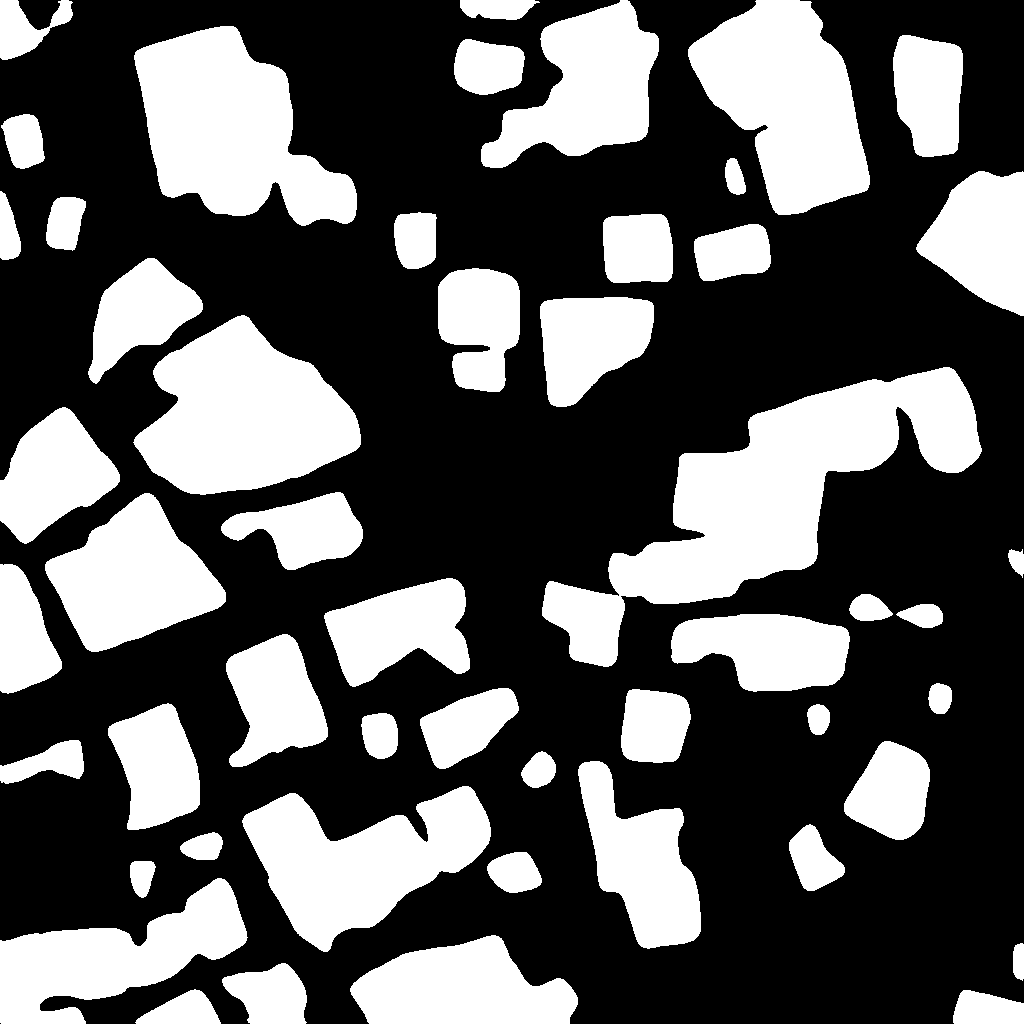} \\

 \includegraphics[width=0.7in]{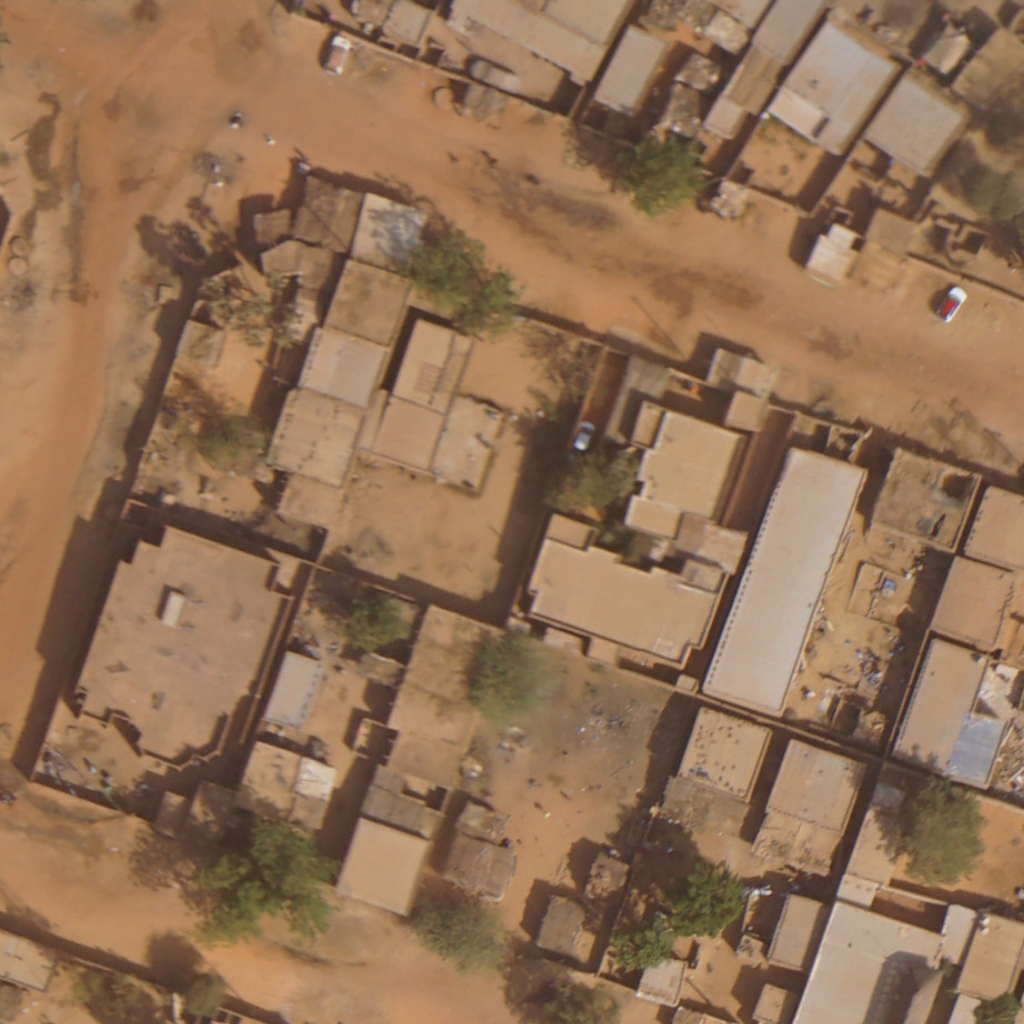} &
 \includegraphics[width=0.7in]{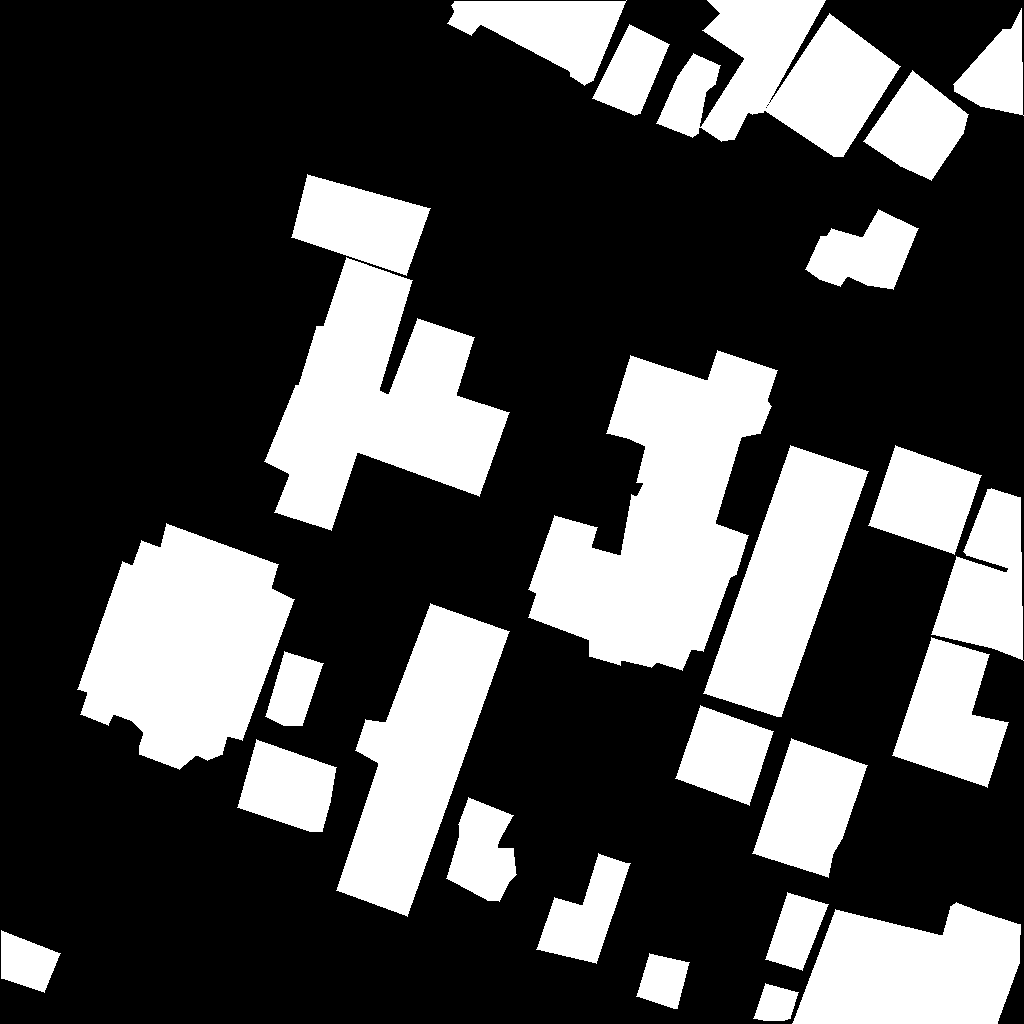} &
 \includegraphics[width=0.7in]{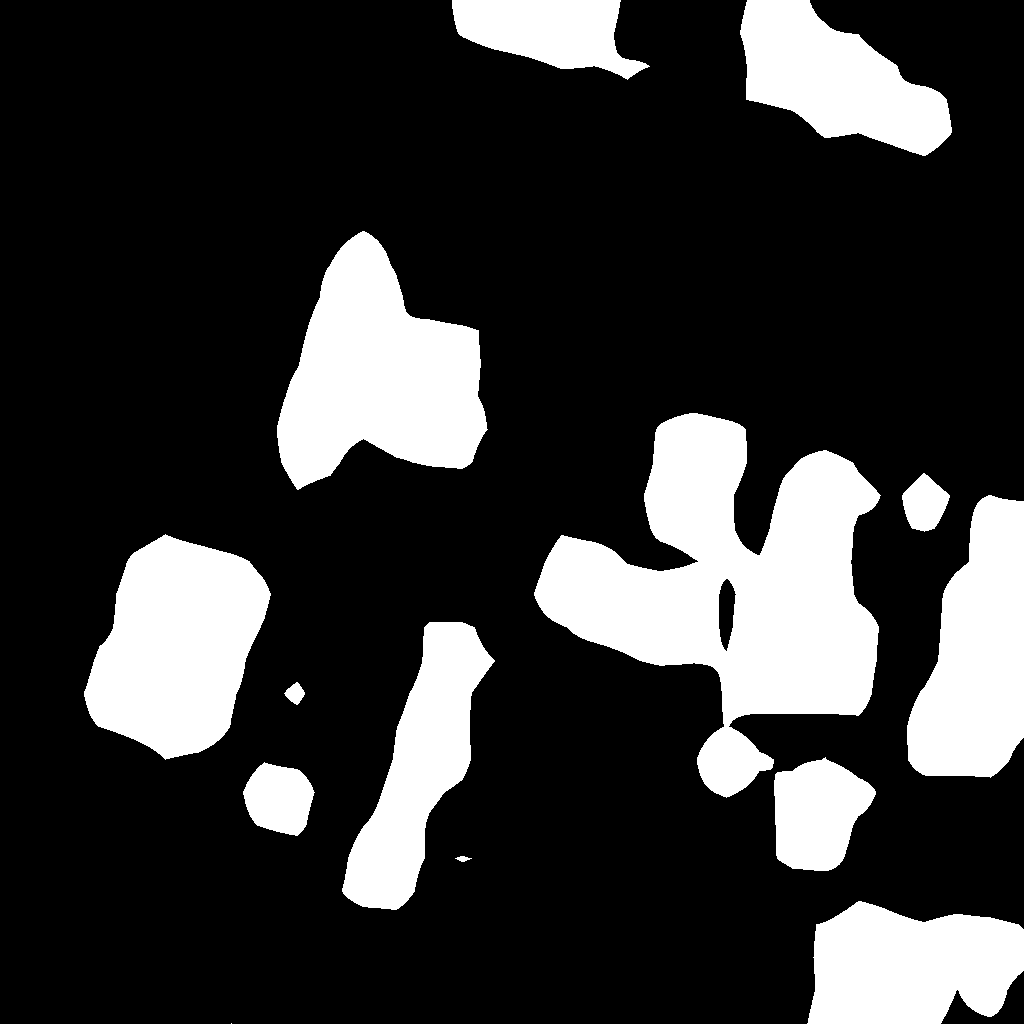} &
 \includegraphics[width=0.7in]{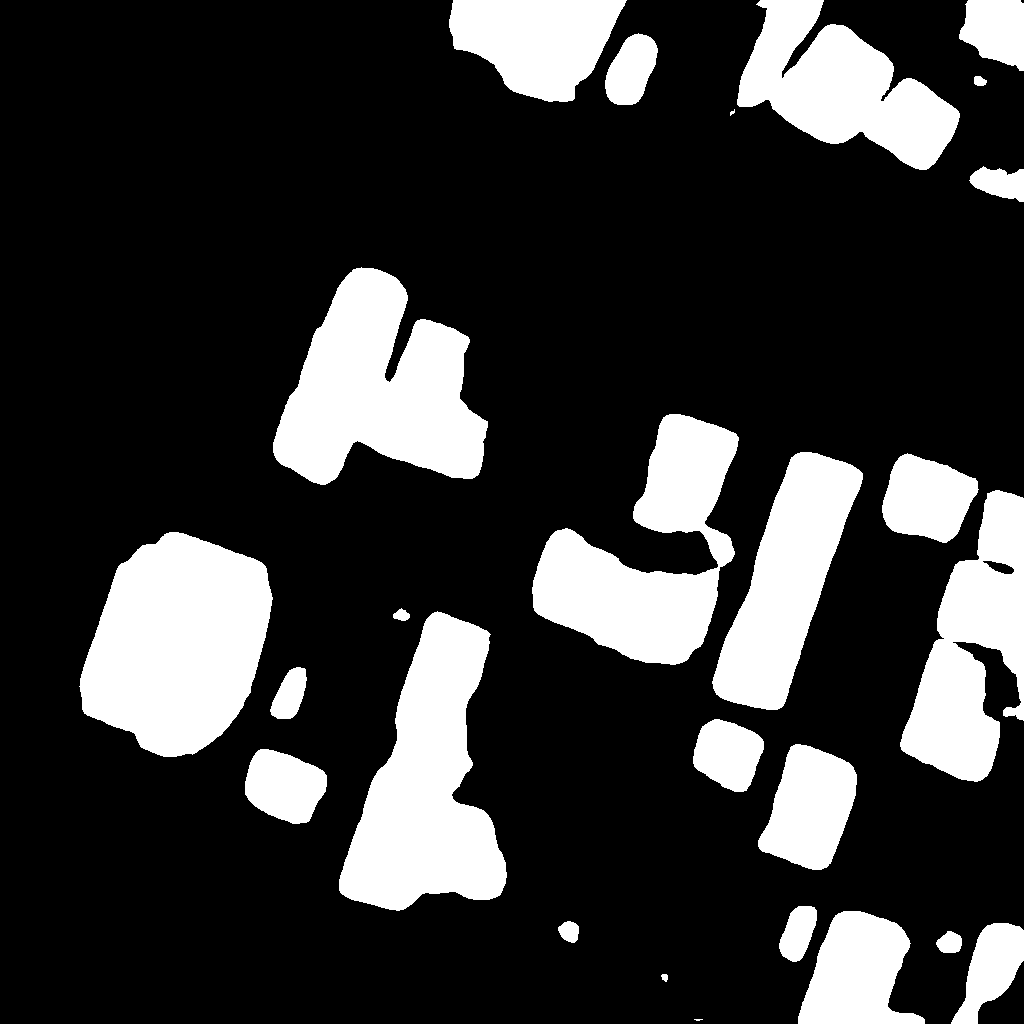} &
 \includegraphics[width=0.7in]{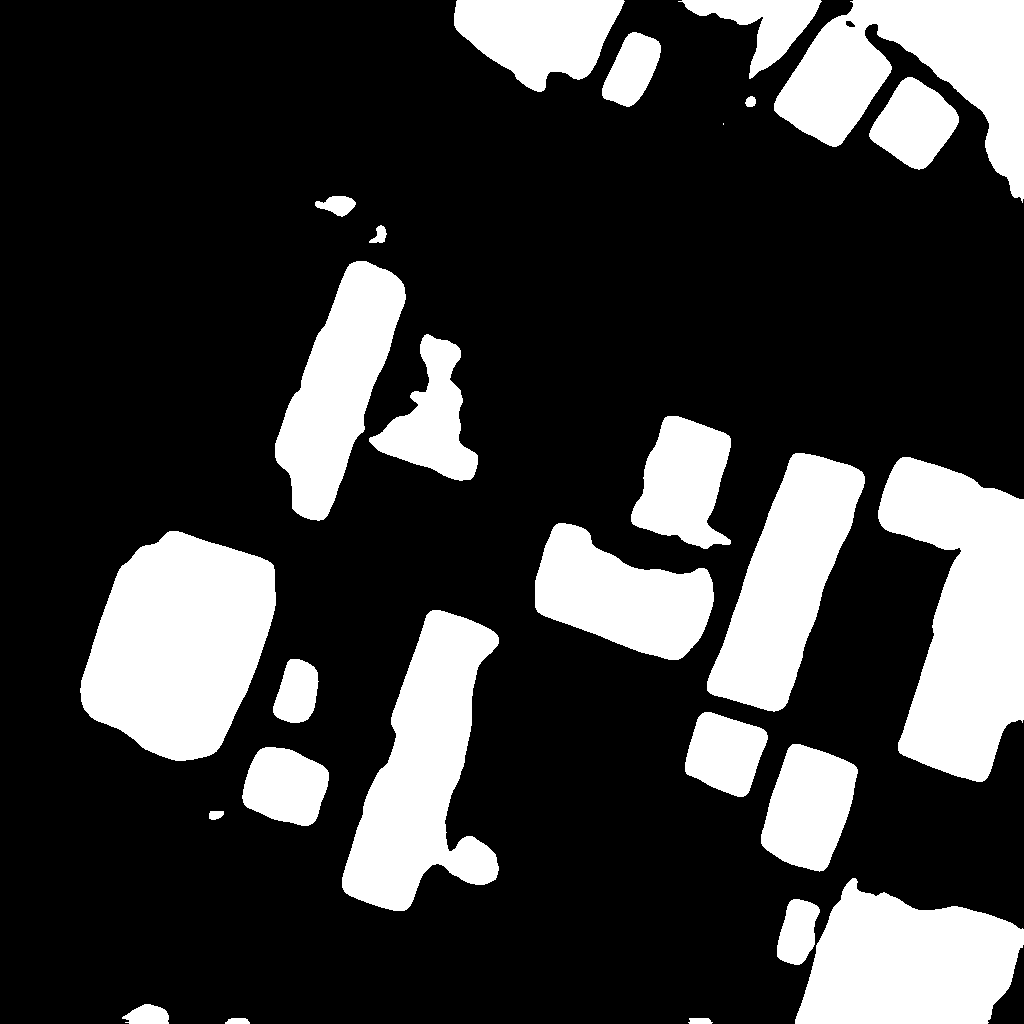} &
 \includegraphics[width=0.7in]{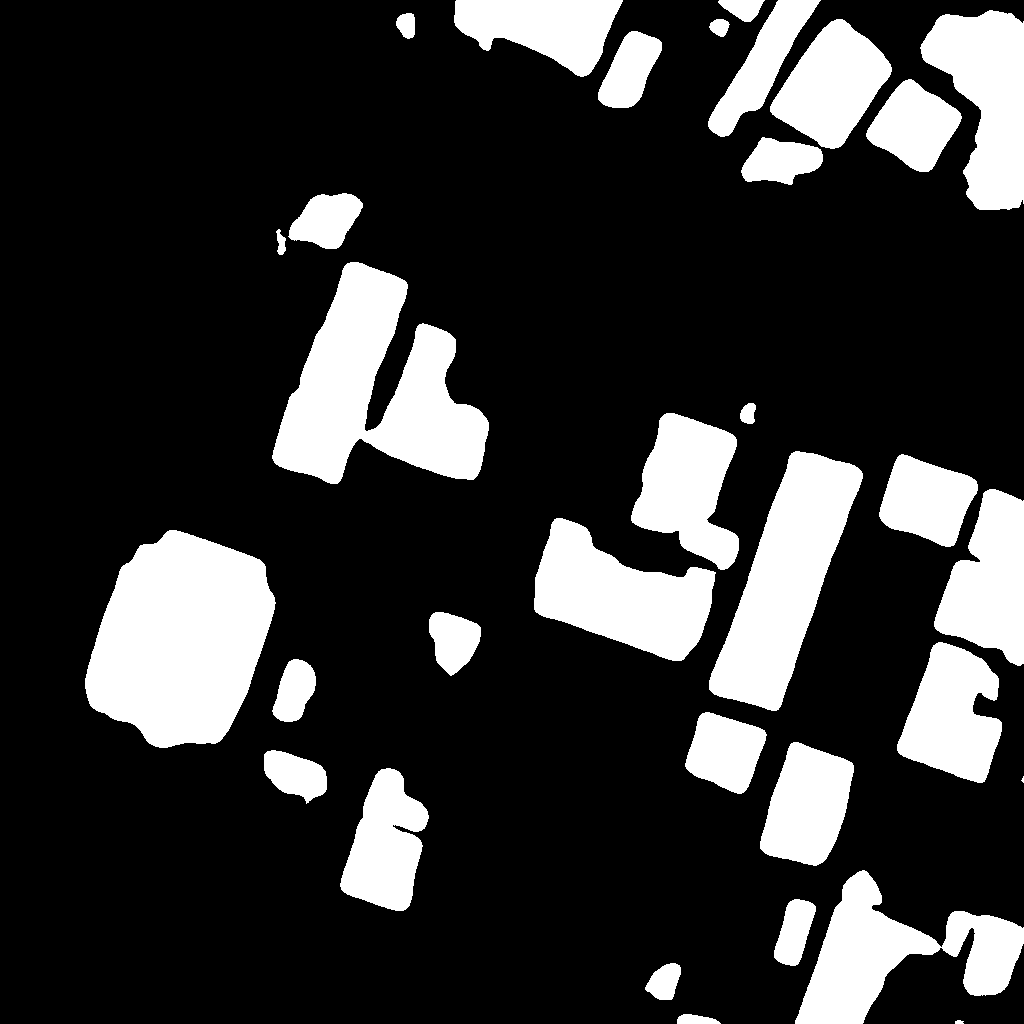} &
 \includegraphics[width=0.7in]{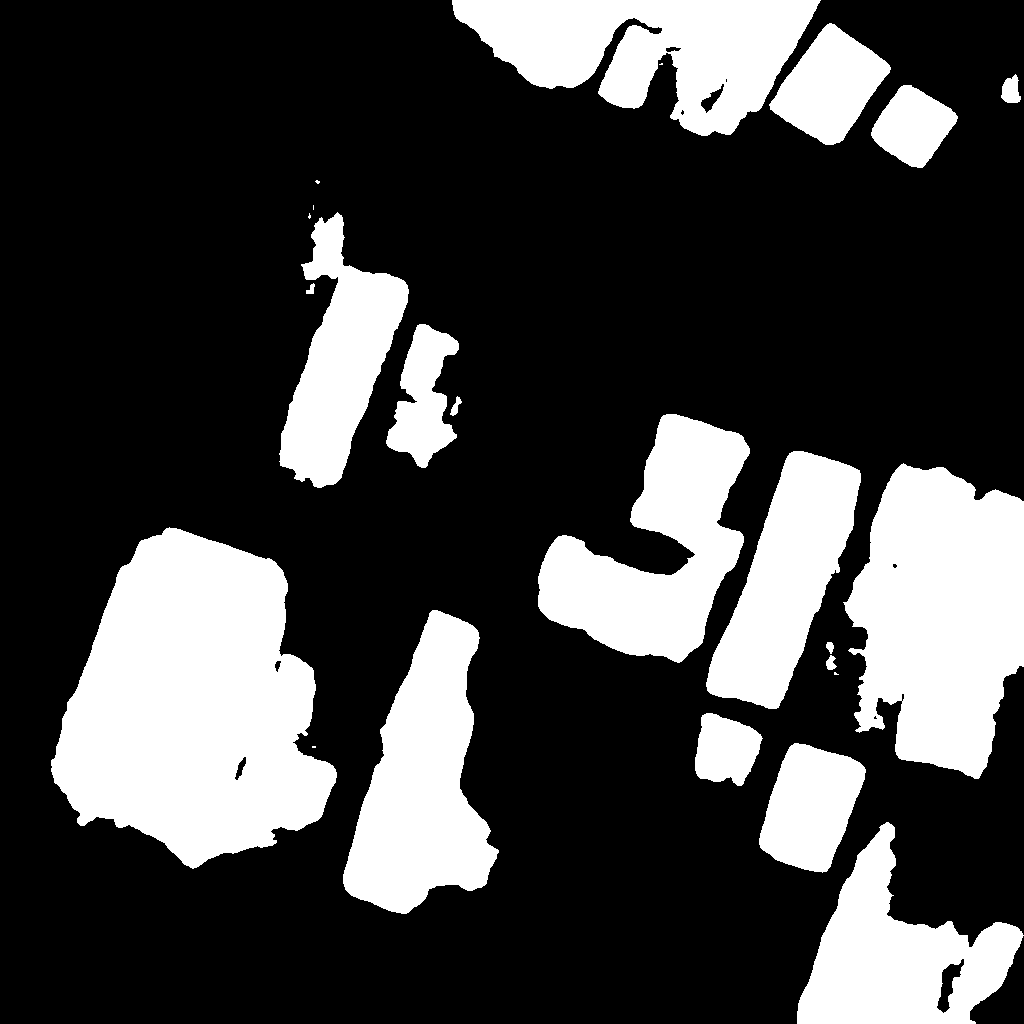} &
 \includegraphics[width=0.7in]{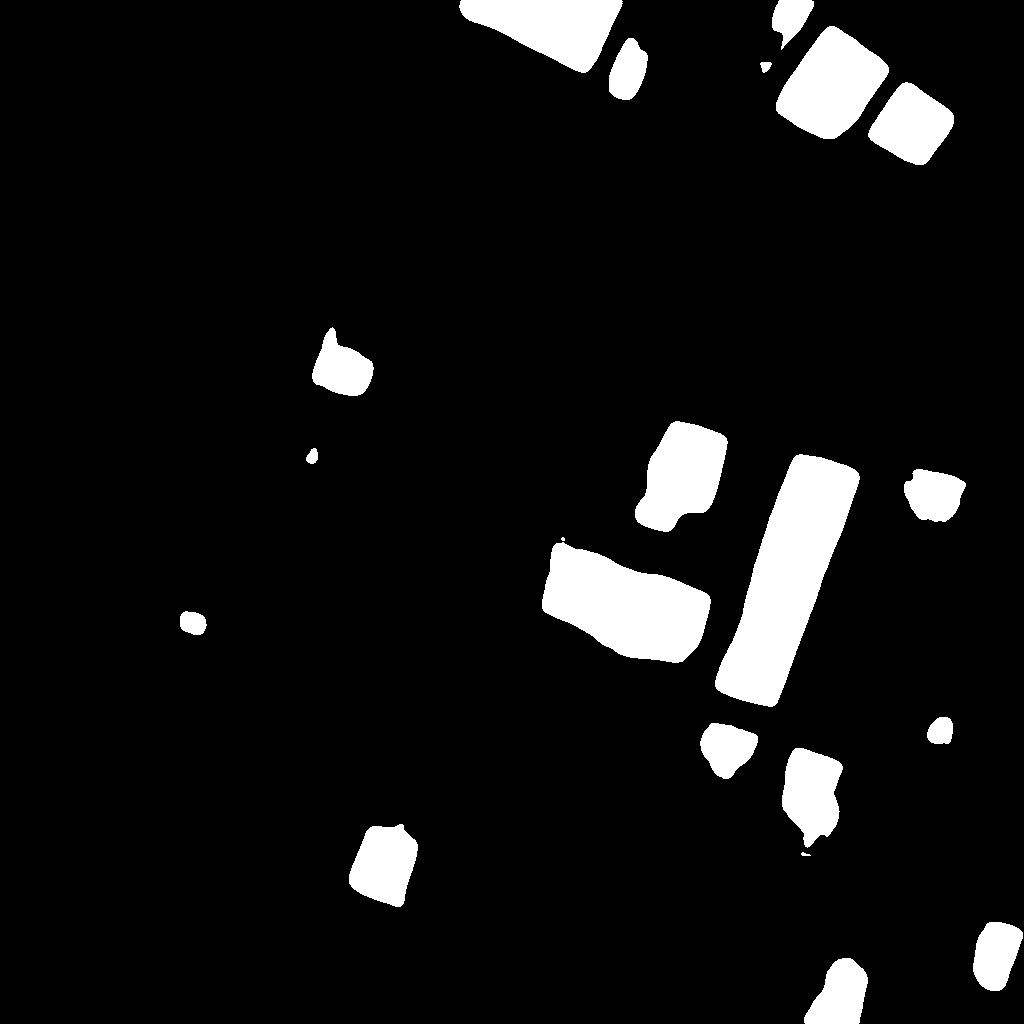} &
 \includegraphics[width=0.7in]{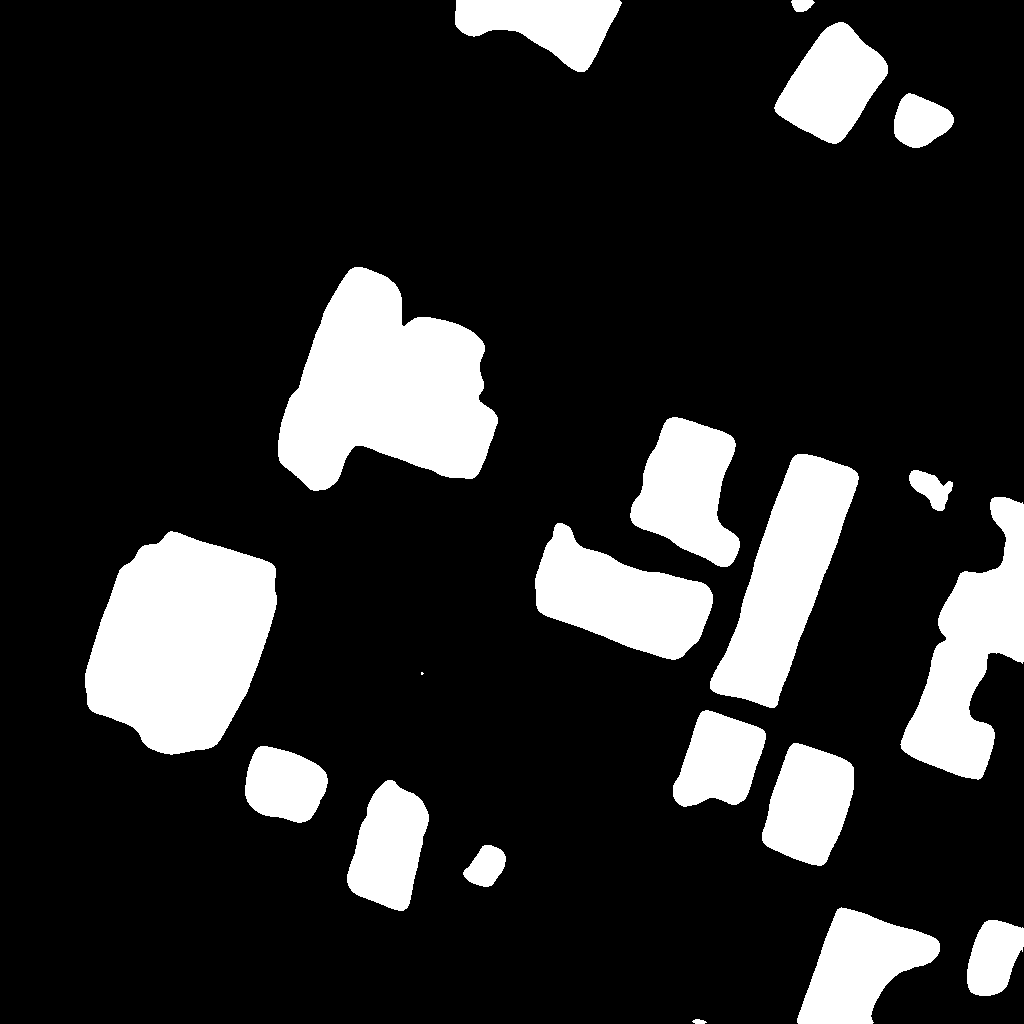} \\

 \end{tabular}
\caption{Ground Truth (GT) data and predicted masks for Sci-Net and benchmarked SoA models for various images from the OCAIC testset at different resolutions revealing fragmentation, under- and over-segmentation issues.}
\label{visualpreds}
\end{center}

\end{figure*}

\section{Experimental Results}
\label{resultss}
In this section, we present detailed experimental results and ablation studies of the proposed Sci-Net architecture.

\subsection{Evaluation on the OCAIC dataset}

Experimental results reveal the performance superiority of the proposed Sci-Net over several SoA models in terms of IoU and F1-score across varying resolutions. At inference time, the images are fed to the neural network at full scale ($1024\times1024$). Table~\ref{results1} shows that the proposed Sci-Net model provides a significant improvement of at least $2\%$ score over benchmarked SoA models on the OCAIC testset. Sci-Net attained a micro-IoU score of $82.25\%$. Table~\ref{results1} also shows that Sci-Net scored the highest macro-IoU value of $88.42\%$, which indicates that it leverages the best per-image performance.
Furthermore, at least $2\%$ score improvement margin is also observed for micro and macro F1-scores ($91.04\%$ and $92.62\%$ respectively). Thus, the proposed model attains better precision and recall against competitor models. Models like PSPNet and DeepLabV3+ failed to provide near good results at some resolutions leading to a degradation in their scores.

\begin{table*}[h]
\begin{center}
\begin{tabular}{c|cccc} \toprule
     {Model} & {micro-IoU} & {micro-F1} & {macro-IoU} & {macro-F1}  \\ \midrule
     {PSPNet \cite{psp}} & {57.42} & {72.95} & {54.90} & {69.72}  \\
     {DeepLabV3+ \cite{deeplabv3+}} & {74.16} & {85.17} & {71.58} & {82.84}  \\
     {MANet \cite{manet}} & {73.38} & {84.65} & {72.37} & {83.69}  \\
     {MSB \cite{msb_dataset}} & {72.30} & {83.92} & {70.34} & {82.18}  \\
     {\bf Sci-Net} & {\bf 75.18} & {\bf 85.84} & {\bf 73.71} & {\bf 84.53}  \\
     \bottomrule

\end{tabular}
\end{center}
\caption{Performance metrics of Sci-Net architecture compared to existing SoA models on the MSB testset. Sci-Net outperforms all benchmarked models in terms of micro-IoU, micro-F1, macro-IoU and macro-F1 scores, respectively.}
\label{results_msb}
\end{table*}

\begin{figure*}[h]
\begin{center}
\begin{tabular}{ccc | ccc}
 \textbf{Image} &
 \textbf{GT} &
 \textbf{Sci-Net} &
 \textbf{Image} &
 \textbf{GT} &
 \textbf{Sci-Net}  \\

 \includegraphics[width=0.7in]{"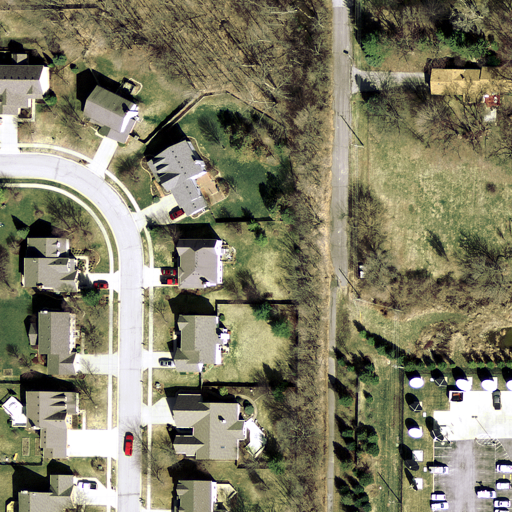"} &
 \includegraphics[width=0.7in]{"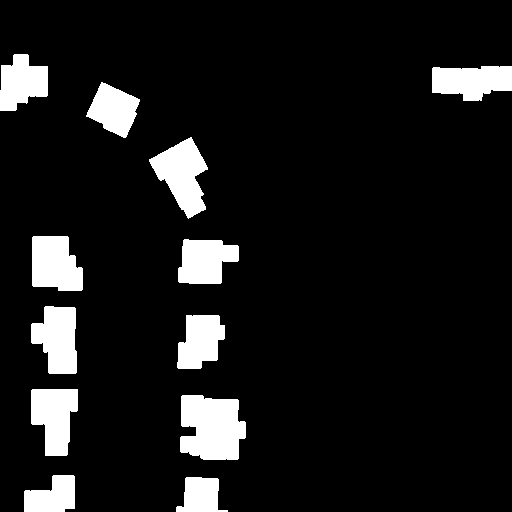"} &
 \includegraphics[width=0.7in]{"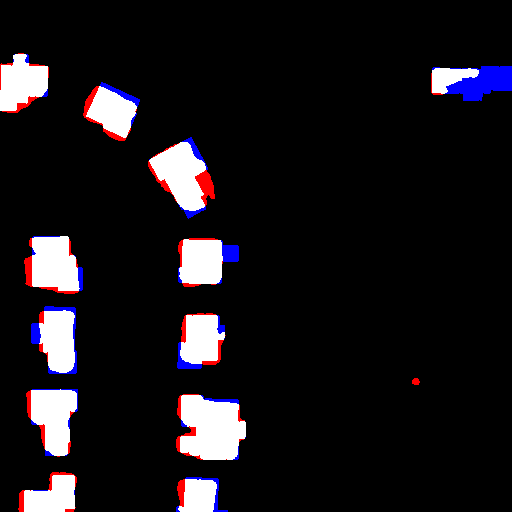"} &

 \includegraphics[width=0.7in]{"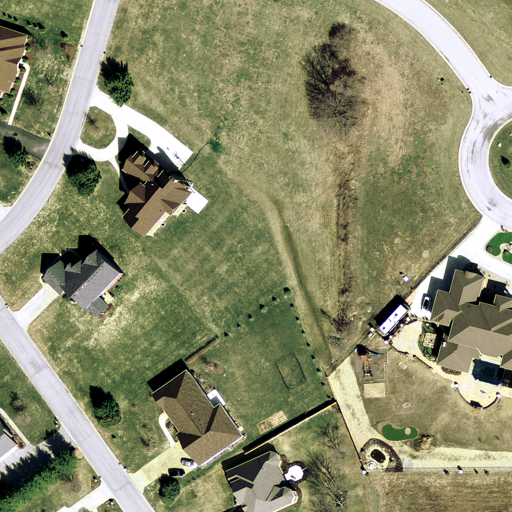"} &
 \includegraphics[width=0.7in]{"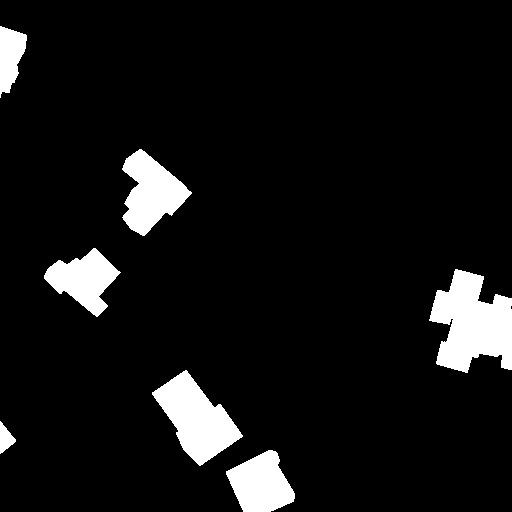"} &
 \includegraphics[width=0.7in]{"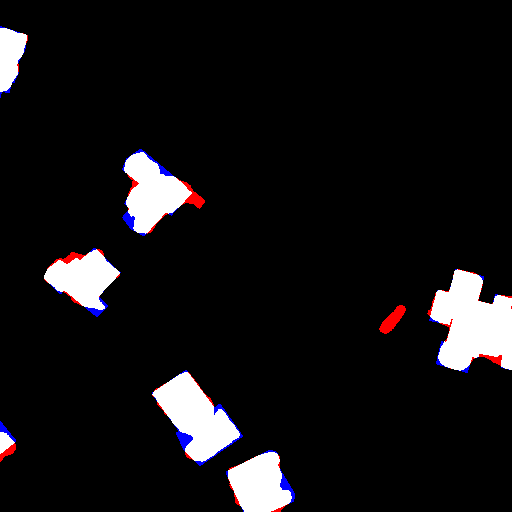"} \\

  \includegraphics[width=0.7in]{"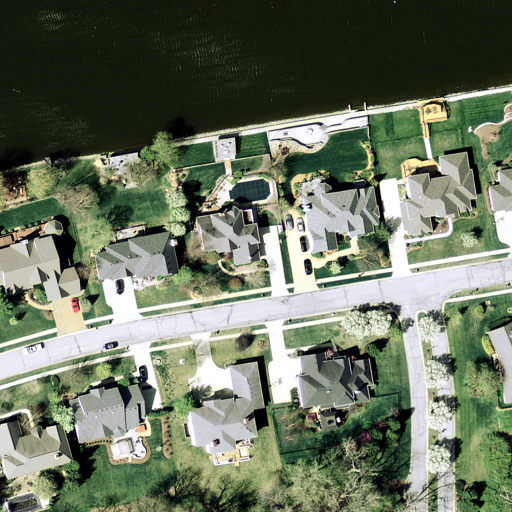"} &
 \includegraphics[width=0.7in]{"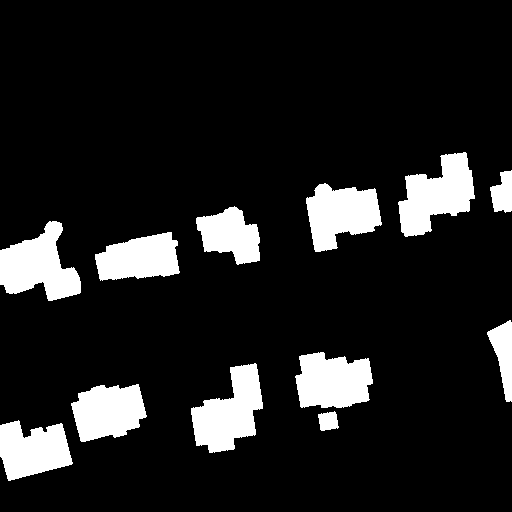"} &
 \includegraphics[width=0.7in]{"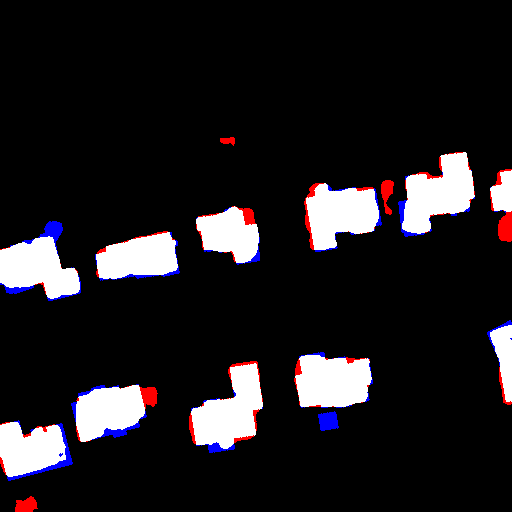"} &

  \includegraphics[width=0.7in]{"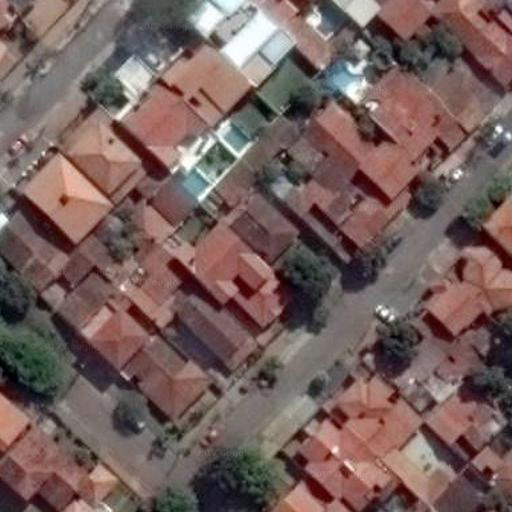"} &
 \includegraphics[width=0.7in]{"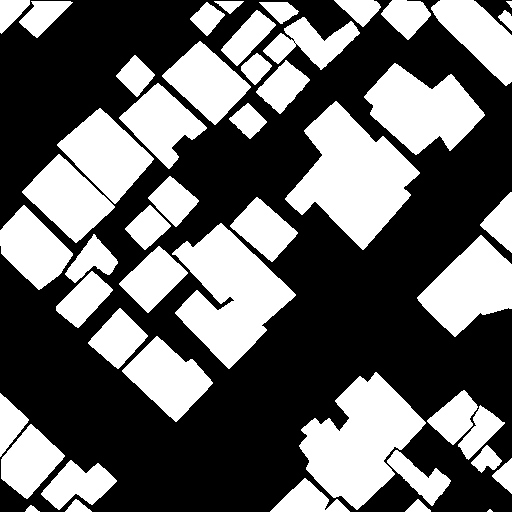"} &
 \includegraphics[width=0.7in]{"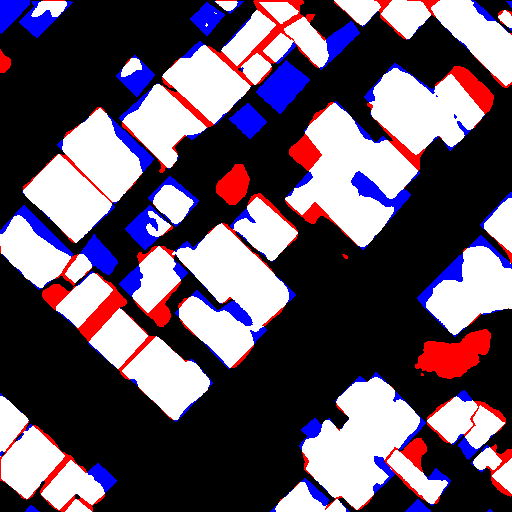"} \\

  \includegraphics[width=0.7in]{"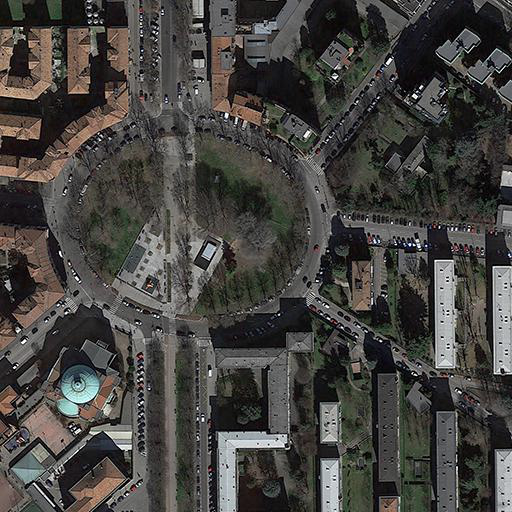"} &
 \includegraphics[width=0.7in]{"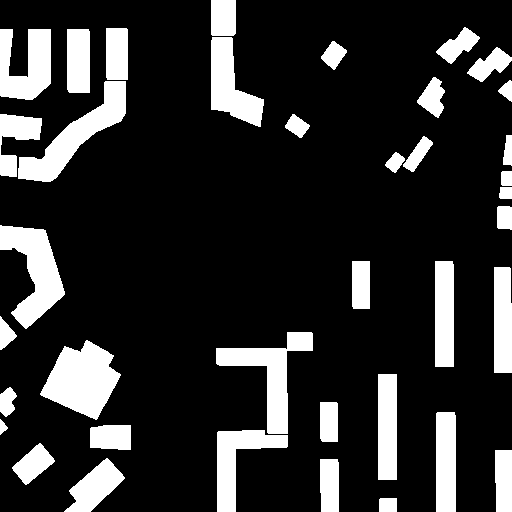"} &
 \includegraphics[width=0.7in]{"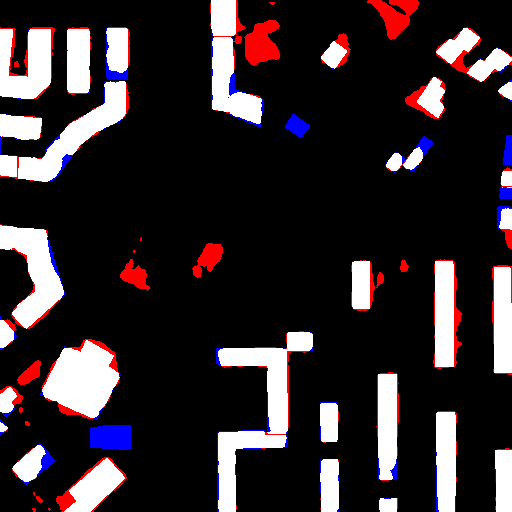"} &

  \includegraphics[width=0.7in]{"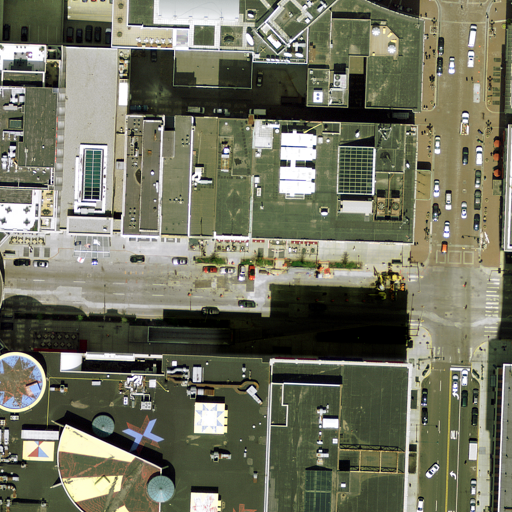"} &
 \includegraphics[width=0.7in]{"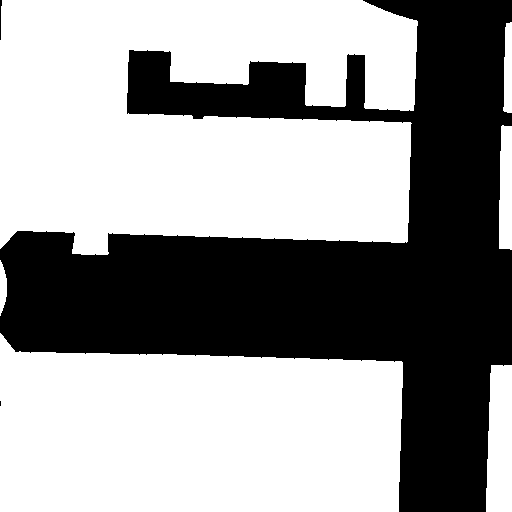"} &
 \includegraphics[width=0.7in]{"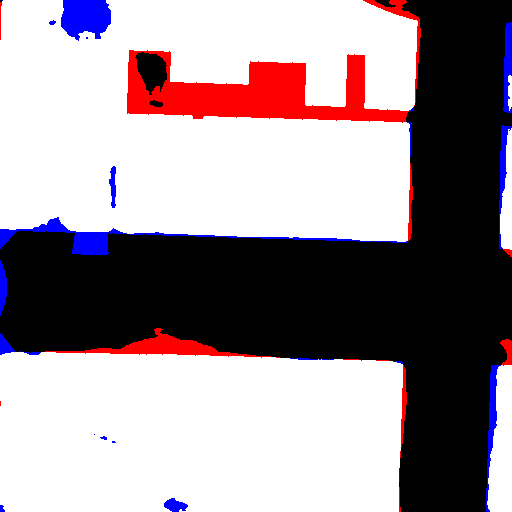"} \\

 \end{tabular}
\caption{Ground Truth (GT) data and predicted masks using Sci-Net for buildings with complex roofs in the MSB testset. White pixels denote TP, red pixels denote FP, black pixels denote TN and blue pixels denote FN. Sci-Net proved to be robust to different shapes and colors of complex roofs.}
\label{complex_roofs_results}
\end{center}

\end{figure*}

\begin{figure}[t]
\begin{center}
\begin{tabular}{p{-0.4cm}ccc}
\centering

 & \hspace{0.6cm} \textbf{Image} \hspace{0.85cm}& \hspace{0.85cm} \textbf{ERF Mask} & \hspace{0.3cm} \textbf{Masked Image} \\

\rotatebox{90}{\hspace{0.6cm}   UNet} & \multicolumn{3}{c}{\includegraphics[width=0.9\linewidth]{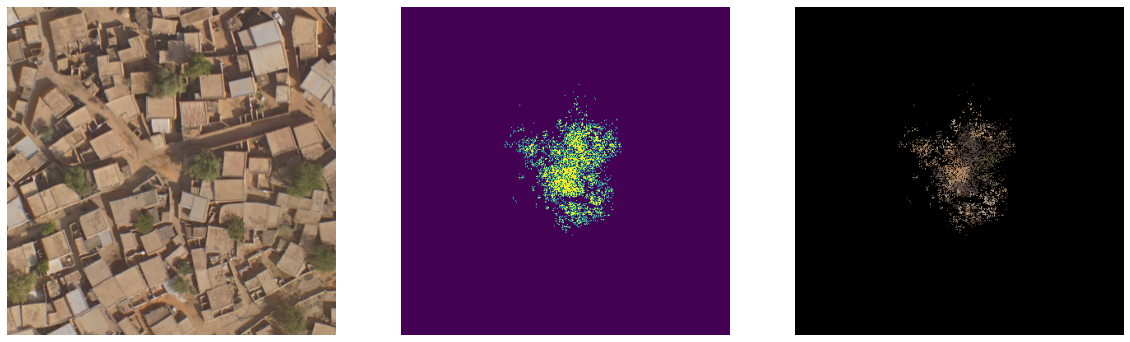}} \\
\rotatebox{90}{\hspace{0.5cm}   Sci-Net} & \multicolumn{3}{c}{\includegraphics[width=0.9\linewidth]{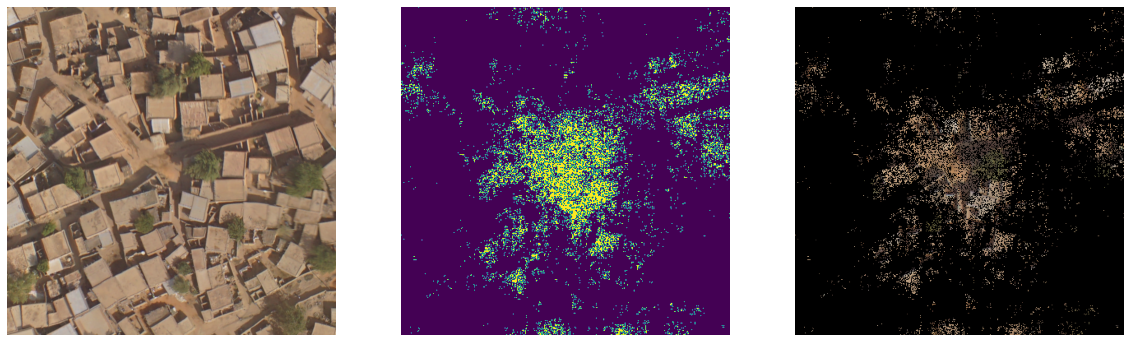}}  \\ \midrule

\rotatebox{90}{\hspace{0.6cm}   UNet} & \multicolumn{3}{c}{\includegraphics[width=0.9\linewidth]{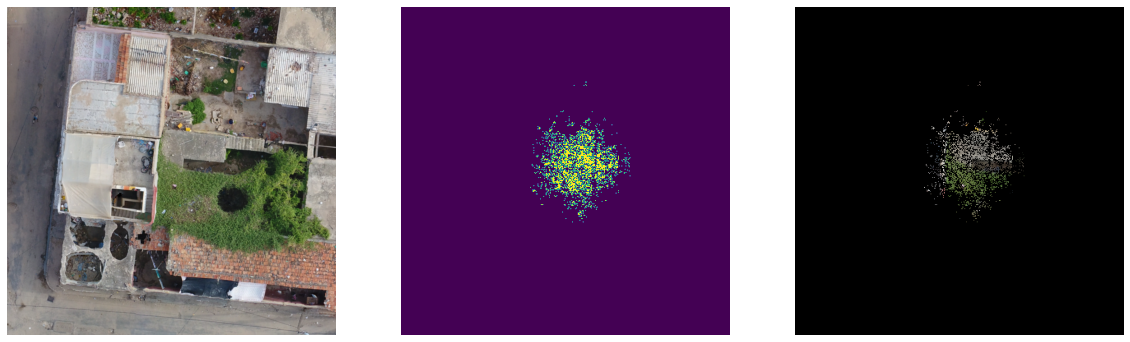}} \\
\rotatebox{90}{\hspace{0.5cm}   Sci-Net} & \multicolumn{3}{c}{\includegraphics[width=0.9\linewidth]{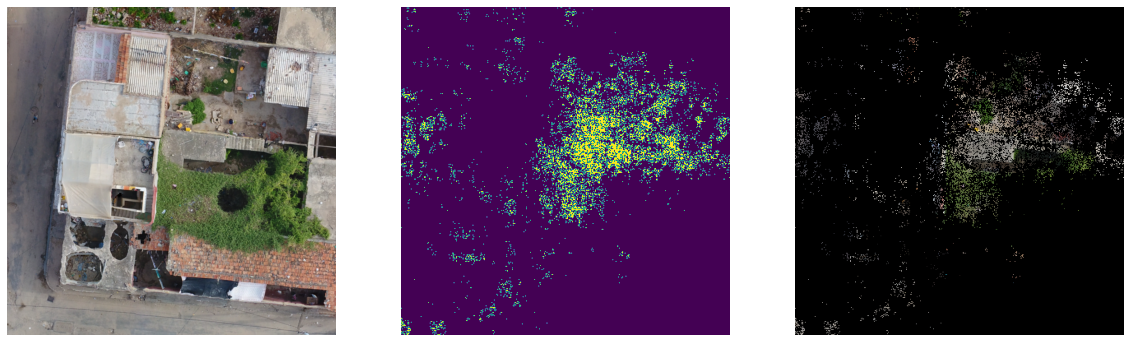}}  \\ \midrule

 \end{tabular}
\caption{Visualization of ERF: Sci-Net captures wider ERF compared to UNet.}
\label{erf}
\end{center}

\end{figure}

\begin{table*}[h]
\begin{center}
\begin{tabular}{c|cccc} \toprule
     {Model} & {micro-IoU} & {micro-F1} & {macro-IoU} & {macro-F1}\\ \midrule
     {UNet \cite{unet}} & {80.18} & {89.00} & {86.01} &{90.16} \\
     {UNet + ASPP} & {80.47} & {89.18} & {86.65} &{90.79}\\
     {\bf Sci-Net} & {\bf 82.25} & {\bf 91.04} & {\bf 88.42} &{\bf 92.62} \\
     \bottomrule

\end{tabular}
\end{center}
\caption{Ablation Study; Sci-Net brings a significant improvement on the OCAIC testset, compared to  UNet and  UNet + ASPP.}
\label{tab:ablation}
\end{table*}

\begin{table}[h]
\begin{center}
\begin{tabular}{c|cc} \toprule
     {Loss Function} & {micro-IoU} & {micro-F1} \\ \midrule
     {BCE} & {71.57} & {83.072}   \\
     {Dice} & {72.03} & {83.41} \\
     {\bf BCE+Dice} & {\bf 75.18} & {\bf 85.84}  \\
     \bottomrule

\end{tabular}
\end{center}
\caption{Micro-IoU and micro-F1 scores attained after training Sci-Net on the MSB testset \cite{msb_dataset} using different loss functions. The model achieves higher score when using a combination of both Dice and BCE losses.}
\label{tab:ab_study_loss}
\end{table}

Furthermore, we plot in Figure~\ref{resultsgraphs} the micro and macro IoU and F1-scores per resolution (cm/pixel) for every benchmarked model using the OCAIC testset. The green curve corresponding to the Sci-Net model always performs better than all other models across different metrics and resolutions, which shows that it can effectively extract better multi-scale representations than the existing models. Sci-Net curve is consistently above all other curves for the four presented score graphs, which indicates that it is less prone to performance degradation when the scale changes. It is unclear which model holds the "runner up" spot, as these models alternate places on varying resolutions. For instance, UNet+ASPP (in red) achieves competitive scores for resolutions in range (2cm/pixel up-to 8cm/pixel), however its performance deteriorates for larger resolutions. PSPNet (in brown) benchmarks the worst performance for all resolutions.

To better visualize our results, a sample of images from the OCAIC testset and the corresponding predicted masks by each model are shown in Figure~\ref{visualpreds}. For instance, problems like fragmentation and under-fitting are solved using Sci-Net by acquiring sufficiently large receptive fields capable of relating far pixels that belong to large building instances at high resolutions (Masks in rows 1, 2, 3, and 4). In the first four rows, it is clear that Sci-Net succeeds in segmenting large building instances. For example, models like DeepLabV3+, UNet, and HRNet showed a critical level of mask fragmentation for that large building instance in the first row. Also, at lower resolutions (rows 6 and 7), Sci-Net can avoid over-segmentation, unlike other architectures such as HRNet that misclassified a significant amount of background pixels, as shown in row 6.

While models that use pyramid pooling like PSPNet performed well in segmenting large building instances (row 1 and 2), they often fails in capturing small to medium-sized buildings at lower resolutions. Unet+ASPP could not capture enough multi-scale information to segment large structures properly (rows 1, 2, and 3). Hence, Sci-Net proved to be the most efficient approach across various spatial resolutions.

In terms of model complexity, Table \ref{results1} shows that our model is smaller than MANet and HRNet, which indicates that the improvements is not due only to adding more modules/parameters.

\subsection{Evaluation on the MSB dataset}
We conducted more experiments using the MSB dataset\cite{msb_dataset} to verify and validate claimed advantages of our proposed Sci-Net architecture.
As shown in Table~\ref{results_msb}, Sci-Net attained a micro-F1 score of 85.84\% and a macro-F1 score of 84.53\% which is 2\% higher than the MSB model results reported in \cite{msb_dataset}.\\
DeepLabV3+ falls short behind by 1\% in terms of micro-F1 but is 2\% lower than Sci-Net in terms of macro-F1 scores. This shows that Sci-Net is ensured to perform well on a per-image and per-dataset basis. Indeed, Sci-Net outperforms other benchmarked models in Table~\ref{results_msb} in terms of all measured metrics.

Similar to \cite{msb_dataset}, we explored the robustness of our proposed model for difficult scenarios with complex roofs and evaluated how sensitive is Sci-Net toward outliers. In Figure~\ref{complex_roofs_results}, we plot colored masks inference results for images with complex roofs. White pixels denote TP, red pixels denote FP, black pixels denote TN and blue pixels denote FN. Sci-Net performs outstanding building segmentation task with low False Positives and low False Negatives for those complex roofs. Sci-Net resulted in an F1-score between 85\% and 94\% which outperforms results reported in \cite{msb_dataset} for all complex roof image scenarios. For space limitations, we did not report F1-score for each image in Figure~\ref{complex_roofs_results}.

\subsection{Effective Receptive Field (ERF) Analysis}
The Effective Receptive Field (ERF) or Field of View is of crucial importance for dense prediction tasks, such as semantic segmentation. The larger the receptive field the better to capture a larger context about a given pixel, which is desirable, especially for large objects. The Receptive Field (RF) can be computed manually, however, in \cite{luo2016understanding_erf}, authors shows that the ERF is much lower than the theoretical RF.
Similarly to \cite{luo2016understanding_erf}, we use the back-propagation of gradients to visualize ERF. Specifically, we back-propagate the gradient of the central pixel backward to the input image and visualize the gradients. Results are shown in Figure \ref{erf}, where we show that the ERF of Sci-Net is wider than that of UNet, validating the importance of the proposed architecture.

\subsection{Ablation Studies}
In this subsection, we investigate the importance of our design choices in terms of Dense ASPP and loss function adapted.

\paragraph{Dense ASPP:}
From Table \ref{tab:ablation}, we can notice that the dilated convolutions at the bottleneck (UNet+ASPP), brings only slight improvements to the UNet model. On the other hand, the Dense ASPP module leads to significant improvements compared to ASPP for the UNet model. This observation also holds for the evaluation on several resolutions as shown in Figure \ref{resultsgraphs} and the quantitative comparison in Figure \ref{visualpreds}.

\paragraph{Loss function:}
To justify the choice of loss function used for model training, we conducted three experiments where Sci-Net was trained on the Multi-Scale Building dataset \cite{msb_dataset} to test the model performance with/without each component of the loss function. As stated in Section \ref{train_pipeline}, we are using a weighted combination of Dice loss and Binary Cross-Entropy ($BCE$) loss.  In Table \ref{tab:ab_study_loss}, we show that the model achieves 3\% higher micro-IoU score when using BCE+Dice loss compared to using each one of them alone.

\section{Conclusion}
\label{conclusion}
This paper proposes Sci-Net, a new model capable of accurately segmenting buildings' footprint at multi-scale spatial resolutions. We compare the performance of Sci-Net with other well-known SoA models using both OCAIC and MSB datasets. We show that the proposed Sci-Net architecture is useful to mitigate fragmentation, over-segmentation, and under-segmentation problems that SoA models suffer from.

\section*{Declarations}

Ethical Approval: not applicable

Competing interests: none

Authors' contributions: H.N and A.G wrote the manuscript text. H.N. and M.S. prepared the figures. All authors reviewed the manuscript.

Funding: not applicable

Availability of data and materials: not applicable


{\small
\bibliographystyle{ieee_fullname}
\bibliography{references}
}


\end{document}